\documentclass[acmtog]{acmart}

\usepackage{booktabs} 
\setcitestyle{square}

\usepackage[ruled]{algorithm2e} 
\SetAlFnt{\small}
\SetAlCapFnt{\small}
\SetAlCapNameFnt{\small}
\SetAlCapHSkip{0pt}
\IncMargin{-\parindent}

\usepackage{soul}
\usepackage{physics}
\usepackage{mathtools}
\usepackage{xfrac}
\usepackage{xcolor}
\usepackage{graphicx}
\usepackage{epstopdf}
\usepackage{subcaption}
\usepackage{geometry}\usepackage{tabularx,ragged2e}

\renewcommand{\v}[1]{\vb{#1}}

\newcommand{\gv}[1]{\vb*{#1}}
\newcommand{\pddv}[2]{\dfrac{\partial #1}{\partial #2}}

\newcommand{\q}{\v{q}}
\newcommand{\qdot}{\dot{\v{q}}}
\newcommand{\qddot}{\ddot{\v{q}}}

\renewcommand{\u}{\v{u}}

\newcommand{\dt}{h}

\newcommand{\addrev}[1]{#1}
\newcommand{\remrev}[1]{}

\newcommand{\add}[1]{#1}
\newcommand{\rem}[1]{}

\setcopyright{acmlicensed}
\acmJournal{TOG}
\acmYear{2019} \acmVolume{38} \acmNumber{5} \acmArticle{20} \acmMonth{6} \acmPrice{15.00}\acmDOI{10.1145/3338695}

\begin{document}
\title{Non-Smooth Newton Methods for Deformable Multi-Body Dynamics} 

\author{Miles Macklin}
\affiliation{ \institution{NVIDIA}}
\affiliation{ \institution{University of Copenhagen}}
\email{mmacklin@nvidia.com}

\author{Kenny Erleben}
\affiliation{ \institution{University of Copenhagen}}
\email{kenny@di.ku.dk}

\author{Matthias M\"uller}
\affiliation{ \institution{NVIDIA}}
\email{matthiasm@nvidia.com}

\author{Nuttapong Chentanez}
\affiliation{ \institution{NVIDIA}}
\email{nchentanez@nvidia.com}

\author{Stefan Jeschke}
\affiliation{ \institution{NVIDIA}}
\email{sjeschke@nvidia.com}

\author{Viktor Makoviychuk}
\affiliation{ \institution{NVIDIA}}
\email{vmakoviychuk@nvidia.com}

\begin{abstract}
We present a framework for the simulation of rigid and deformable bodies in the presence of 
contact and friction. \rem{In contrast to previous methods which solve linearized models,} 
Our method is based on a non-smooth Newton iteration that solves the underlying nonlinear complementarity problems (NCPs) directly. \add{This approach allows us to support nonlinear dynamics models, including hyperelastic deformable bodies and articulated rigid mechanisms, coupled through a smooth isotropic friction model. The fixed-point nature of our method means it requires only the solution of a symmetric linear system as a building block.} We propose a new complementarity preconditioner \add{for NCP functions} that improves convergence, and we develop an efficient GPU-based solver based on the conjugate residual (CR) method that is suitable for interactive simulations. We show how to improve robustness using a new geometric stiffness approximation and evaluate our method's performance on a number of robotics simulation scenarios, including dexterous manipulation and training using reinforcement learning.
\end{abstract}



\begin{teaserfigure}
	\includegraphics[width=\columnwidth]{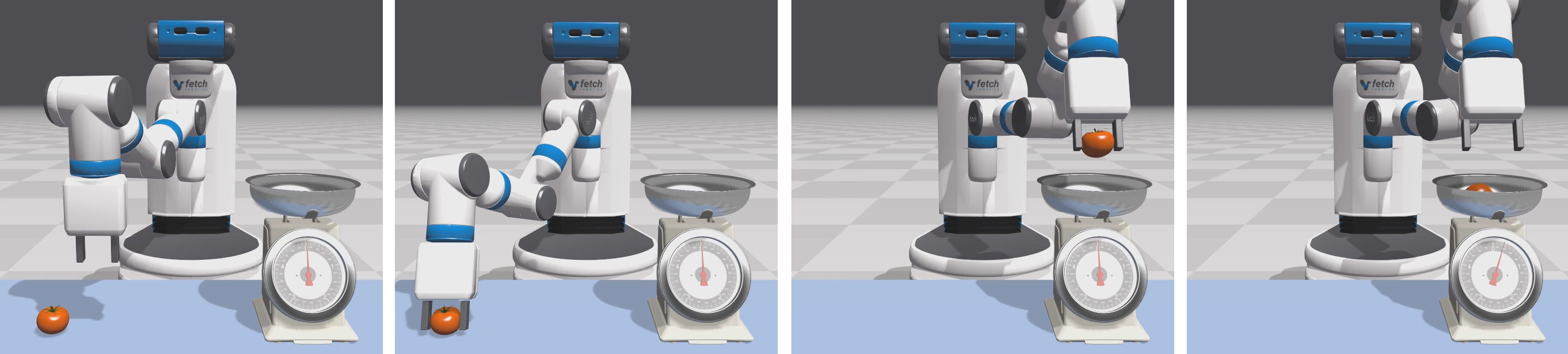}
	\caption{The Fetch robot picking up and transferring a tomato to a mechanical scale. The tomato is modeled using tetrahedral FEM, while the robot and working mechanical scale are modeled as rigid bodies connected by revolute and prismatic joints. Our method provides full two-way coupling that allows for stable grasping and force sensing on the gripper. The robot is controlled by a human operator in real-time. Model provided courtesy of Fetch Robotics, Inc.}
\label{fig:fetch_tomato}
\end{teaserfigure}	

\maketitle

\section{Introduction}

Enabling the next-generation of robots that can, for example, enter a kitchen and prepare dinner, requires new control algorithms capable of navigating complex environments and performing dexterous manipulation of real-world objects, as illustrated in Figure \ref{fig:fetch_tomato}. Machine-learning based approaches hold the promise of unlocking this capability, but require large amounts of data to work effectively \cite{levine2018learning}. For many tasks, gathering this data from the real-world may be inefficient, or impractical due to safety concerns. In contrast, simulation is relatively inexpensive, safe, and has been used to learn and transfer behaviors such as walking and jumping to real robots \cite{tan2018sim}\cite{sadeghi2017sim2real}. Extending transfer learning beyond locomotion to a wider range of behaviors requires the robust and efficient simulation of richer environments incorporating multiple physical models \cite{heess2017emergence}.

\begin{figure*}
	\begin{subfigure}{0.5\columnwidth}
		\includegraphics[width=\columnwidth,trim=300 100 300 0, clip]{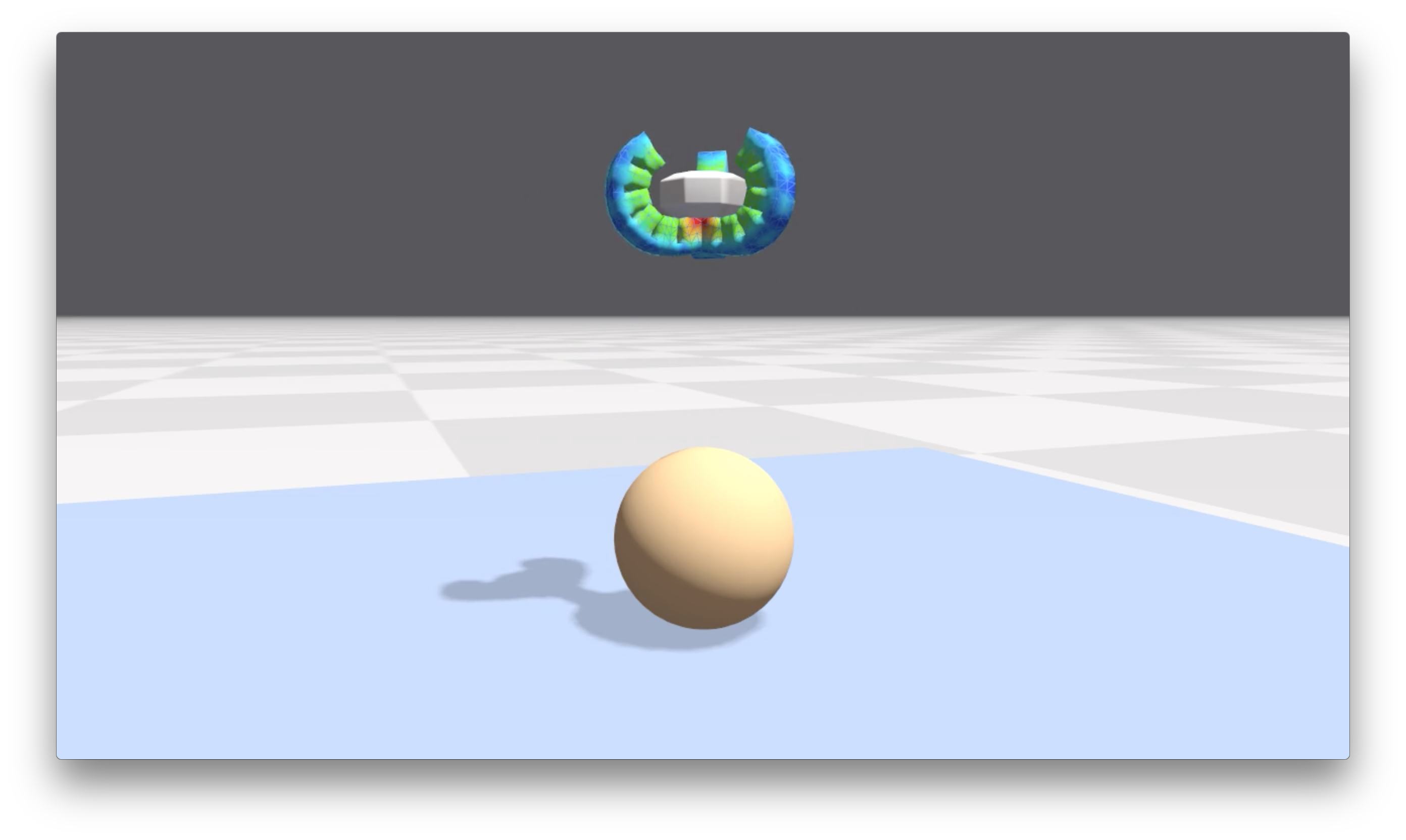}
	\end{subfigure}
	~
	\begin{subfigure}{0.5\columnwidth}
		\includegraphics[width=\columnwidth,trim=300 100 300 0, clip]{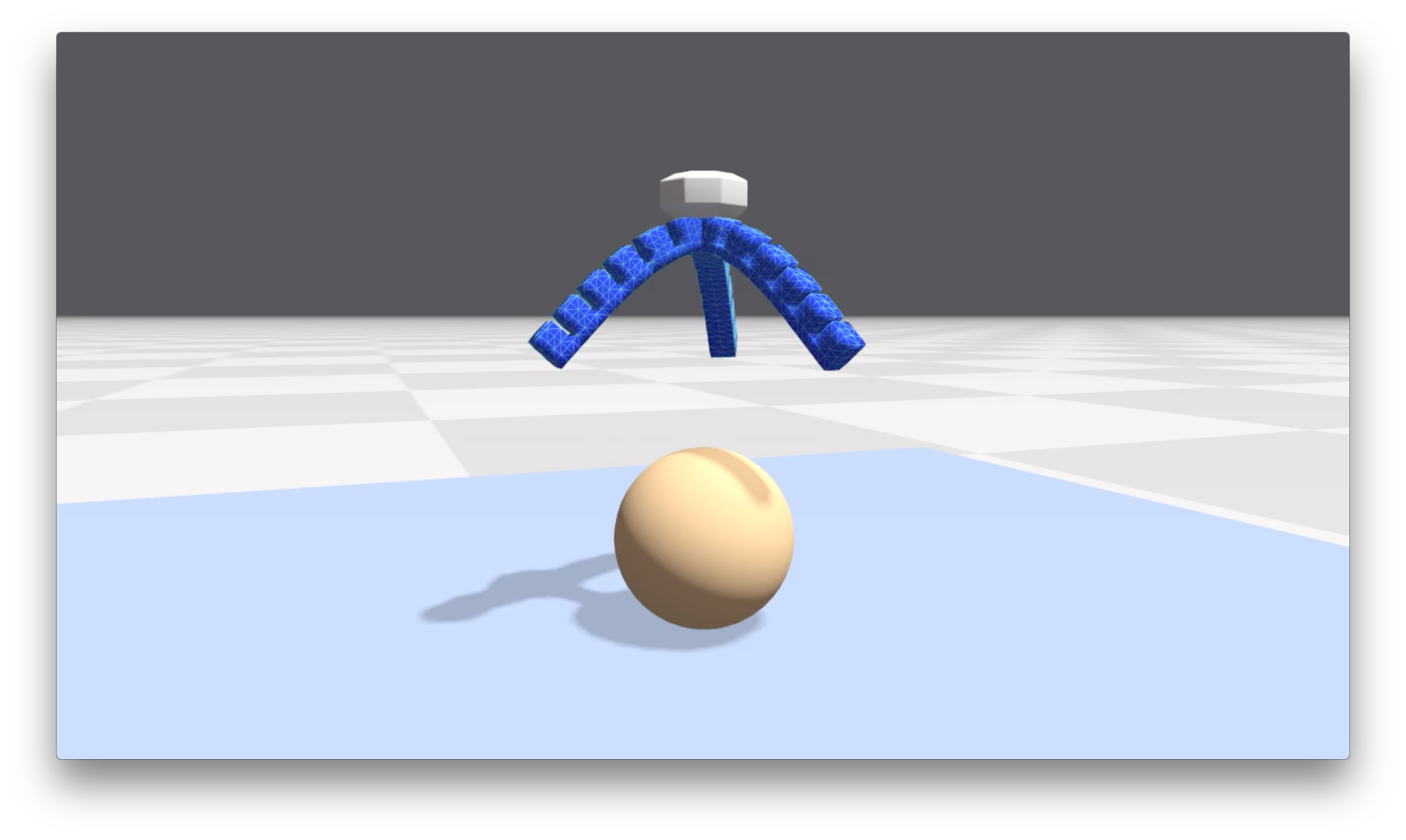}
	\end{subfigure}
	~
	\begin{subfigure}{0.5\columnwidth}
		\includegraphics[width=\columnwidth,trim=300 100 300 0, clip]{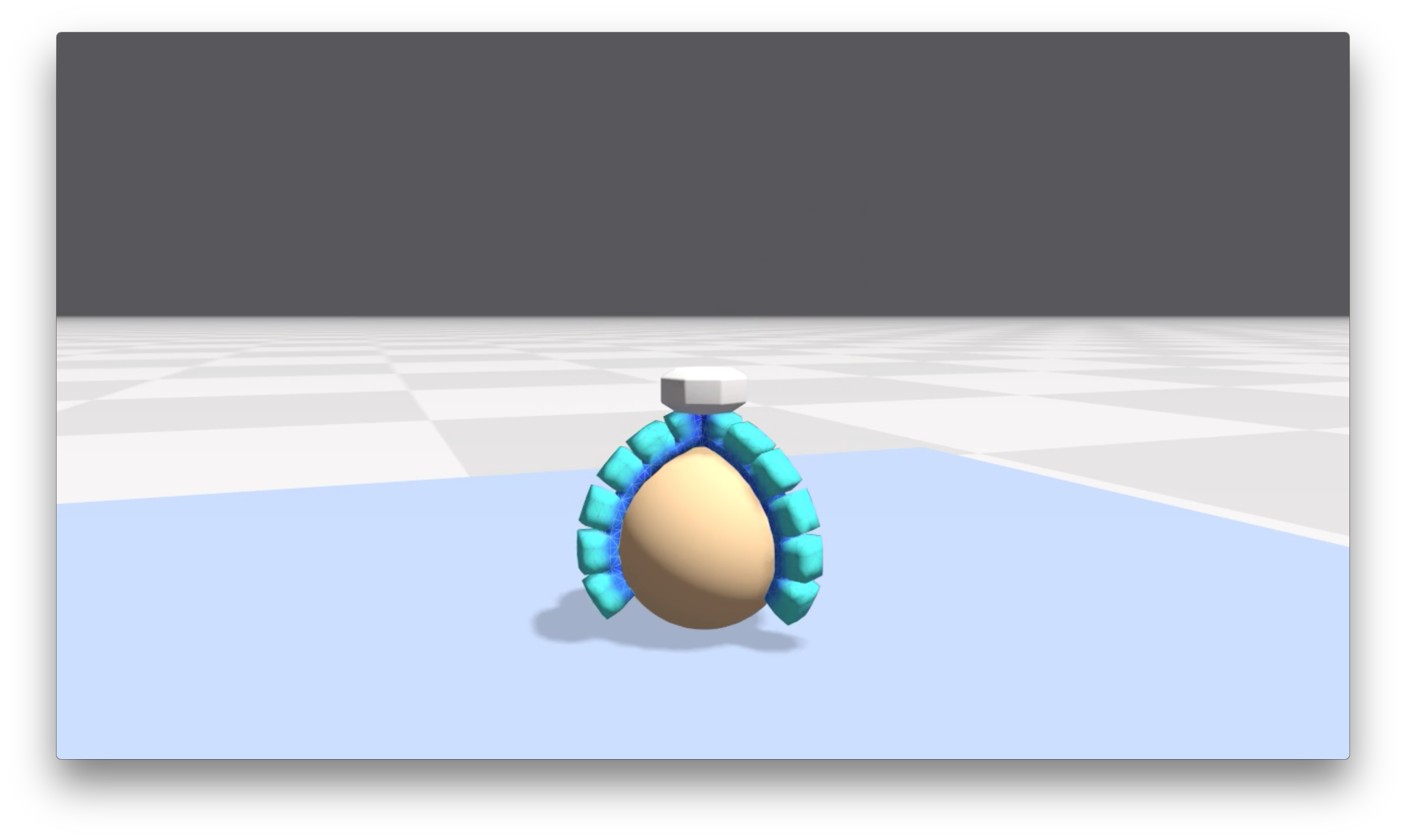}
	\end{subfigure}
	~
	\begin{subfigure}{0.5\columnwidth}
		\includegraphics[width=\columnwidth,trim=300 100 300 0, clip]{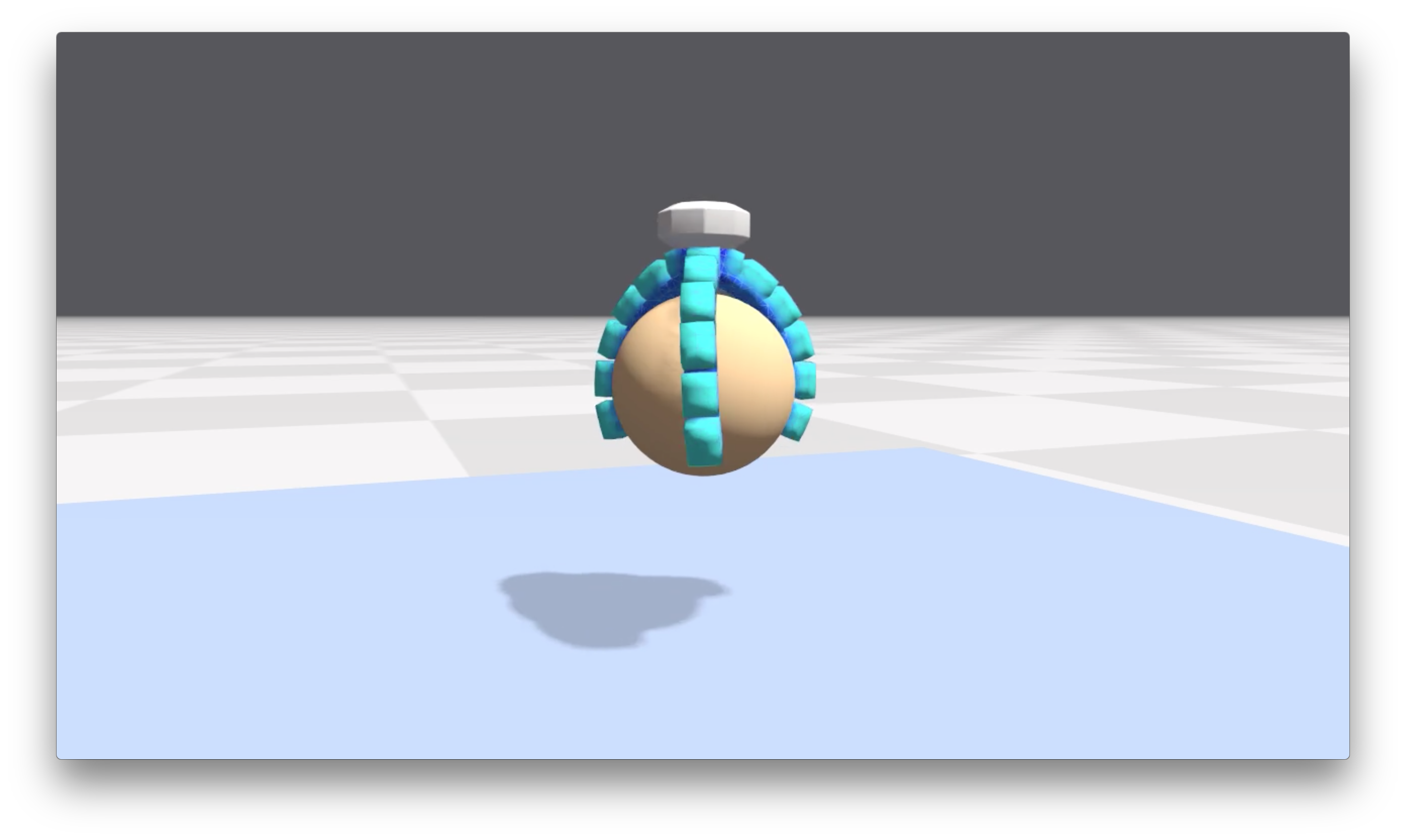}
	\end{subfigure}
																			\caption{A simulation of a soft-robotic grasping mechanism based on the pneumatic networks (PneuNet) design \cite{ilievski2011soft}. The three fingers are modeled using tetrahedral FEM, while the sphere is a rigid body. When the chambers on the back of the fingers are pressurized they cause the inextensible lower layer to curve. Our method provides robust contact coupling between the two dynamics models. The element color visualizes the volumetric strain field.}
	\label{fig:pnet}
\end{figure*}

\begin{figure}
	\begin{subfigure}{0.5\columnwidth}
		\includegraphics[width=\columnwidth]{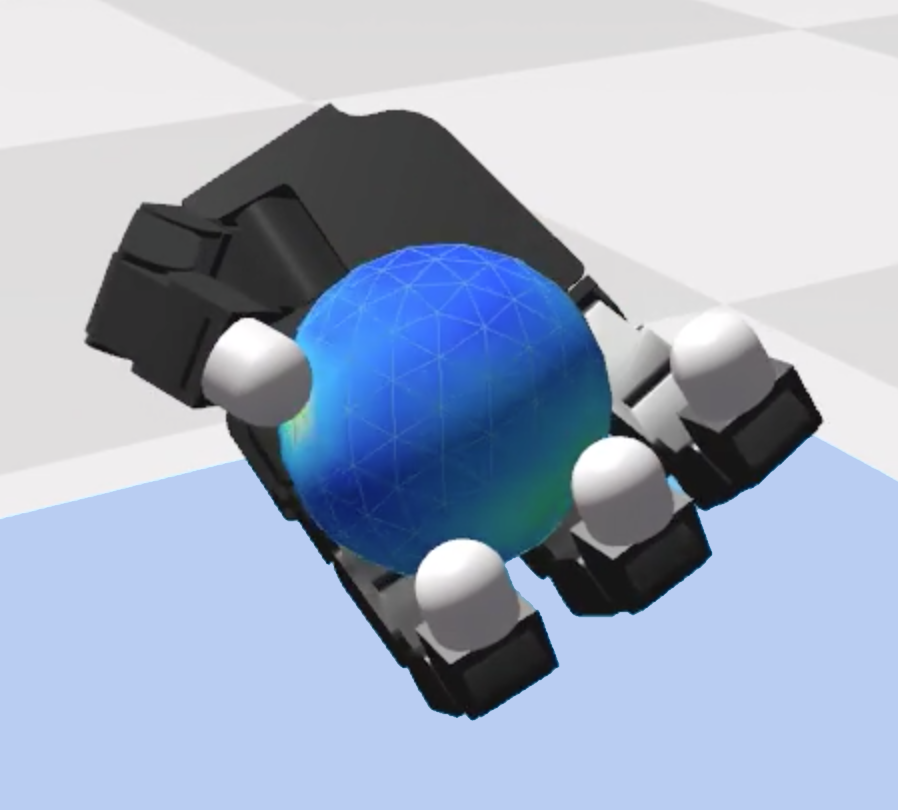}
	\end{subfigure}
	~
	\begin{subfigure}{0.5\columnwidth}
		\includegraphics[width=\columnwidth]{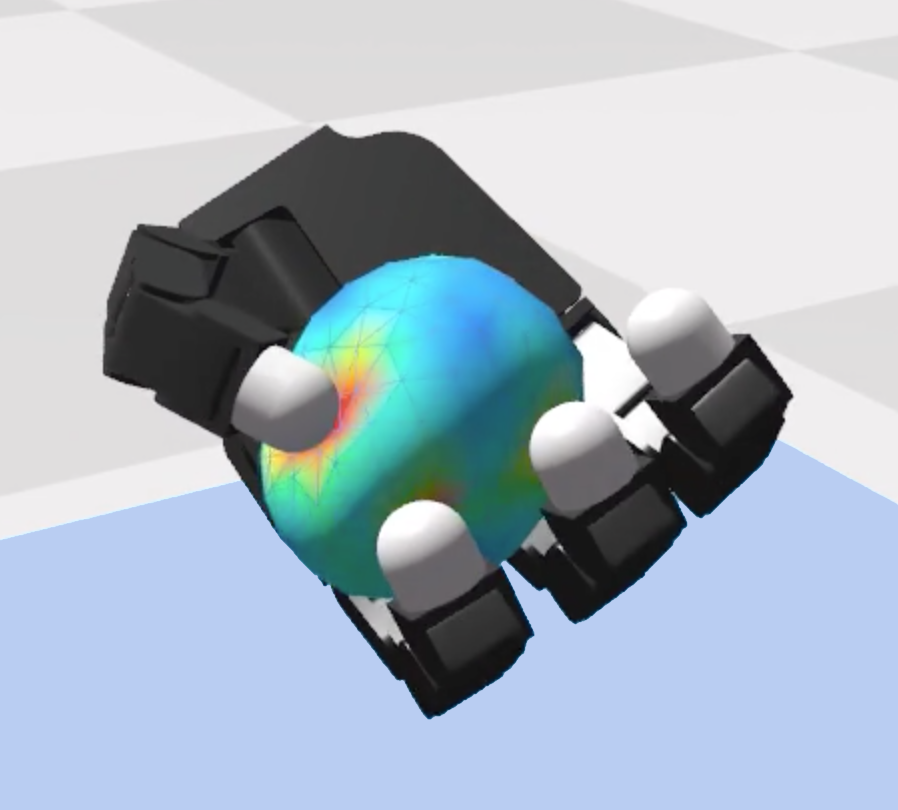}
	\end{subfigure}
	\caption{The Allegro hand squeezing a deformable ball. Our framework supports full coupling between the articulated fingers and the ball's internal dynamics. The color field visualizes volumetric strain. Model provided courtesy of SimLab Co., Ltd.}
	\label{fig:allegro_squeeze}
\end{figure}

We believe the computer graphics community is uniquely placed to address the simulation needs of robotics. One area that computer graphics has studied extensively is two-way coupled simulation of rigid and deformable objects. Such algorithms are necessary to simulate tasks involving dexterous manipulation of soft objects, or even soft robots themselves. An example of the latter is found in the PneuNet gripper \cite{ilievski2011soft} as shown in Figure \ref{fig:pnet}. This design incorporates a flexible elastic gripper, an inextensible paper layer, and a chamber that is pressurized to cause the finger to curve. An additional example is shown in Figure \ref{fig:allegro_squeeze} where a deformable ball is manipulated by a humanoid robot hand made of rigid parts. This type of multiphysics scenario is challenging for simulation because the internal forces may be large, and they must interact strongly with contact to allow robust grasping.

Methods for simulating multi-body systems in the presence of contact and friction most commonly formulate a linear complementarity problem (LCP) that is solved in one of two ways: relaxation methods such as projected Gauss-Seidel (PGS), or direct methods such as Dantzig's pivoting algorithm. Relaxation methods are popular due to their simplicity, but suffer from slow convergence for poorly conditioned problems \cite{erleben2013numerical}. In contrast, direct methods can achieve greater accuracy, but are typically serial algorithms that scale poorly with problem size. Finding methods that combine the simplicity and robustness of relaxation methods with the accuracy of direct methods remains a challenge in computer graphics and robotics. While solving LCP problems efficiently is still the subject of active research, as a model they may not capture all of the dynamics we wish to simulate. For example, hyperelastic materials have highly nonlinear forces that significantly affect behavior compared to linear models \cite{smith2018stable}. In addition, contact models themselves may be nonlinear particularly when considering compliance and deformation \cite{li2001review}.

In this work we develop a framework based on Newton's method that solves the underlying nonlinear complementarity problems (NCPs) arising in multi-body dynamics. We combine a nonlinear contact model, articulated rigid-body model, and a hyperelastic material model as a system of semi-smooth equations, and show how it can be solved efficiently. A key advantage of our Newton-based approach is that it allows the use of off-the-shelf linear solvers as the fundamental computational building block. This flexibility means we can choose to use accurate direct solvers, or take advantage of highly-optimized iterative solvers available for parallel architectures such as graphics-processing units (GPUs).  In summary, our main contributions are:
\vspace{0.0625em}
\begin{itemize}
		\item A formulation of smooth isotropic Coulomb friction in terms of non-smooth complementarity functions.
	\item A new complementarity preconditioner that significantly improves convergence for contact problems.
	\item A generalized compliance formulation that supports hyperelastic material models.
	\item A simple approximation of geometric stiffness to improve robustness without changing system dynamics.
\end{itemize}

\section{Related Work}

\subsection{Contact}

\add{The seminal work of Jean and Moreau \shortcite{jean1992unilaterality} introduced an implicit time-stepping scheme for contact problems with friction.} This work was further popularized by Stewart and Trinkle \shortcite{stewart1996implicit} who linearize the Coulomb friction cone and solve LCPs using Lemke's method in a fixed-point iteration to handle nonlinear forces. We also make use of a fixed-point iteration, but in contrast to their work we use non-smooth functions to model nonlinear friction cones. Kaufman et al. \shortcite{kaufman2008staggered} proposed a method for implicit time-stepping of contact problems by solving two separate QPs for normal and frictional forces in a staggered manner. In our work we do not stagger our system updates but solve for contact forces and friction forces in a combined system, and without using a linearized cone. In addition, rather than using QP solves our method requires only the solution of a symmetric linear system as a building block, allowing for the use of off-the-shelf linear solvers. 

Relaxation methods such as projected Gauss-Seidel are popular in computer graphics thanks to their simplicity and robustness \cite{bender2014survey}. While robust, these methods suffer from slow convergence for poorly conditioned problems, e.g.: those with high-mass ratios \cite{erleben2013numerical}. In addition, Gauss-Seidel iteration suffers from order dependence and is challenging to parallelize, while Jacobi methods require modifications to ensure convergence \cite{Tonge:2012:MSJ:2185520.2185601}. \addrev{Daviet et al. \shortcite{daviet2011hybrid} used a change of variables to restate the Coulomb friction cone into a self-dual complementarity cone problem followed by a modified Fischer-Burmeister reformulation to obtain a local non-smooth root search problem. In comparison, we work directly with the friction cone as limit surfaces as we believe this provides us with more modeling freedom, e.g.: for anisotropic or non-symmetric friction cones. Another difference is that we solve for all contacts simultaneously rather than one-by-one.}

Otaduy et al. \shortcite{otaduy2009implicit} presented an implicit time-stepping scheme for deformable objects that solves a mixed linear complementarity problem (MLCP) using a nested Gauss-Seidel relaxation over primal and dual variables. Prior work on simulating smooth friction models has used proximal-map projection operators \add{\cite{jourdan1998gauss, jean1999non, erleben2017rigid}} which work by projecting contact forces to the friction cone one contact at a time until convergence. There has been considerable work to address the slow convergence of relaxation methods, Mazhar et al. \shortcite{mazhar2015using} use a convexification of the frictional contact problem \cite{anitescu2004constraint} to obtain a cone complementarity problem (CCP) and solve it using an accelerated version of projected gradient descent. Silcowitz et al. \shortcite{silcowitz2009nonsmooth, silcowitz2010nonsmooth} developed a method for solving LCPs based on non-smooth nonlinear conjugate gradient (NNCG) \add{applied to a PGS iteration}. Francu et al. \shortcite{francu2015virtual} proposed an improved Jacobi method based on a projected conjugate residual (CR) method. \rem{Our method is related, however rather than go through an LCP we consider the underlying NCP directly. We also advocate for conjugate residual methods, but our framework is independent any specific linear solver. }
\add{We also make use of Krylov space linear solvers, however our Newton-based iteration is decoupled from the underlying linear backend, allowing the application of matrix-splitting relaxation methods, or even direct solvers}.

\add{Early work on Newton methods for contact problems used a formulation based on a generalized projection operator and an augmented Lagrangian approach to unilateral constraints \cite{alart1991mixed, curnier1988generalized}.} Newton-based approaches found in libraries such as PATH \cite{dirkse1995path} have proved successful in practice, and have been applied to smooth Coulomb friction models in fiber assemblies in computer graphics \add{\cite{bertails2011nonsmooth, daviet2011hybrid, kaufman2014adaptive}}. These approaches formulate the complementarity problem in terms of non-smooth functions, and solve them with a generalized version of Newton's method. This approach can yield quadratic convergence, although with a higher per-iteration cost than relaxation methods.

\add{Todorov \shortcite{todorov2010implicit} observed that if solving a nonlinear time-stepping problem, e.g.: due to an implicit time-discretization, it makes little sense to perform the friction cone linearization, since the smooth contact model can be treated simply as an additional set of nonlinear equations in the system}. This observation is at the heart of our method, but in contrast to their work we formulate friction in terms of arbitrary NCP-functions, and extend our framework to handle deformable bodies. Our approach is based on Newton's method, and combined with our proposed preconditioner we show that it enables handling considerably more ill-conditioned problems than relaxation methods, while naturally accommodating nonlinear friction models. For a review of numerical methods for linear complementarity problems we refer to the book by Niebe and Erleben \shortcite{niebe2015numerical}. \add{For a review of non-smooth methods applied to dynamics problems we refer to the book by Acary \& Brogliato \shortcite{acary2008numerical}.}

\subsection{Coupled Systems}

There has been considerable work in computer graphics on coupling between rigid and deformable bodies. Shinar et al. \shortcite{shinar2008two} proposed a method for the coupled simulation of rigid and deformable bodies through a multi-stage update that peforms collisions, contacts, and stabilization in separate passes. In contrast, we formulate a system update that includes elastic and contact dynamics in a single phase. Duriez \shortcite{duriez2013control} showed real-time control of soft-robots using a co-rotational FEM model and Gauss-Seidel based constraint solver. Servin et al. \shortcite{servin2006interactive} introduced a compliant version of elasticity that fits naturally inside constrained rigid body simulators. Their work was extended by Tournier et al. \shortcite{tournier2015stable} who include a geometric stiffness term to improve stability. They use temporal coherence of Lagrange multipliers to build the system Jacobian and compute a Cholesky decomposition, followed by a projected-Gauss Seidel solve for contact. \add{For smaller problems our approach is compatible with direct solvers, however we avoid the requirement of dense matrix decompositions by using a diagonal geometric stiffness approximation inspired by the work of Andrews et al. \shortcite{andrews2017geometric}, this improves stability and allows us to apply iterative methods.}

In this work we propose a generalized view of compliance, and give a recipe for constructing the compliance form of an arbitrary material model given its strain-energy density function in terms of principle stretches, strains, or other parameterization. As an example we show how to formulate the stable Neo-Hookean material proposed by Smith et al. \shortcite{smith2018stable}. Liu et al. \shortcite{liu2016towards} propose a quasi-Newton method for hyperelastic materials based on Projective Dynamics \cite{bouaziz2014projective} and model contact through stiff penalty forces. In contrast we model contact through complementarity constraints that naturally fit into existing multi-body simulations.

\begin{table}
	\centering
	\label{tab:notation}
	\caption{Glossary of terms.}
	\begin{tabular}{ c l }
		\hline  
		\textbf{Symbol} & \textbf{Meaning} \\
		\hline  \hline \\
		$\q$ & Generalized system coordinates   \\
		$\qdot$ & First time derivative of generalized coordinates \\
		$\qddot$ & Second time derivative of generalized coordinates \\
		$\tilde{\q}$ & Predicted or inertial system configuration \\
		$\v{f}$ & Generalized force function \\
		$\v{u}$ & Generalized velocity vector \\
		$\v{M}$ & System mass matrix \\
		$\v{K}$ & Geometric stiffness matrix \\
		$\v{E}$ & Compliance matrix \\				
		$\v{D}$ & Friction force basis \\
		$\v{W}$ & Friction compliance matrix\\
		$\v{G}$ & Kinematic mapping transform \\
		$\v{N}$ & Material stiffness matrix \\
		$\Psi$ & Strain energy density function\\		
		$\v{c}_b$ & Bilateral constraint vector \\
		$\v{c}_n$ & Normal contact constraint vector \\
		$\v{c}_f$ & Frictional contact constraint vector \\
		$\v{n}$ & Contact normal \\
		$d$ & Separation distance for a contact \\
		$\gv{\phi}_n$ & Normal contact NCP constraint vector \\
		$\gv{\phi}_f$ & Frictional contact NCP constraint vector \\
		$\v{J}_b$ & Jacobian of bilateral constraints \\
		$\v{J}_n$ & Jacobian of contact normal NCP constraints \\
		$\v{J}_f$ & Jacobian of contact friction NCP constraints \\
		$\gv{\lambda}_b$ & Lagrange multiplier vector for bilateral constraints \\
		$\gv{\lambda}_n$ & Lagrange multipliers for normal contact constraints \\
		$\gv{\lambda}_f$ & Lagrange multipliers for friction contact constraints\\
		$n_d$ & Number of system degrees of freedom \\
		$n_b$ & Number of bilateral constraints \\
		$n_c$ & Number of contacts \\	
		\hline  
	\end{tabular}
	\label{tab:notation}
\end{table}

\section{Background}

At the most general level, the dynamics of the systems we wish to simulate can be described through the following second order differential equation:

\begin{align}\label{eq:newtons_2nd}
\v{M}(\q)\qddot - \v{f}(\q, \qdot) = \v{0},
\end{align}

\noindent where for a system with $n_d$ degrees of freedom, $\v{M} \in \mathbb{R}^{n_d\times n_d}$ is the system mass matrix, $\v{f} \in \mathbb{R}^{n_d}$ is a generalized force vector, and  $\q \in \mathbb{R}^{n_d}$ is a vector of generalized coordinates describing the system configuration with $\qdot, \qddot$ the first and second time derivatives respectively. We refer the reader to Table \ref{tab:notation} for a list of symbols used in this paper. 

\subsection{Hard Kinematic Constraints}

We impose hard bilateral (equality) constraints on the system configuration through constraint functions of the form $C_b(\q) = 0$.  Using D'Alembert's principle \cite{lanczos1970variational} the force due to such a constraint is of the form $\v{f}_{b} = \nabla C_b(\q)^T\lambda_b$ where $\nabla C_b$ is the constraint's gradient with respect to the system coordinates, and $\lambda_b$ is a Lagrange multiplier variable used to enforce the constraint. Combining these algebraic constraints with the differential equation \eqref{eq:newtons_2nd} gives the following Differential Algebraic Equation (DAE):

\begin{align}
\v{M}(\q)\qddot - \v{f}(\q, \qdot) - \nabla\v{c}_b^T\gv{\lambda}_b &= \v{0} \\
\v{c}_b(\q) &= \v{0}. \label{eq:czero}
\end{align}
	
For a system with $n_b$ equality constraints we define the constraint vector $\v{c}_b(\q) = [C_{b,1}, C_{b,2}, \cdots, C_{b,n_b}]^T \in \mathbb{R}^{n_b}$, $\nabla\v{c}_b \in \mathbb{R}^{n_b\times n_d}$ its gradient with respect to system coordinates, and $\gv{\lambda}_b \in \mathbb{R}^{n_b}$ the vector of Lagrange multipliers. \add{In general bilateral constraints may depend on velocity and time, we assume scleronomic constraints for the remainder of the paper to simplify the exposition}.

\subsection{Soft Kinematic Constraints}\label{sec:softconstraints}

In addition to hard constraints, we make extensive use of the compliance form of elasticity \cite{servin2006interactive}. This may be derived by considering a quadratic energy potential defined in terms of a constraint function $C_b(\q)$ and stiffness parameter $k > 0$

\begin{align}
U(\q) = \frac{k}{2}C_b(\q)^2.
\end{align}

The force due to this potential is then given by 

\begin{align}
\v{f}_b = -\nabla U^T = -k\nabla C_b^T C_b(\q) = \nabla C_b^T\lambda_b.
\end{align}

\rem{where the variable $\lambda_b$ is defined as $\lambda_b = -kC_b(\q)$. This may be re-arranged to give a constraint equation of the form,}

\add{Here we have introduced the variable $\lambda_b$, defined as}

\begin{align}
\add{\lambda_b = -kC_b(\q).}
\end{align}

\add{We can move all terms to one side to write this as a constraint equation,}

\begin{align}
C_b(\q) + e\lambda_b = 0,
\end{align}

\noindent where $e = k^{-1}$ is the compliance, or inverse stiffness coefficient. We can incorporate this into our equations of motion by defining the diagonal matrix $\v{E} = \text{diag}(e_1, e_2, \cdots, e_{n_b}) \in \mathbb{R}^{n_b\times n_b}$, and updating \eqref{eq:czero} as follows,

\begin{align}
\v{c}_b(\q) + \v{E}\gv{\lambda}_b = \v{0}.
\end{align}

This form is mathematically equivalent to including quadratic energy potentials defined in terms of a stiffness. The benefit of the compliance form is that it remains numerically stable even as $k\to\infty$ \cite{tournier2015stable}. We describe how to extend this model to handle general energy potentials in Section \ref{sec:deformables}.

\begin{figure}[h]
	\includegraphics[width=2.8in]{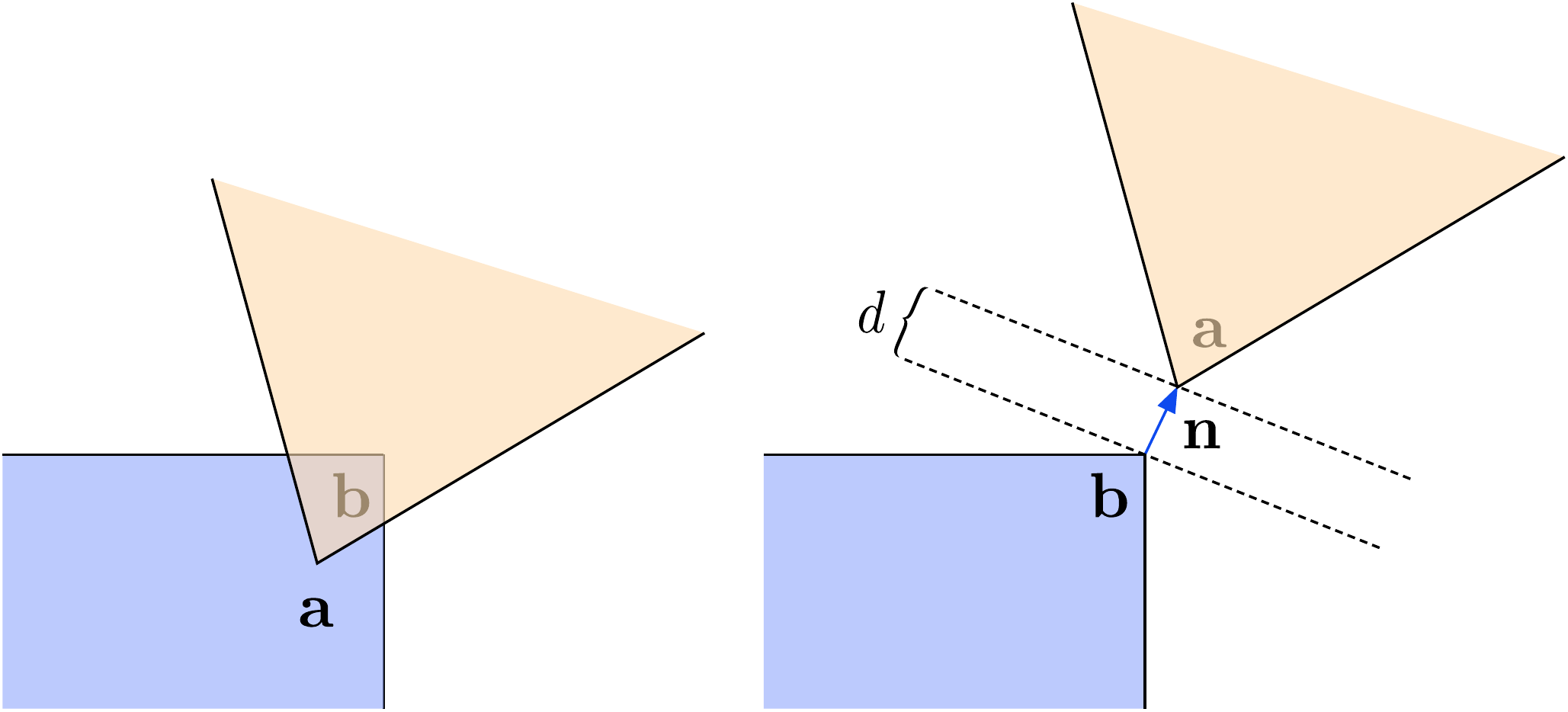}
	\caption{We model contact as a constraint on the distance between two points $\v{a}$ and $\v{b}$ as measured along a fixed normal direction $\v{n}$ being greater than some minimum $d$.}
	\label{fig:contact}
\end{figure}

\subsection{Contact}

We treat contacts as inelastic and prevent interpenetration between bodies through inequality constraints of the form

\begin{align}
C_n(\q) = \v{n}^T\left[\v{a}(\q) - \v{b}(\q)\right] - d \ge 0.
\end{align}

Here $\v{n} \in \mathbb{R}^3$ is the contact plane normal. \add{We use the normalized direction vector between closest points of triangle-mesh features as the normal vector}. \add{The points $\v{a}$ and $\v{b} \in \mathbb{R}^3$ may be points defined in terms of a rigid body frame, or particle positions in the case of a deformable body}. The constant $d$ is a thickness parameter that represents the distance along the contact normal we wish to maintain. Non-zero values of $d$ can be used to model shape thickness, as illustrated in Figure \ref{fig:contact}. The normal force for a contact is given by $\v{f}_n = \nabla C_n^T\lambda_n$, with the additional Signorini-Fichera complementarity condition 

\begin{align}
0 \le C_n(\q) \perp \lambda_n \ge 0,
\end{align}

\noindent which ensures contact forces only act to separate objects \cite{stewart2000rigid}. We treat the contact normal $\v{n}$ as fixed in world-space. For a system with $n_c$ contacts, we define the vector of contact constraints as $\v{c}_n = [C_{n,1},\cdots, C_{n,n_c}] \in \mathbb{R}^{n_c}$, their gradient $\nabla\v{c}_n \in \mathbb{R}^{n_c\times n_d}$, and the associated Lagrange multipliers as $\gv{\lambda}_n \in \mathbb{R}^{n_c}$. \add{In our implementation contact constraints are created when body features come within some fixed distance of each other. This approach works well for reasonably small time-steps, but can lead to over-constrained configurations. More sophisticated non-interpenetration constraints can be formulated to avoid this problem \cite{williams2017geometrically}.}

\begin{figure}[h]
	\includegraphics[width=3.2in]{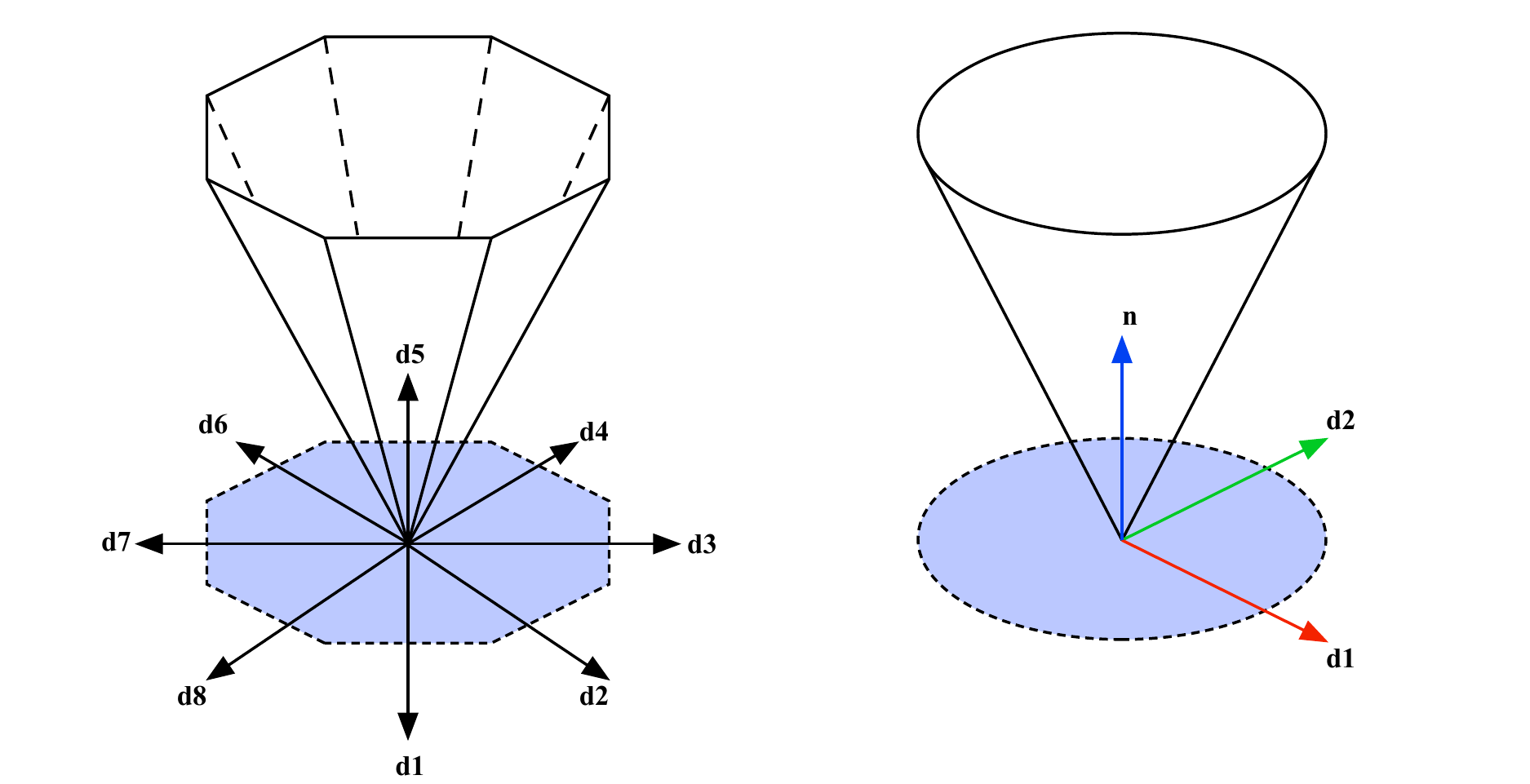}
	\caption{\textbf{Left}: A Coulomb friction cone approximated by linearizing into 8 directions leads to a larger LCP problem and introduces bias in some directions. \textbf{Right}: We project frictional forces directly to the smooth friction cone, and require only 2 tangential direction vectors.}
	\label{fig:friction_cone}
\end{figure}

\subsection{Friction}

We derive our friction model from a principle of maximal dissipation that requires the frictional forces remove the maximum amount of energy from the system while having their magnitude bounded by the normal force \cite{stewart2000rigid}. \add{For each contact we define a two-dimensional basis $\v{D} = \v{\Gamma}(\v{q})[\v{d}_1 \v{d}_2] \in \mathbb{R}^{n_d\times 2}$. The contact basis vectors $\v{d}_1, \v{d}_2 \in \mathbb{R}^{3\times 1}$ are lifted from spatial to generalized coordinates by the transform $\v{\Gamma} \in \mathbb{R}^{n_d\times 3}$ using the notation of Kaufman et al. \shortcite{kaufman2014adaptive}.} The generalized frictional force for a single contact is then $\v{f}_f = \v{D}\gv{\lambda}_f$, where $\gv{\lambda}_f \in \mathbb{R}^2$ is the solution to the following minimization

\begin{equation}
\begin{aligned}\label{eq:friction_min}
\underset{\gv{\lambda}_f}{\text{argmin}} & \hspace{1em} \qdot^T\v{D}\gv{\lambda}_f \\ 
\text{subject to} & \hspace{1em}|\gv{\lambda}_f| \le \mu\lambda_n.
\end{aligned}
\end{equation}
\\
Here $\mu$ is the coefficient of friction, and $\lambda_n$ is the Lagrange multiplier for the normal force at this contact which for the moment we assume is known. This minimization defines an admissible cone that the total contact force must lie in, as illustrated in Figure \ref{fig:friction_cone}. The Lagrangian associated with this minimization is

\begin{align}
\mathcal{L}(\gv{\lambda}_f, \lambda_n) = \qdot^T\v{D}\gv{\lambda}_f + \lambda_s(|\gv{\lambda}_f| - \mu\lambda_n).
\end{align}

\noindent where $\lambda_s$ is a slack variable used to enforce the Coulomb constraint that the friction force magnitude is bounded by the \add{$\mu$ times the normal force magnitude}. When $\mu\lambda_n > 0$ the problem satisfies Slater's condition \cite{boyd2004convex} and we can use the first-order Karush-Kuhn-Tucker (KKT) conditions for \eqref{eq:friction_min} given by

\begin{align}
\v{D}^T\qdot + \lambda_s\pdv{|\gv{\lambda}_f|}{\gv{\lambda}_f} = \v{0} \label{eq:kkt1} \\
0 \le \lambda_s ~~ \perp ~~ \mu\lambda_n - |{\gv{\lambda}_f}| \ge 0 \label{eq:kkt2}.
\end{align}

Equation \eqref{eq:kkt1} requires that the frictional force act in a direction opposite to any relative tangential velocity. Equation \eqref{eq:kkt2} is a complementarity condition that describes the sticking and slipping behavior characteristic of dry friction. In the case that $\mu\lambda_n = 0$ \rem{the optimal point may not satisfy these KKT conditions} \add{Slater's condition is not satisfied}, however in this case the only feasible point is $\gv{\lambda}_f = \v{0}$, which we handle explicitly. We now make a transformation to remove the slack variable $\lambda_s$. To do this we observe that the derivative of a vector's length is the vector normalized, so we have $\pdv{|\gv{\lambda}_f|}{\gv{\lambda}_f} = \gv{\lambda}_f/|\gv{\lambda}_f|$, which by definition is a unit-vector. We can use this fact to express the optimality conditions in terms of $\qdot$ and $\gv{\lambda}_f$ only:

\begin{align}
\v{D}^T\qdot + \frac{|\v{D}^T\qdot|}{|\gv{\lambda}_f|}\gv{\lambda}_f = \v{0} \label{eq:friction1} \\
0 \le |\v{D}^T\qdot| ~~ \perp ~~ \mu\lambda_n - |\gv{\lambda}_f| \ge 0 \label{eq:friction2}.
\end{align}

Finally we combine the frictional basis vectors and multipliers for all contacts in the system as a single matrix $\nabla\v{c}_f = [\v{D}_1, \cdots, \v{D}_{n_c}]^T \in \mathbb{R}^{2n_c\times n_d}$ and vector $\gv{\lambda}_f = [\gv{\lambda}_{f,1},\cdots, \gv{\lambda}_{f, n_c}]^T \in \mathbb{R}^{2n_c}$.

\subsection{Governing Equations}\label{sec:ceom}

Assembling the above components, our continuous equations of motion are given by the following nonlinear system of equations:

\begin{align*} 
\v{M}(\q)\qddot - \v{f}(\q, \qdot) - \nabla\v{c}_b^T(\q)\gv{\lambda}_b - \nabla\v{c}_n^T(\q)\gv{\lambda}_n - \nabla\v{c}^T_f(\q)\gv{\lambda}_f = \v{0} &  \\ 
\v{c}_b(\q) + \v{E}\gv{\lambda}_b = \v{0} & \\
\v{0} \le \v{c}_n(\q) \perp \gv{\lambda}_n \ge \v{0} & \\
i \in \mathcal{A}, \quad \v{D}_i^T\qdot + \frac{|\v{D}_i^T\qdot|}{|\gv{\lambda}_{f,i}|}\gv{\lambda}_{f,i} = \v{0} & \\
i \in \mathcal{A}, \quad 0 \le |\v{D}_i^T\qdot| ~~ \perp ~~ \mu_i\lambda_{n,i} - |\gv{\lambda}_{f,i}| \ge 0 \\
i \in \mathcal{I}, \quad \gv{\lambda}_{f,i} = \v{0} 
\end{align*}

\noindent where $\mathcal{A} = \{i \in (1,\cdots, n_c) \mid \mu_i\lambda_{n,i} > 0\}$ is the set of all contact indices where the normal contact force is active, and $\mathcal{I} = \{i \in (1,\cdots, n_c) \mid \mu_i\lambda_{n,i} \le 0\}$ is its complement. This combination of a differential equation with equality and complementarity conditions is known as Differential Variational Inequality (DVI) \cite{stewart2000rigid}. The coupling between normal and frictional complementarity problems makes the problem non-convex, and in the case of implicit time-integration leads to an NP-hard optimization problem \cite{kaufman2008staggered}.  In the next section we develop a practical method to solve this problem.

\section{Nonlinear Complementarity}

One successful approach \cite{ferris2000complementarity} to solving nonlinear complementarity problems is to reformulate the problem in terms of a \textit{NCP-function} whose roots satisfy the original complementarity conditions, i.e.: functions where the following equivalence holds

\begin{align}\label{eq:ncp}
\phi(a,b) = 0 \iff 0 \le a \perp b \ge 0.
\end{align}

Combined with an appropriate time-discretization, such a NCP-function turns our DVI problem into a root finding one. In general the functions $\phi$ are non-smooth, but allow us to apply a wider range of numerical methods \cite{munson.facchinei.ea:semismooth}.

\subsection{Minimum-Map Formulation}

The first such function we consider is the \textit{minimum-map} defined as

\begin{align}
\phi_\text{min}(a, b) = \text{min}(a, b) = 0.
\end{align}

The equivalence of this function to the original NCP can be verified by examining the values associated with each conditional case \cite{cottle2008linear}. We now show how this reformulation applies to our unilateral contact constraints. Recall that the complementarity condition associated with a contact constraint $C_{n}(\q)$ and its associated Lagrange multiplier $\lambda_{n}$ is

\begin{align}
0 \le C_{n}(\q) \perp \lambda_{n} \ge 0.
\end{align}

We can write this in the equivalent minimum-map form as

\begin{align}
\phi_{n}(\q, \lambda_{n}) = \text{min}(C_{n}(\q), \lambda_{n}) = 0,
\end{align}

\noindent which has the following derivatives,

\begin{align}
\pdv{\phi_{n}}{\q} &= \begin{cases}
\nabla C_{n}(\q), & C_{n}(\q) \le \lambda_{n} \\
\v{0}, & \text{otherwise} \end{cases} \\
\pdv{\phi_{n}}{\lambda_{n}} &= \begin{cases}
0, &  C_{n}(\q) \le \lambda_{n} \\
1, & \text{otherwise}.\end{cases} \label{eq:grad_phi_lambda}
\end{align}

From these cases we can see that the minimum-map gives rise to an active-set style method where a contact is considered active if $C_{n}(\q) \le \lambda_{n}$. Active contacts are treated as equality constraints, while for inactive contacts the minimum-map enforces that the constraint's Lagrange multiplier is zero.

\subsection{Fischer-Burmeister Formulation}

An alternative NCP-function is given by Fischer-Burmeister \shortcite{fischer1992special}, who observe the roots of the following equation satisfy complementarity:

\begin{align}
\phi_\text{FB}(a, b) = a + b - \sqrt{a^2 + b^2} = 0.
\end{align}

The Fischer-Burmeister function is interesting because, unlike the minimum-map, it is smooth everywhere apart from the point $(a,b) = (0,0)$. For $(a,b) \ne (0,0)$ the partial derivatives of the Fisher-Burmeister function are given by:

\begin{align}
\alpha(a,b) = \pdv{\phi_{FB}}{a} &= 1 - \frac{a}{\sqrt{a^2 + b^2}} \\
\beta(a,b) = \pdv{\phi_{FB}}{b} &= 1 - \frac{b}{\sqrt{a^2 + b^2}}.
\end{align}

At the point $(a,b) = (0,0)$ the derivative is set-valued (see Figure~\ref{fig:subgradient}). For Newton methods it suffices to choose any value from this subgradient. Erleben \shortcite{erleben2013numerical} compared how the choice of derivative at the non-smooth point affects convergence for LCP problems and found no overall best strategy. Thus, for simplicity we make the arbitrary choice of 

\begin{align}
\alpha(0,0) &= 0 \\
\beta(0,0) &= 1.
\end{align}

For a contact constraint $C_{n}$, with Lagrange multiplier $\lambda_{n}$ we may then define our contact function alternatively as,

\begin{align}
\phi_{n}(\q, \lambda_{n}) = \phi_\text{FB}(C_{n}(\q), \lambda_{n}) = 0,
\end{align}

\noindent with derivatives given by

\begin{align}
\pdv{\phi_{n}}{\q} &= \alpha(C_{n}, \lambda_{n})\nabla C_{n} \label{eq:fb_grad1}\\
\pdv{\phi_{n}}{\lambda_{n}} &= \beta(C_{n}, \lambda_{n}).
\end{align}

\begin{figure}
	\begin{subfigure}{0.5\columnwidth}
		\includegraphics[width=\columnwidth]{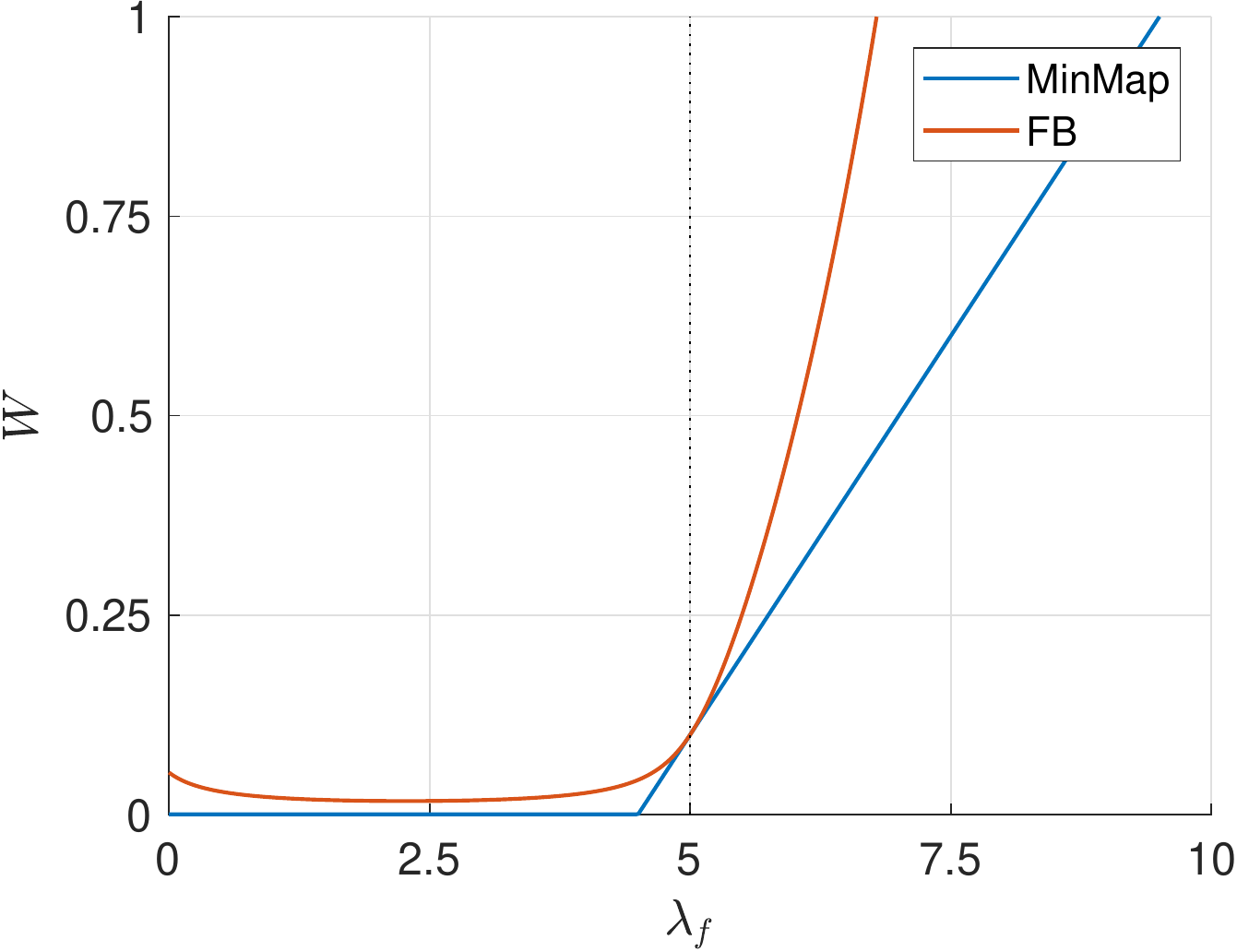}
	\end{subfigure}
	~
	\begin{subfigure}{0.5\columnwidth}
		\includegraphics[width=\columnwidth]{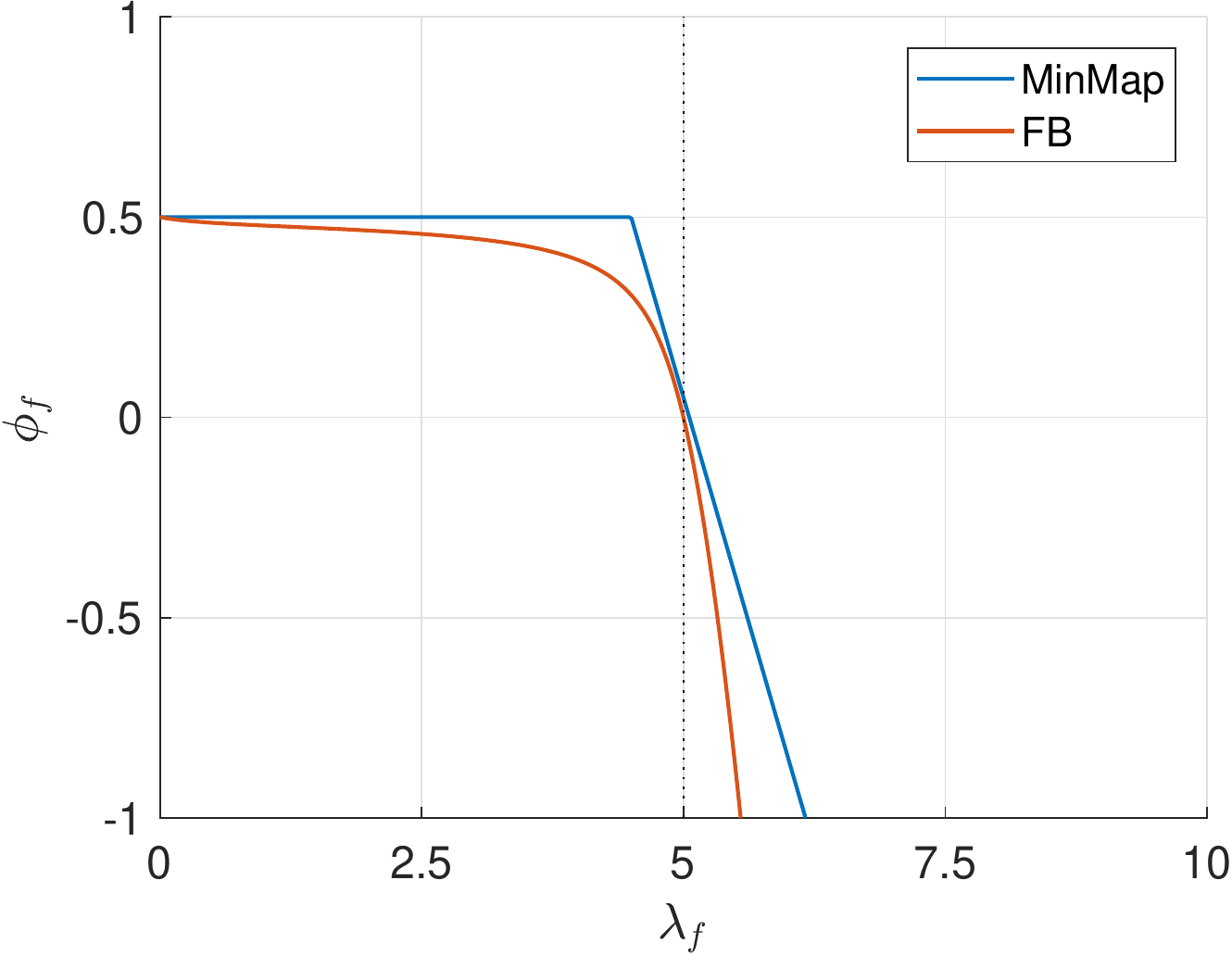}
	\end{subfigure}
	\caption{\textbf{Left:} We plot the value of our frictional compliance term $W$ for a 1-dimensional particle sliding on a plane with velocity $u_0=0.5\text{m/s}, \lambda_n=10\text{N}, \mu=0.5$. The dashed line represents the friction cone limit at $\lambda_f=5\text{N}$, after this point $W$ acts to strongly penalize the Lagrange multiplier. \textbf{Right:} The frictional error function $\phi_f$ for the same scenario. We see that both the Fischer-Burmeister and Minimum-Map functions are zero at the cone limit which indicates sliding.}
	\label{fig:friction_functions}
\end{figure}

\begin{figure*}
	\begin{subfigure}{0.5\columnwidth}
	\includegraphics[width=\columnwidth,trim={140 10 50 70}, clip]{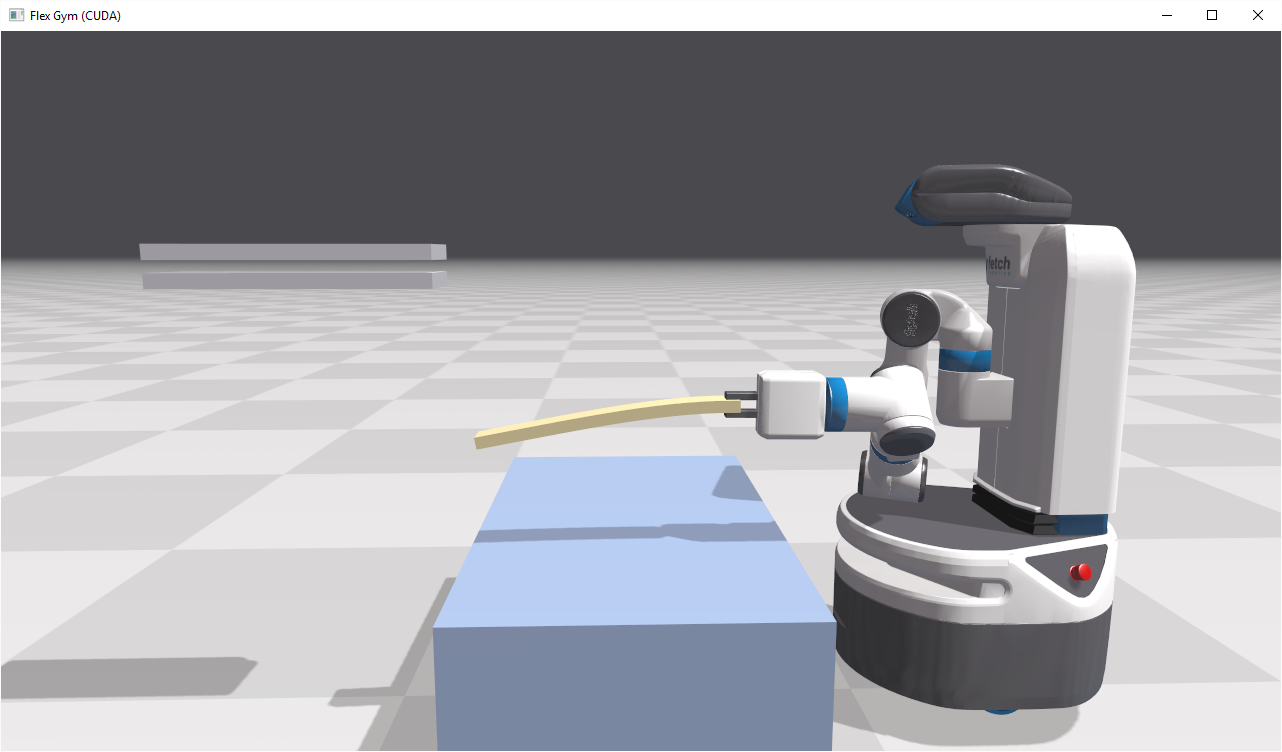}
	\end{subfigure}
	~
	\begin{subfigure}{0.5\columnwidth}
		\includegraphics[width=\columnwidth,trim={140 10 50 70}, clip]{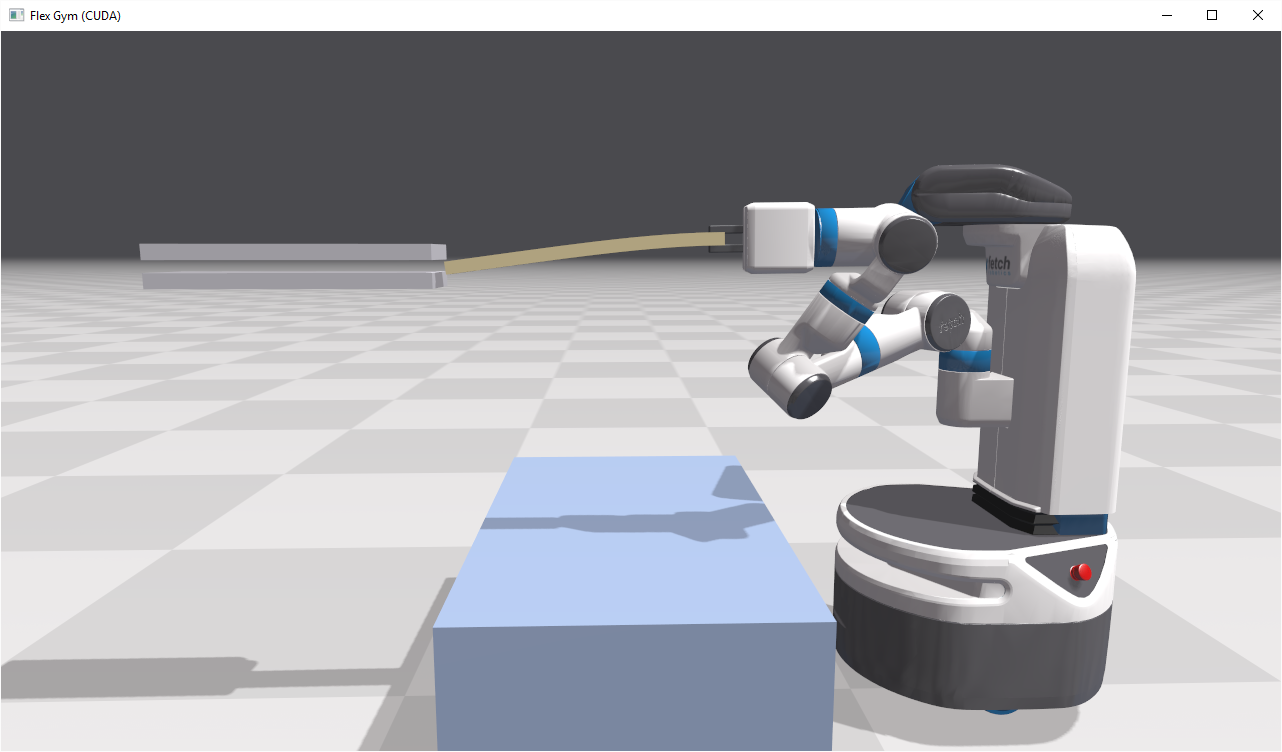}
	\end{subfigure}
	~
	\begin{subfigure}{0.5\columnwidth}
		\includegraphics[width=\columnwidth,trim={140 10 50 70}, clip]{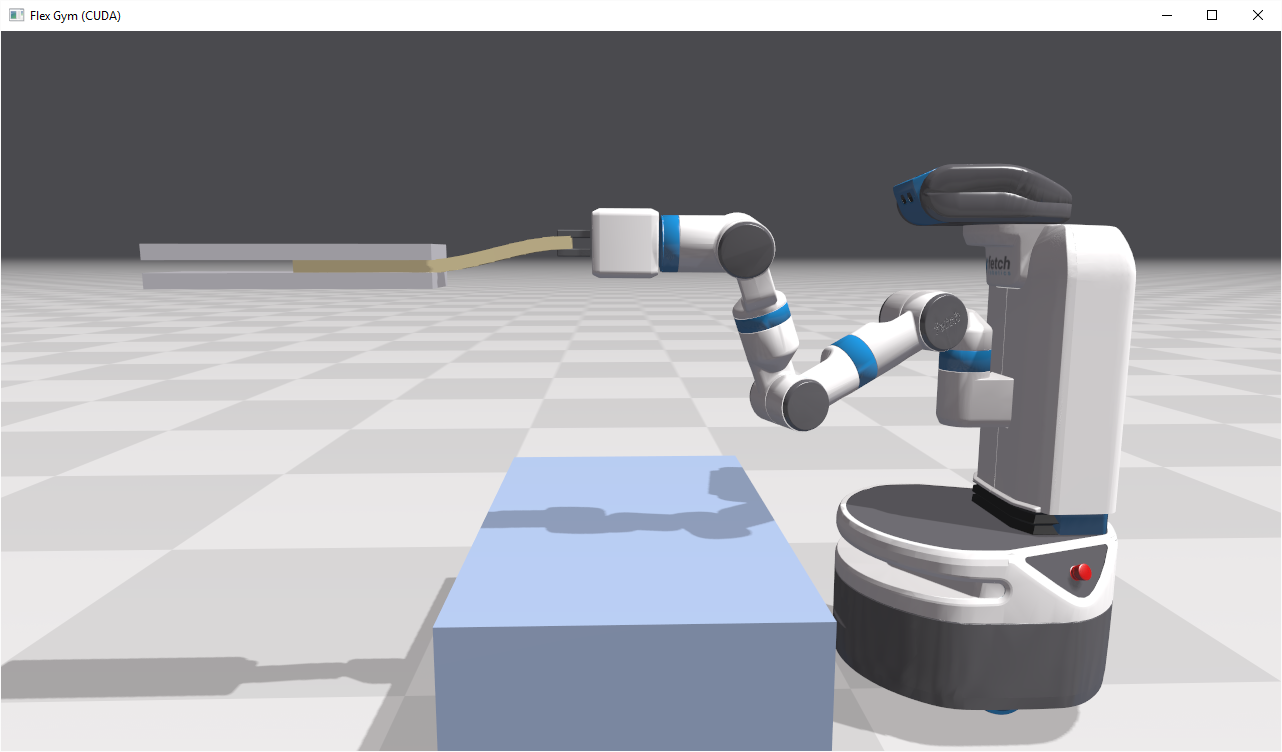}
	\end{subfigure}
	~
	\begin{subfigure}{0.5\columnwidth}
		\includegraphics[width=\columnwidth,trim={140 10 50 70}, clip]{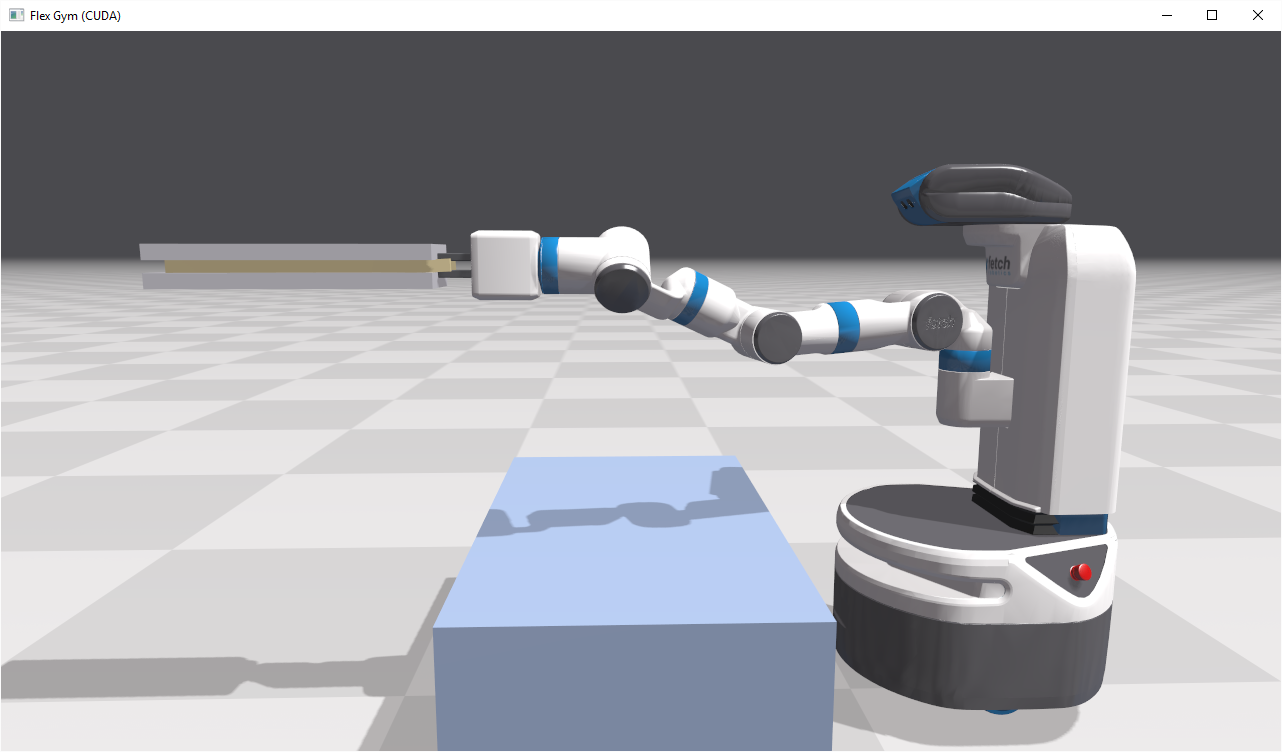}
	\end{subfigure}

	\caption{The Fetch robot performing a flexible beam insertion task. The beam is modeled as 16 rigid bodies connected by joints with a bending stiffness of 250 N$\cdot$m. Relaxation methods such as Jacobi struggle to achieve sufficient stiffness on the joints, while direct methods struggle with large contacting systems near the end of simulation. Our iterative method based on PCR achieves high stiffness and robust behavior in contact.}
	\label{fig:fetch_beam}
\end{figure*}

\subsection{Frictional Constraints}

We now formulate our friction model in terms of non-smooth functions. Our first step is to rewrite the optimality conditions \eqref{eq:friction1}-\eqref{eq:friction2} in terms of an NCP-function. For a single contact with $\mu\lambda_n > 0$ the following must hold:

\begin{align}
\v{D}^T\qdot + \frac{|\v{D}^T\qdot|}{|\gv{\lambda}_{f}|}\gv{\lambda}_{f} &= \v{0} \label{eq:friction_slip} \\
\psi_f(|\v{D}^T\qdot|, \mu\lambda_{n} - |\gv{\lambda}_{f}|) &= 0, \label{eq:friction_complementarity}
\end{align}

where $\psi_f$ may be either the minimum-map or Fischer-Burmeister function. \add{We now introduce a new quantity with the goal of linearizing the relationship between $\v{D}^T\qdot$ and $\gv{\lambda}_f$ such that upon convergence of a fixed-point iteration the initial complementarity problem is solved. We do this by defining the following fixed-point iteration based on the relationship between $|\v{D}^T\qdot|$ and $|\gv{\lambda}_f|$ given by $\psi_f$,}

\add{\begin{align}
|\v{D}^T\qdot|^{n+1} &\leftarrow |\v{D}^T\qdot|^{n} - \psi_f^{n} \\
|\gv{\lambda}_{f}|^{n+1} &\leftarrow |\gv{\lambda}_{f}|^{n} + \psi_f^{n}.
\end{align}}

\add{By construction, a fixed-point of this iteration satisfies the original complementarity condition. We use these expressions as replacements for $\frac{|\v{D}^T\qdot|}{|\gv{\lambda}_{f}|}$ in \eqref{eq:friction_slip}, by defining the following coefficient,}

\begin{align}\label{eq:friction_w}
W \equiv \frac{|\v{D}^T\qdot| - \psi_f(\qdot, \gv{\lambda}_f)}{|\gv{\lambda}_f| + \psi_f(\qdot, \gv{\lambda}_f)}.
\end{align}

\add{Inserting this into \eqref{eq:friction_slip} allows us to write it in the compact form,}

\begin{align}\add{\label{eq:friction_constraint}
	\gv{\phi}_{f}(\qdot, \gv{\lambda}_f) = \v{D}^T\qdot + W\gv{\lambda}_{f}}.
\end{align}

\add{Here $W$ can be interpreted as acting as a compliance term that works to project the frictional force back onto the friction cone as illustrated in Figure \ref{fig:friction_functions}. The derivatives of $\gv{\phi}_f$ are then given by,}

\add{\begin{align}
\pdv{\gv{\phi}_{f}}{\qdot} = \v{D},  \hspace{1.0em} \pdv{\gv{\phi}_{f}}{\gv{\lambda}_{f}} = W\v{1}_{2\times 2},
\end{align}}

\noindent \add{where we have treated $W$ as a constant. Replacing the complementarity condition by a fixed-point iteration ensures that we obtain a symmetric system of equations in the following section. In Appendix \ref{app:phif} we derive the exact form of $W$ for both minimum-map and Fischer-Burmeister functions.}

\add{We note that conical equivalents of the Fischer-Burmeister function exist and have been used to model smooth isotropic friction \cite{fukushima2002smoothing, daviet2011hybrid}. One advantage of our method being based on a fixed-point iteration is that it can be extended to arbitrary friction surfaces for e.g.: anisotropic or even non-symmetric friction cones.}

\section{Implicit Time-Stepping}

The continuous equations of motion are expressed purely in terms of our generalized coordinates $\q$, their derivatives, and the Lagrange multipliers. Although it is possible to discretize and solve these equations for $\v{q}$ directly, it is convenient to introduce an additional re-parameterization in terms of a generalized velocity vector $\v{u}$:

\begin{align}
\dot{\q} &= \v{G}(\q)\v{u} \label{eq:qdot} \\
\ddot{\q} &= \dot{\v{G}}(\q)\v{u} + \v{G}(\q)\dot{\v{u}}.
\end{align}

Here $\v{G}$ is referred to as the \textit{kinematic map}, and its components depend on the choice of system coordinates. For simple particles $\v{G}$ is an identity transform, however in Section \ref{sec:rigids} we discuss the mapping of a rigid body's angular velocity to the corresponding quaternion time-derivative. We use the the kinematic map $\v{G}$ to obtain the following mass matrix,

\begin{align}
\tilde{\v{M}} = \v{G}^T\v{M}\v{G},
\end{align}

\noindent and define the bilateral constraint Jacobians with respect to the generalized velocity $\u$ as follows:

\begin{align}
\v{J}_b = \nabla\v{c}_b\v{G}.
\end{align}

We group the normal and friction NCP-functions for all contacts into two vectors,

\begin{align}
\gv{\phi}_n &= [\phi_{n,1},\cdots,\phi_{n,nc}]^T\\
\gv{\phi}_f &= [\gv{\phi}_{f,1},\cdots,\gv{\phi}_{f,nc}]^T,
\end{align}

\noindent and likewise define their Jacobians with respect to the generalized velocity as

\begin{align}
\v{J}_n = \pdv{\gv{\phi}_n}{\q}\v{G}, \hspace{1.0em} \v{J}_f = \pdv{\gv{\phi}_f}{\qdot}\v{G}.
\end{align}

Using a first-order backwards time-discretization of $\qdot$ gives the following update for the system's generalized coordinates in terms of generalized velocities over a time-step of length $h$,

\begin{align}
\q^+ = \q^{-} + \dt\v{G}(\v{q}^{+})\v{u}^+.
\end{align}

The superscripts $+, -$ represent a variable's value at the end and beginning of the time-step respectively. Discretizing our continuous equations of motion from the previous section gives the implicit time-stepping equations,

\begin{align}
\tilde{\v{M}}\left(\frac{\v{u}^+ - \tilde{\v{u}}}{\dt}\right) - \v{J}^T_b(\v{q}^+)\gv{\lambda}_b^+ - \v{J}^T_n(\v{q}^+)\gv{\lambda}_n^+ - \v{J}^T_f(\v{q}^+)\gv{\lambda}_f^+ = \v{0} & \label{eq:deom_begin} \\
\v{c}_b(\q^+) + \v{E}(\q^+)\gv{\lambda}_b^+ = \v{0} & \\
\gv{\phi}_n(\v{q}^+, \gv{\lambda}^+)  = \v{0} & \\
\gv{\phi}_f(\v{u}^+, \gv{\lambda}^+)  = \v{0} & \\ 
\q^+ - \q^- - \dt\v{G}\v{u}^+ = \v{0} &. \label{eq:deom_end}
\end{align}

Here $\v{u}^+, \gv{\lambda}^+$ are the unknown velocities and multipliers at the end of the time-step. The constant $\tilde{\u} = \v{u}^- + \dt\v{G}^T\v{M}^{-1}\v{f}(\v{q}^-, \qdot^-)$ is the unconstrained velocity that includes the external and gyroscopic forces integrated explicitly. \addrev{Observe that through this time-discretization the original force level model of friction has changed into an impulsive model.}

We highlight a few differences from common formulations. First, the equality and inequality constraints have not been linearized through an index reduction step \cite{hairer2010solving}. Index reduction is a common practice that reduces the order of the DAE by solving only for the constraint time-derivatives, e.g.: $\dot{\v{c}}_b=\v{0}$. Index reduction results in a linear problem, but requires additional stabilization terms to combat drift and move the system back to the constraint manifold. These stabilization terms are often applied as the equivalent of penalty forces \cite{ascher1995stabilization} and are known as a source of instability and tuning issues. Work has been done to add additional post-stabilization passes \cite{cline2003post}, however these require solving nonlinear projection problems, which gives up the primary benefit of performing the index reduction step. Our discrete equations of motion are also nonlinear, but they require no additional stabilization terms or projection passes.

A second point we highlight is that the friction cone defined through \eqref{eq:friction_min} has not been linearized into a faceted cone as is commonly done \cite{stewart2000implicit, kaufman2008staggered}. The faceted cone approximation provides a simpler linear complementarity problem (LCP), but increases the number of Lagrange multipliers required (one per-facet), and introduces an approximation bias where frictional effects are stronger in some directions than others. In the next section we propose a method that solves the NCP corresponding to smooth isotropic friction using a fixed number of Lagrange multipliers.

\add{The implicit time-discretization above treats contact forces as impulsive, meaning they are able to prevent sliding and interpenetration instantaneously (over a single time-step in the discrete setting). This avoids the inconsistency raised by Painlev\'e in the continuous setting.}

\begin{figure}
	\includegraphics[width=0.75\columnwidth]{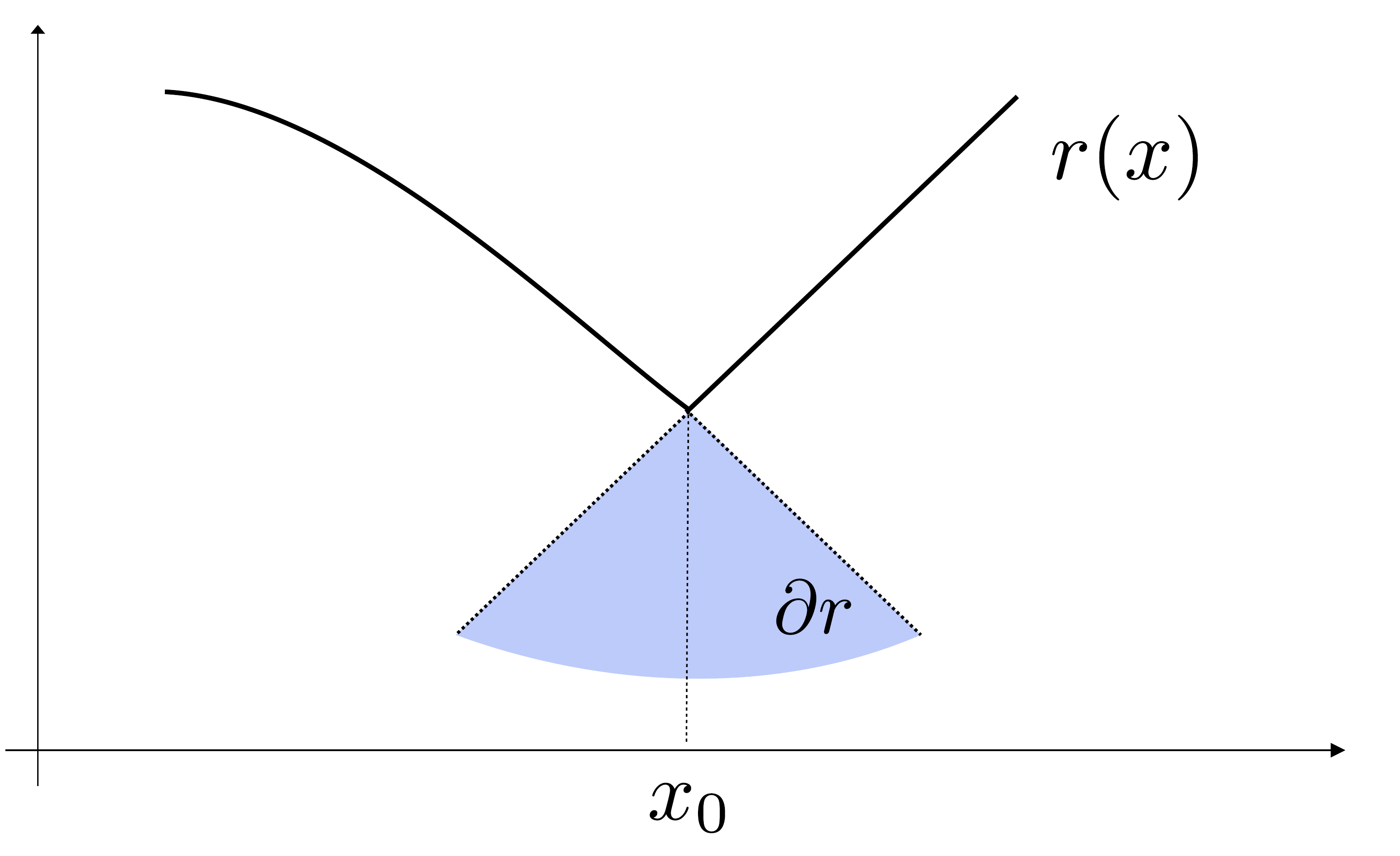}
	\caption{The derivative of a non-smooth function is set valued at discontinuities. The shaded area \rem{at} represents the generalized Jacobian $\partial r(\v{x}_0)$, defined as the convex hull of directional derivatives at $x_0$.}
	\label{fig:subgradient}
\end{figure}

\section{Non-Smooth Newton}

We now develop a method to solve the discretized equations of motion \eqref{eq:deom_begin}-\eqref{eq:deom_end}. Our starting point is Newton's method which we briefly review here. Given a set of nonlinear equations $\v{r}(\v{x}) = \v{0}$ in terms of a vector valued variable $\v{x}$, Newton's method gives a fixed point iteration in the following form:

\begin{align}\label{eq:newton_iteration}
\v{x}^{n+1} = \v{x}^n - \v{A}^{-1}(\v{x}^n)\v{r}(\v{x}^n),
\end{align}

\noindent where $n$ is the Newton iteration index. Newton's method chooses $\v{A}$ specifically to be the matrix of partial derivatives evaluated at the current system state, i.e.:

\begin{align}
\v{A} = \pdv{\v{r}}{\v{x}} = \begin{bmatrix} 
\pddv{\v{r}_1}{\v{x}_1} & \cdots & \pddv{\v{r}_1}{\v{x}_m} \\ \vdots & \ddots & \vdots \\
\pddv{\v{r}_n}{\v{x}_1} & \cdots & \pddv{\v{r}_n}{\v{x}_m} \end{bmatrix} .
\end{align}

Although Newton's method is most commonly used for solving systems of smooth functions it may also be applied to non-smooth functions by generalizing the concept of a derivative \cite{pang1990newton}. Qi and Sun \shortcite{qi1993nonsmooth} showed that Newton will converge for non-smooth functions provided $\v{A} \in \partial\v{r}(\v{x}^k)$, where $\partial\v{r}$ is the \textit{generalized Jacobian} of $\v{r}$ at $\v{x}^k$ defined by Clarke \shortcite{clarke1990optimization}. Intuitively, this is the convex hull of all directional derivatives at the non-smooth point, as illustrated in Figure \ref{fig:subgradient}. The derivatives of our NCP-functions given in the previous section belong to this subgradient, and we can use them in the fixed-point iteration of \eqref{eq:newton_iteration} directly. Algorithms of this type are sometimes referred to as \textit{semismooth} methods, \add{we refer the reader to the article by Hinterm\"uller \shortcite{hintermuller2010semismooth} for a more mathematical introduction and the precise definition of semismoothness}. In many cases $\v{A}$ is not inverted directly, and the linear system for $\Delta\v{x}$ may only be solved approximately. When this is the case we refer to the method as an inexact Newton method. Additionally, when the partial derivatives in $\v{A}$ are also approximated we refer to it as a quasi-Newton method. In our method we make use of both approximations.

\begin{algorithm}[t]
			\While{Simulating}{
		Perform Collision Detection\;
		
		\For{N \rem{Outer} \add{Newton} Iterations
		}{
			Assembly $\v{M}, \v{H}, \v{J}, \v{C}, \v{g}, \v{h}, \v{b}$\;
			\For{M \rem{Inner} \add{Linear} Iterations
			}{
				Update Solution to $[\v{J}\v{H}^{-1}\v{J}^T + \v{C}]\Delta\gv{\lambda} = \v{b}$\;
			}

			Perform Line Search to find $t$ (optional)
	
			$\gv{\lambda}^{n+1} = \gv{\lambda}^{n} + t\Delta\gv{\lambda}$ \\
			$\v{u}^{n+1} = \v{u}^{n} + t\Delta\v{u}$ \\
			$\v{q}^{n+1} = \v{q}^{n} + \dt\v{G}\v{u}^{n+1}$
		}
	}
	\caption{Simulation Loop}
	\label{alg:loop}
\end{algorithm}

\subsection{System Assembly}

Linearizing our discrete equations of motion \eqref{eq:deom_begin}-\eqref{eq:deom_end} each Newton iteration provides the following linear system in terms of the change in system variables:

\begingroup
\renewcommand*{\arraystretch}{1.3}
\begin{align}\label{eq:newton_system}
\begin{bmatrix}
\tilde{\v{M}} - \dt^2\v{K} & -\v{J}_b^T & -\v{J}_n^T & - \v{J}_f^T \\ 
\v{J}_b & \frac{\v{E}}{\dt^2} & \v{0} & \v{0} \\
\v{J}_n & \v{0} & \frac{\v{S}}{\dt^2} & \v{0}  \\
\v{J}_f & \v{0} & \v{0} & \frac{\v{W}}{\dt}\end{bmatrix}
\begin{bmatrix*}[l]\Delta\u \\ \Delta\gv{\lambda}_b\dt \\ \Delta\gv{\lambda}_n\dt \\ \Delta\gv{\lambda}_f\dt\end{bmatrix*} = -\begin{bmatrix} \v{g} \\ \v{h}_b \\ \v{h}_n \\ \v{h}_f\end{bmatrix}
\end{align}
\endgroup \\

where $\v{K}$ is the geometric stiffness matrix arising from the spatial derivatives of constraint forces discussed in Section \ref{sec:geometric_stiffness}. The lower-diagonal blocks are the derivatives of our contact functions with respect to their Lagrange multipliers

\begin{align}
\v{S}=\pdv{\gv{\phi}_n}{\gv{\lambda}_n}, \hspace{1.0em} \v{W}=\pdv{\gv{\phi}_f}{\gv{\lambda}_f}.
\end{align}

Grouping the sub-block components such that

\begin{align}
\v{H} = \begin{bmatrix}\tilde{\v{M}} - \dt^2\v{K}\end{bmatrix}, \hspace{0.5em}\v{J} = \begin{bmatrix} \v{J}_b \\ \v{J}_n \\ \v{J}_f \end{bmatrix}, \hspace{0.5em} \v{C} = \begin{bmatrix}\frac{\v{E}}{h^2} & \v{0} & \v{0} \\
\v{0} & \frac{\v{S}}{h^2} & \v{0}  \\
\v{0} & \v{0} & \frac{\v{W}}{h}\end{bmatrix}, \hspace{0.5em} \gv{\lambda} = \begin{bmatrix}\gv{\lambda}_b \\ \gv{\lambda}_n \\ \gv{\lambda}_f\end{bmatrix}
\end{align}

\noindent we can write the system compactly as,

\begin{align}\label{eq:group_system}
\begin{bmatrix}
\v{H} & -\v{J}^T \\ 
\v{J} & \v{C}\end{bmatrix}
\begin{bmatrix*}[l]\Delta\u \\ \Delta\gv{\lambda}\dt\end{bmatrix*} = -\begin{bmatrix} \v{g} \\ \v{h}\end{bmatrix}.
\end{align}

The right-hand side is given by evaluating our discrete equations of motion \eqref{eq:deom_begin}-\eqref{eq:deom_end} at the current Newton iterate. Here, $\v{g}$ is our momentum balance equation,

\begin{align}
\v{g} &= \tilde{\v{M}}\left(\v{u}^{n} - \tilde{\v{u}}\right) - \dt\v{J}^T\gv{\lambda}^n,
\end{align}

\noindent and $\v{h}$ is a vector of constraint errors,

\begin{align}
\v{h} &= \begin{bmatrix} \v{h}_b \\ \v{h}_n \\ \v{h}_f\end{bmatrix} = \begin{bmatrix}
\frac{1}{\dt}\v{1} & & \\
& \frac{1}{\dt}\v{1} & \\
& & \v{1}
\end{bmatrix}\begin{bmatrix}\v{c}_b(\q^n) + \v{E}(\q^n)\gv{\lambda}_b^n \\ \gv{\phi}_n(\q^n, \gv{\lambda}^n_n) \\ \gv{\phi}_f(\v{u}^n, \gv{\lambda}^n_f) \end{bmatrix}.
\end{align}

Here the left-hand side matrix should be considered acting block-wise on the constraint error. Since the frictional constraints are measured at the velocity level they are not scaled by $\frac{1}{h}$ like the positional constraints.

\add{The mass block matrix $\v{M}$ is evaluated at the beginning of the time-step using $\v{q}^-$, while the $\v{H}$ matrix is evaluated each Newton iteration as described in Section \ref{sec:geometric_stiffness}}. The friction compliance block, $\v{W}$, is evaluated at each Newton iteration using the current Lagrange multipliers. This means the frictional forces are bounded by the normal force from the previous Newton iteration. This is similar to the fixed point iteration described by Stewart and Trinkle \shortcite{stewart2000implicit} but with a smooth friction cone. It is also similar in principle to the Kaufman et al.'s Staggered Projections \shortcite{kaufman2008staggered}, with the difference that we update both normal and frictional forces during a single Newton iteration, rather than solving two separate minimizations. \add{For inactive contacts with $\lambda_n \ge 0$ we conceptually disable their frictional constraint equation rows by removing them from the system. In practice this can be performed by zeroing the corresponding rows to avoid changing the system matrix structure.}

\subsection{Schur-Complement}

The system \eqref{eq:group_system} is a saddle-point problem \cite{benzi2005numerical} that is indefinite and possibly singular. To obtain a reduced positive semi-definite system, we take the Schur-complement with respect to the mass-block to obtain

\begin{align}\label{eq:schur_system}
\left[\v{J}\v{H}^{-1}\v{J}^T + \v{C}\right]\Delta\gv{\lambda} = \frac{1}{\dt}\left(\v{J}\v{H}^{-1}\v{g} - \v{h}\right).
\end{align}

After solving this system for $\Delta\gv{\lambda}$ the velocity change $\Delta\v{u}$ may be evaluated directly, 

\begin{align}
\Delta\v{u} = \v{H}^{-1}\left(\v{J}^T\Delta\gv{\lambda}\dt - \v{g}\right)
\end{align}

\noindent and the system updated accordingly,

\begin{align}
\gv{\lambda}^{n+1} &= \gv{\lambda}^{n} + t\Delta\gv{\lambda} \\
\v{u}^{n+1} &= \v{u}^{n} + t\Delta\v{u} \\
\v{q}^{n+1} &= \v{q}^{n} + \dt\v{G}\v{u}^{n+1}.
\end{align}

Here $t$ is a step-size determined by line-search or other means (see Section \ref{sec:robustness}). We refer to \eqref{eq:schur_system} as the Newton \textit{compliance formulation}. Under certain conditions we could alternatively have taken the Schur-complement of \eqref{eq:group_system} with respect to the compliance block $\v{C}$, instead of the mass block $\v{H}$. This transformation is only possible if $\v{C}$ is non-singular, but it leads to what we call the Newton \textit{stiffness formulation},

\begin{align}
\left[\v{H} + \v{J}^T\v{C}^{-1}\v{J}\right]\Delta\u = -\v{g} - \v{J}^T\v{C}^{-1}\v{h}.
\end{align}

This form is closely related to that of Projective Dynamics \cite{bouaziz2014projective}, and arises from a linearization of the elastic forces due to a quadratic energy potential. Having $\v{C}$ be invertible corresponds to having compliance everywhere, or in other words, no perfectly hard constraints. For stiff materials, where $\v{C}$ is poorly conditioned, this approach leads to numerical problems in calculating $\v{C}^{-1}$. However, if the system has fewer degrees of freedom than constraints this transformation can result in a smaller system. In this work we are interested in methods that combine stiff constraints with deformable bodies, so we use the compliance form which accommodates both.

\begin{figure}
	\begin{subfigure}{0.5\columnwidth}
		\includegraphics[width=\columnwidth]{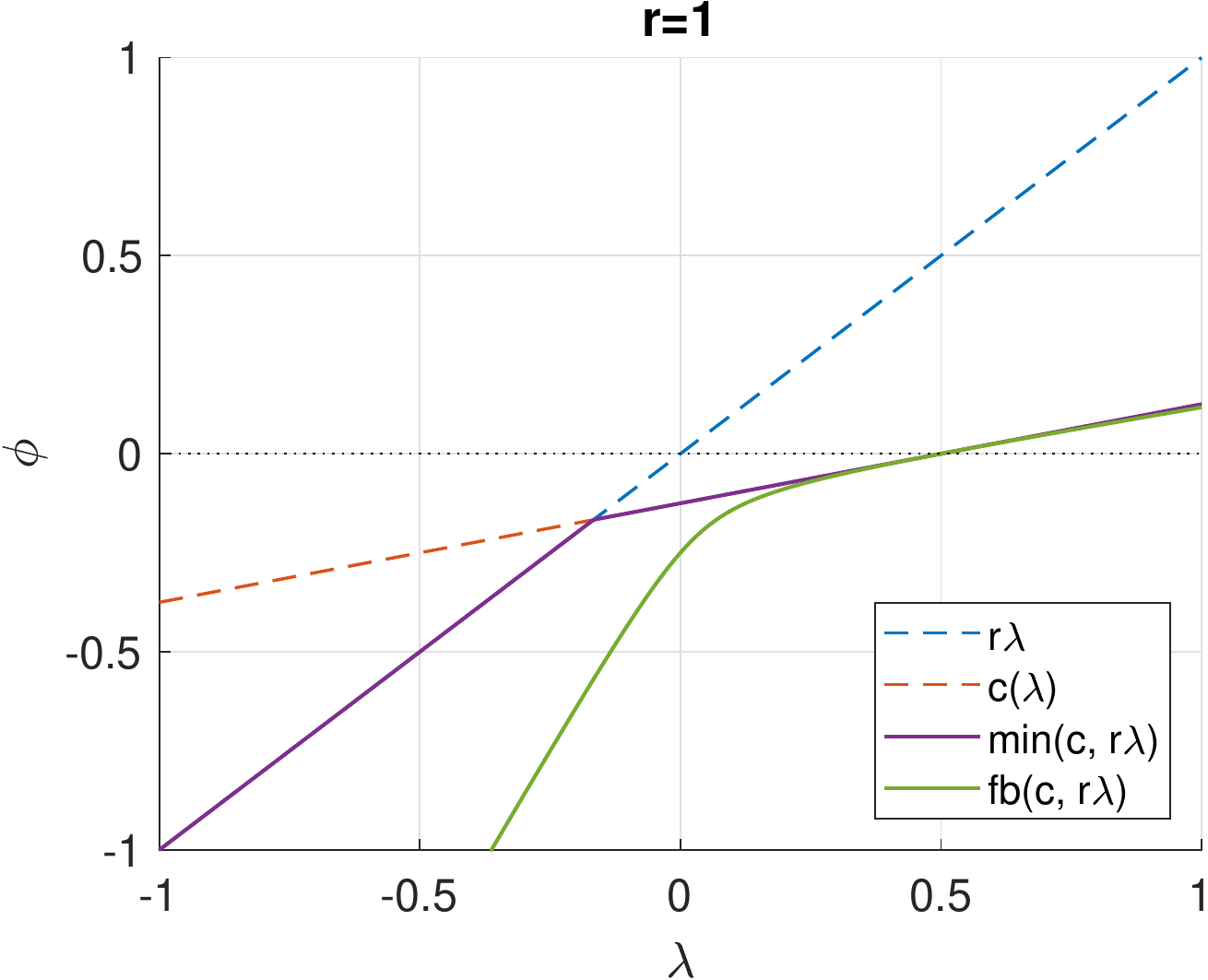}
	\end{subfigure}
	~
	\begin{subfigure}{0.5\columnwidth}
		\includegraphics[width=\columnwidth]{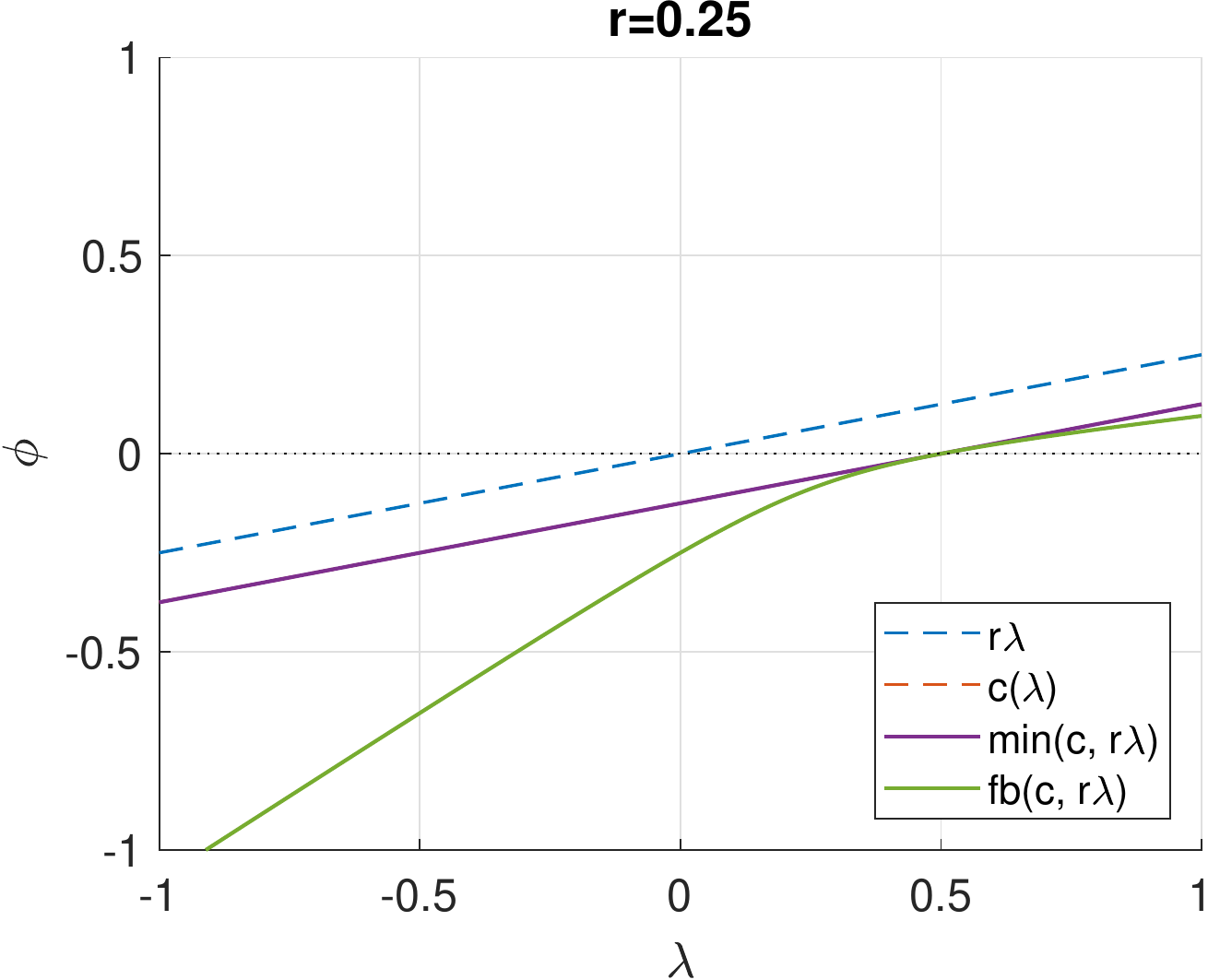}
	\end{subfigure}
	\caption{\textbf{Left:} The minimum-map of a unilateral constraint has a kink in it at the cross over point. \textbf{Right: } We propose a preconditioner that removes this discontinuity by forcing both terms to be parallel. For a single constraint this results in a straight-line error function that can be solved in one step regardless of starting point (right). In the case of Fischer-Burmeister (green) the error function's curvature is reduced. For illustration purposes we have shown the constraint function $c = \frac{1}{4}\lambda - \frac{1}{8}$ > 0, which has a unique solution at $\lambda=\frac{1}{2}$.}
	\label{fig:rscalingvis}
\end{figure}

\section{Complementarity Precondtioning}

An interesting property of the non-smooth complementarity formulations is that their solutions are invariant up to a constant positive scale $r$ applied to either of the arguments. Specifically, a solution to $\phi(a, b) = 0$, is also a solution to the scaled problem, $\phi(a, rb) = 0$. \add{Alart \shortcite{alart1997methode} presented an analysis of the optimal choice of $r$ in the context of linear elasticity in a quasistatic setting}. Erleben \shortcite{erleben2017rigid} also explored this free parameter in the context of proximal-map operators for rigid bodies.

In this section we propose a new complementarity preconditioning strategy to improve convergence by choosing $r$ based on the system properties and discrete time-stepping equations. To motivate our method, we make the observation that the two sides of the complementarity condition typically have different physical units. For example, a contact NCP function

\begin{align}
\phi(C_n, \lambda_n) = 0
\end{align}

\noindent has the units of meters for the first parameter, and units of Newtons for the second parameter. This mismatch can lead to poor convergence in a manner similar to the effect of row-scaling in traditional iterative linear solvers.

Inspired by the use of diagonal preconditioners for linear solvers, our idea is to use the effective system mass and time-stepping equations to put both sides of the complementarity condition into the same units. The appropriate $r$-factor to perform this scaling comes from the relation between $\gv{\lambda}$ and our constraint functions, given through the equations of motion. Specifically, for a unilateral constraint with index $i$ we set $r_i$ to be,

\begin{equation}\label{eq:rscaling}
r_i = \dt^2\left[\v{J}\tilde{\v{M}}^{-1}\v{J}^T\right]_{ii}.
\end{equation}
\\
Intuitively, this is the time-step scaled effective mass of the system, and relates how a change in Lagrange multiplier affects the corresponding constraint value. This has the effect of making both sides of the complementarity equation have the same slope, as illustrated in Figure \ref{fig:rscalingvis}. \add{For a friction constraint $\phi_f$ the correct scaling factor is $r_i = \dt\left[\v{J}\tilde{\v{M}}^{-1}\v{J}^T\right]_{ii}$ since this is a velocity-force relationship. When using a Jacobi preconditioner the diagonal of the system matrix is already computed, so applying $r_i$ to the NCP function incurs little overhead.} We discuss the effect of preconditioning strategies in Section \ref{sec:rscaling_discussion}.

\begin{figure}

	\begin{subfigure}{0.5\columnwidth}
		\includegraphics[width=\columnwidth]{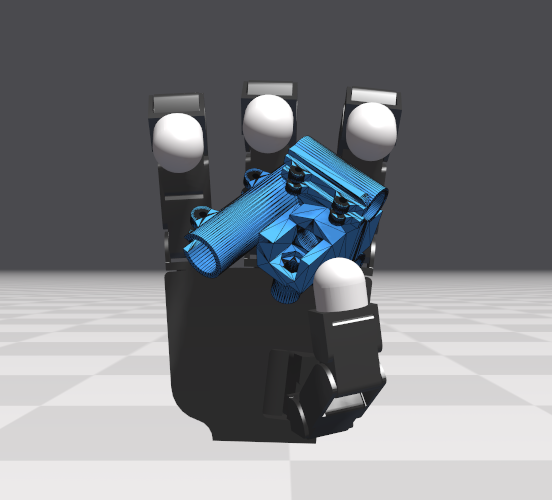}
	\end{subfigure}
	~
	\begin{subfigure}{0.5\columnwidth}
		\includegraphics[width=\columnwidth]{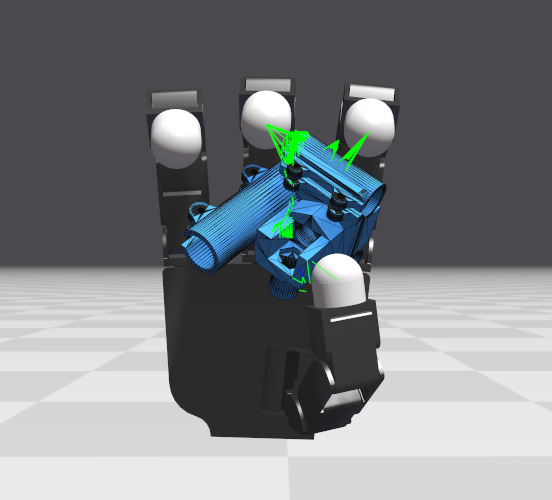}
	\end{subfigure}
	\par\vspace{0.1em}
	\begin{subfigure}{0.5\columnwidth}
		\includegraphics[width=\columnwidth]{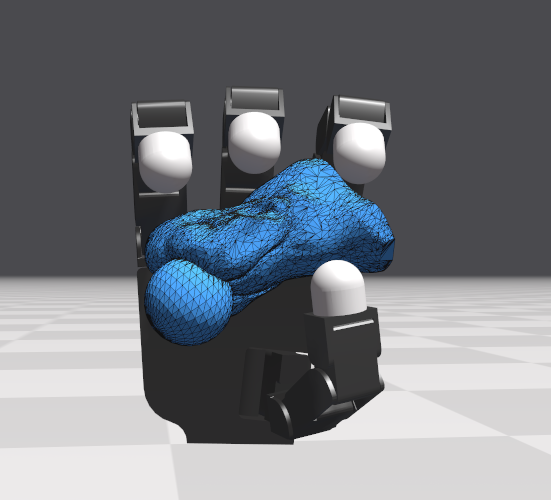}
	\end{subfigure}
	~
	\begin{subfigure}{0.5\columnwidth}
		\includegraphics[width=\columnwidth]{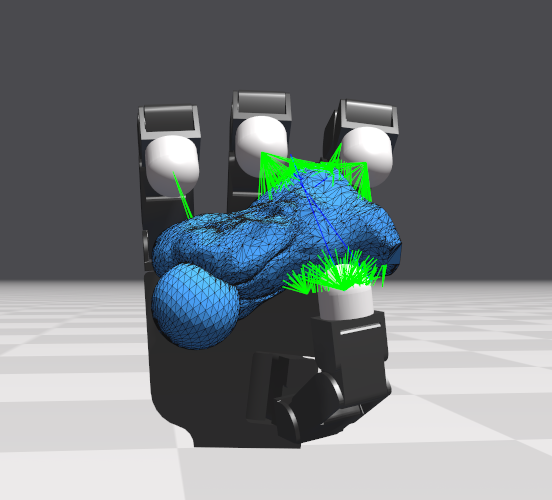}
	\end{subfigure}
	
	\caption{A selection of grasping tests using the Allegro hand and objects from the DexNet adversarial mesh collection \cite{mahler2017dex}. Our method simulates stable grasps around the irregular objects, and is robust to the over-specified contact sets generated by the high-resolution and often non-manifold meshes (right).}
	\label{fig:dexnet}
\end{figure}

\section{Robustness}\label{sec:robustness}

\subsection{Line Search and Starting Iterate}

We implemented a back-tracking line-search based on a merit function defined as the $L_2$ norm of our residual vector. \add{For frictionless contact problems we found this worked well to globalize the solution. However, for frictional problems we found line search would often cause the iteration to stall and make no progress. We believe this is related to the fact that the frictional problem is non-convex and our search direction is not necessarily a descent direction.} Instead, all our results use a simple damped Newton strategy that accepts a constant fraction of the full Newton step with a factor $t \in (0, 1)$. This strategy may cause a temporary increase in the merit function, but can result in overall better convergence \cite{maratos1978exact}. Watchdog strategies \cite{nocedal2006numerical} may be employed to make stronger convergence guarantees, however we did not find them necessary. We have used a value of $t=0.75$ for all examples unless otherwise specified. \add{This value is ad-hoc, but we have not found our method to be particularly sensitive to its setting.}

Starting iterates have a strong effect on most optimization methods, and ours is no different. Although it is common to initialize solvers with the unconstrained velocity at the start of the time-step we found the most robust method was to use the zero-velocity solution as a starting iterate, i.e.:

\begin{align}
\v{u}^0 &= \v{0} \\
\gv{\lambda}^0 &= \v{0}.
\end{align}

This choice is robust in the case of reinforcement-learning, where large random external torques are applied to bodies that can lead to initial points far from the solution. Warm-starting for constraint forces is possible, and our tests indicate this can give a good improvement in efficiency, however due to the additional book-keeping we have not used warm-starting in our reported results.

\begin{figure}
	\begin{subfigure}{0.5\columnwidth}
	\includegraphics[width=\columnwidth]{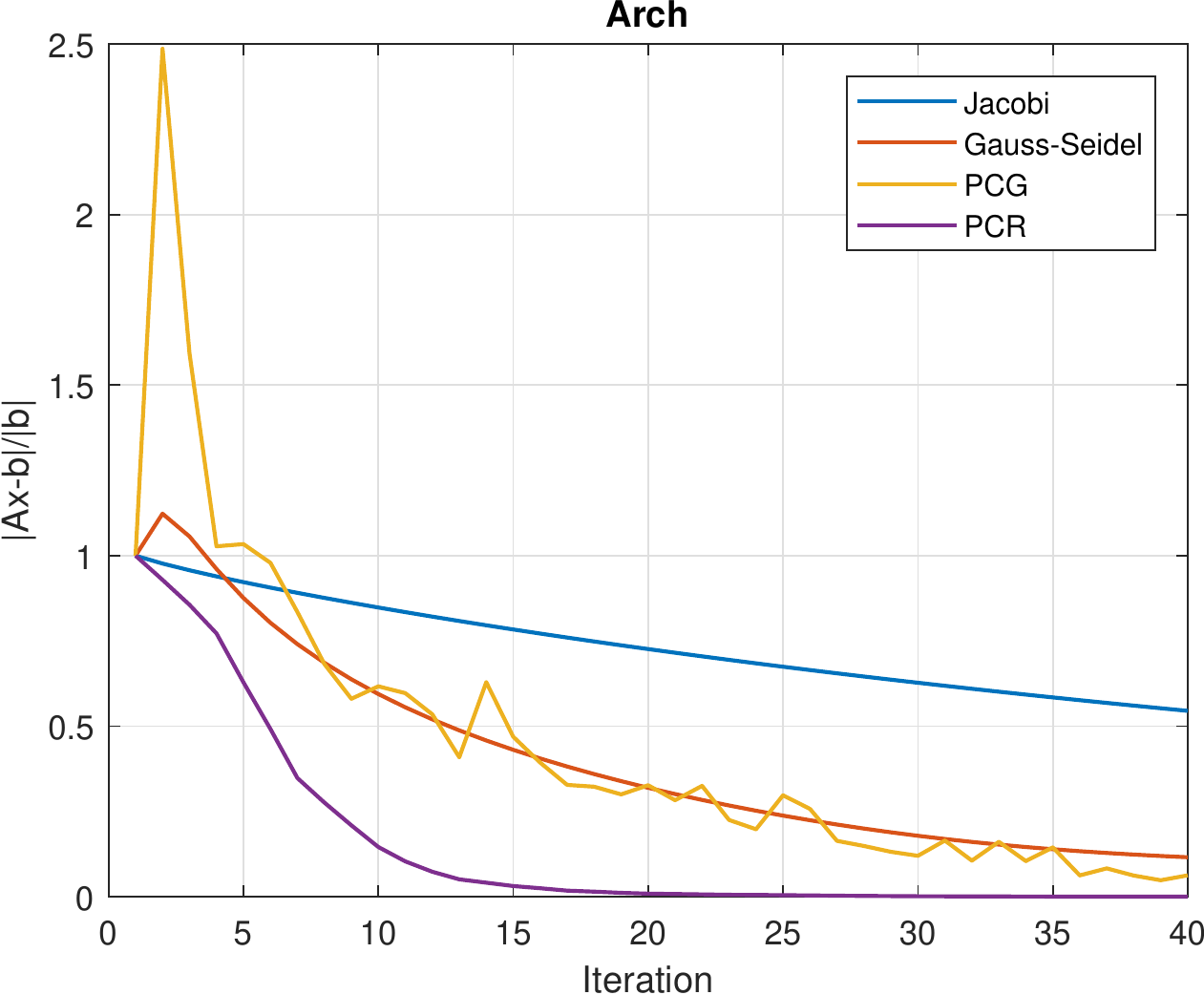}
	\end{subfigure}
	~
	\begin{subfigure}{0.5\columnwidth}
	\includegraphics[width=\columnwidth]{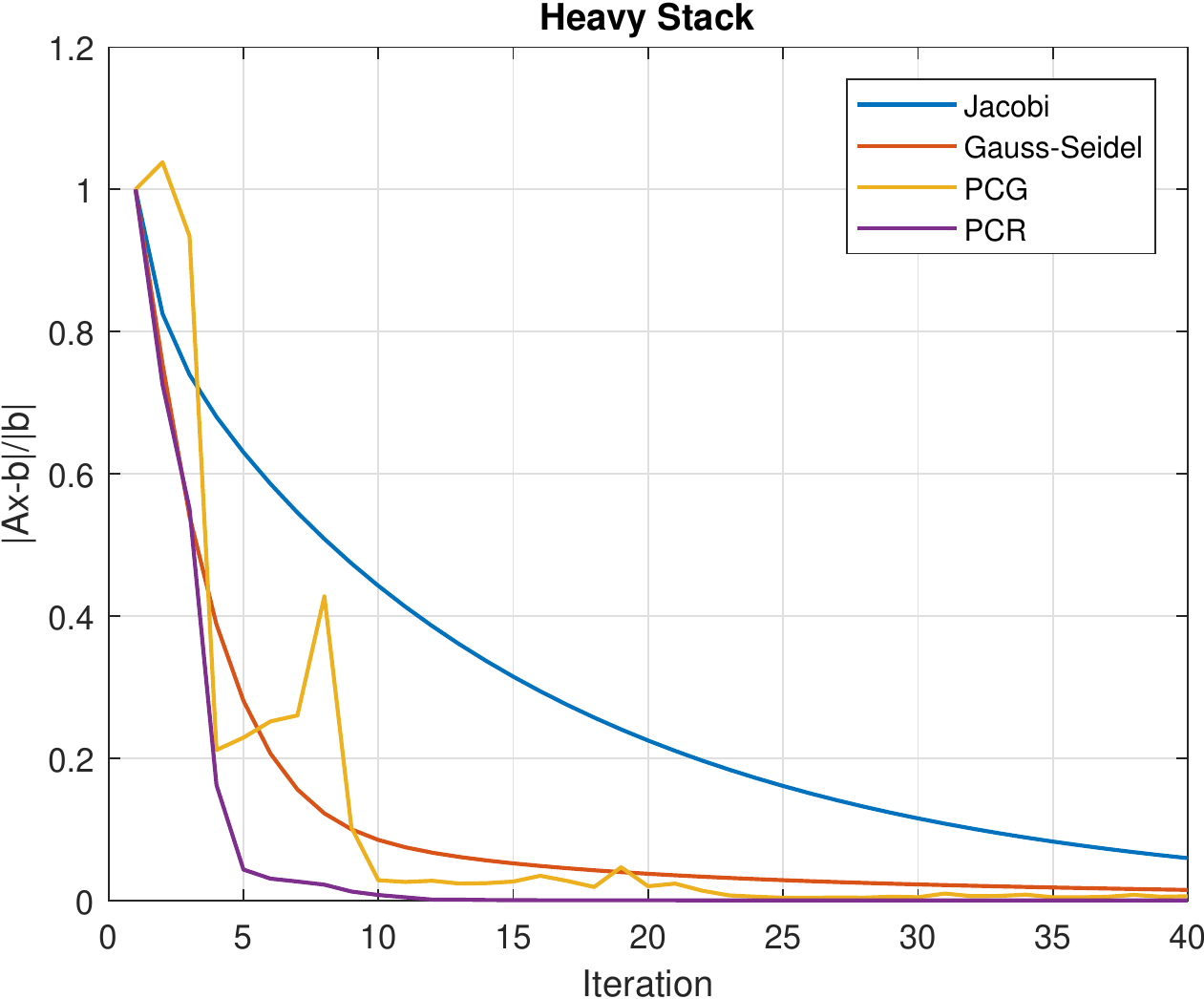}
	\end{subfigure}
	\caption{Convergence of different iterative methods for a single linear sub-problem on two different examples. Preconditioned Conjugate Gradient (PCG) shows characteristic non-monotone behavior that causes problems with early termination. Preconditioned Conjugate Residual (PCR) is monotone (in the preconditioned norm), which ensures a useful result even at low iteration counts.}
	\label{fig:linear_solve_convergence}
\end{figure}

\subsection{Preconditioned Conjugate Residual}

A key advantage of our non-smooth formulation is that it allows black-box linear solvers to be used for nonlinear complementarity problems. Nevertheless, the choice of solver is still an important factor that affects performance and robustness. A common issue we encountered is that even simple contact problems may lead to singular Newton systems. Consider for example a tessellated cylinder resting flat on the ground. In a typical simulator a ring of contact points would be created, leading to an under-determined system, i.e.: there are infinitely many possible contact force magnitudes that are valid solutions. This indeterminacy results in a singular Newton system and causes problems for many common linear solvers. The ideal numerical method should be insensitive to these poorly conditioned problems.

In addition, we seek a method that allows solving each Newton system only inexactly. Since the linearized system is only an approximation it would be wasteful to solve it accurately far from the solution. Thus, another trait we seek is a method that smoothly and monotonically decreases the residual so that it can be terminated early e.g.: after a fixed computational time budget. Finally, since our application often requires real-time updates we also look for a method that is amenable to parallelization.

\add{The conjugate gradient (CG) method \cite{hestenes1952methods} is a popular method for solving linear systems in computer graphics. It is a Krylov space method that minimizes the quadratic energy $e = \frac{1}{2}\v{x}^T\v{A}\v{x} - \v{b}^T\v{x}$. One side effect of this structure is that the residual $r = |\v{A}\v{x} - \v{b}|$ is not monotonically decreasing. This behavior leads to problems with early termination since, although a given iterate may be closer to the solution, the true residual may actually be larger. This manifests as constraint error changing unpredictably between iterations.}

\add{A related method that does not suffer from this problem is the conjugate residual (CR) algorithm \cite{saad2003iterative}. It is a Krylov space method similar to conjugate gradient (CG), however, unlike CG each CR iteration $k$ minimizes the residual}

\begin{align}
r_k = ||\v{A}\v{x} - \v{b}||_{\mathcal{K}_k}
\end{align}

\noindent within the Krylov space \add{$\mathcal{K}_k = \text{span}\{\v{b}, \v{A}\v{b}, \v{A}^2\v{b},\cdots, \v{A}^{k-1}\v{b}\}$}. A remarkable side effect of this minimization property is that CR is monotonically decreasing in the residual norm $r = |\v{A}\v{x} - \v{b}|$, and monotonically increasing in the solution variable norm \cite{fong2012cg}. These are exactly the properties we would like for an inexact Newton method. From a computational viewpoint it is nearly identical to CG, requiring one matrix--vector multiply, and two vector inner products per-iteration. CR requires an additional vector multiply-add per-iteration, but each stage is fully parallelizable, and in our experience we did not observe any performance difference to standard CG.

In Figure \ref{fig:linear_solve_convergence} we compare the convergence of four common linear solvers, Jacobi, Gauss-Seidel, preconditioned conjugate gradient (PCG), and preconditioned conjugate residual (PCR). We use a diagonal preconditioner for both PCR and PCG. Our surprising finding is that PCR often has an order of magnitude lower residual for the same iteration count compared to other methods. A similar result was reported by Fong et al. \shortcite{fong2012cg} in the numerical optimization literature. \add{For symmetric positive definite (SPD) systems CR will generate the same iterates as MINRES, however it does not handle semidefinite or indefinite problems in general.} In practice we observed that CR will converge for close to singular contact systems when CG and even direct Cholesky solvers may fail. We evaluate and discuss the effect of linear solver on a number of test cases in Section \ref{sec:linear_discussion}.

\begin{figure}
	\begin{subfigure}{0.5\columnwidth}
		\includegraphics[width=\columnwidth]{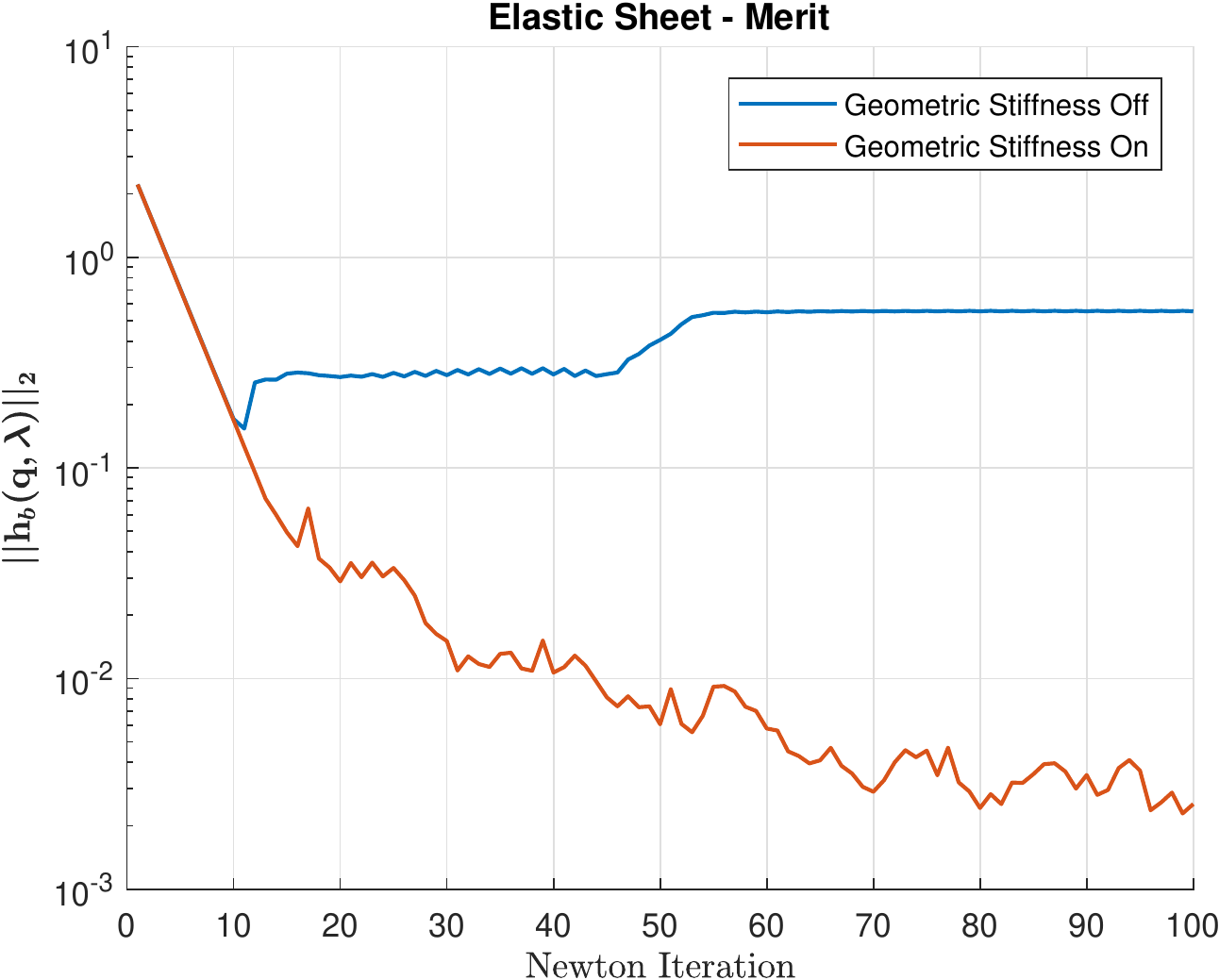}
	\end{subfigure}
	~
	\begin{subfigure}{0.5\columnwidth}
		\includegraphics[width=\columnwidth]{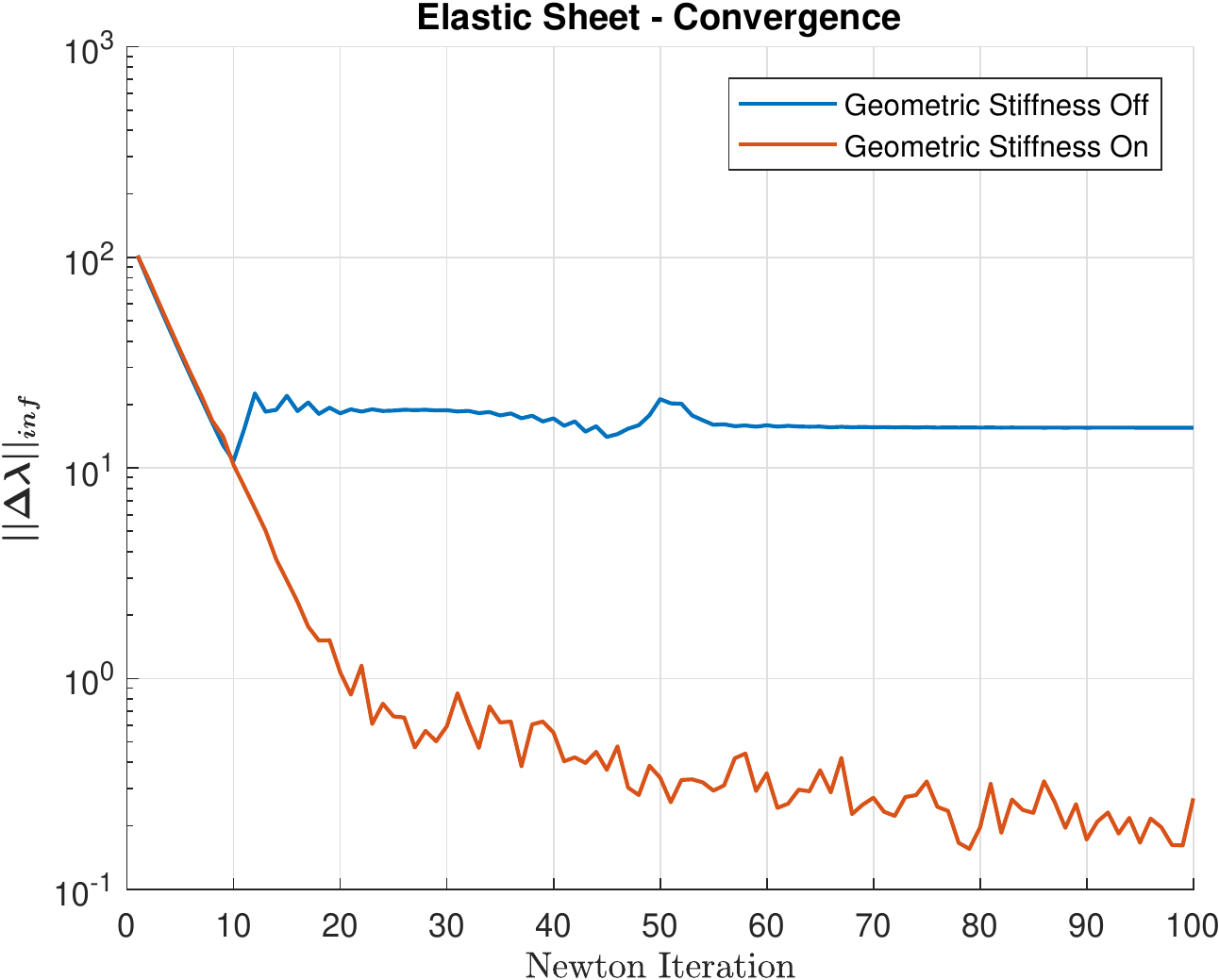}
	\end{subfigure}
	\caption{\add{A plot of the merit function and convergence with and without our geometric stiffness approximation on the stretched elastic sheet example. With geometric stiffness disabled we observe overshooting and oscillating iterates at large strains.}}
	\label{fig:geometric_merit}
\end{figure}

\subsection{Geometric Stiffness}\label{sec:geometric_stiffness}

The upper-left block of the system matrix, $\v{H} = \v{M} - \dt^2\v{K}$, consists of the mass matrix and a second term $\v{K}$, referred to as the \textit{geometric stiffness} of the system. This is defined as:

\begin{align}
\v{K} &= \pdv{\u}(\v{J}^T\gv{\lambda_i}). 
\end{align}

$\v{K}$ is not a physical quantity in the sense that it does not appear in the continuous or discrete equations of motion. It appears only as a side-effect of the numerical method, in this case the linearization performed at each Newton iteration. In practice $\v{K}$ is often dropped from the system matrix, leading to a quasi-Newton method. Tournier et al. \shortcite{tournier2015stable} observed that ignoring geometric stiffness can lead to instability, as the Newton search direction is free to move the iterate away from the constraint manifold in the transverse direction, eventually causing the iteration \add{to} diverge. However, including $\v{K}$ directly is also problematic because the mass block is no longer simple block-diagonal, and in addition, the possibly indefinite constraint Hessian restricts the numerical methods that can be applied. Inspired by the work of Andrews et al. \shortcite{andrews2017geometric} we \rem{now} introduce an approximation of geometric stiffness that improves the robustness of our Newton method without these drawbacks.

\begin{table*}
	\centering
	\caption{Parameters and statistics for different examples. All scenarios have used a time-step of $h=0.0083$s. Contact counts are for a representative step of the simulation. Performance cost is typically dominated by the linear solver step, while the matrix assembly cost is small. We have reported solver times for \add{a single time-step of} our GPU-based PCR solver. We note that for small problems the performance of iterative methods is limited by fixed cost kernel launch overhead and that scaling with problem size is typically sub-linear up to available compute resources. \add{For contacts we also report the maximum number of contacts in a single simulation island in brackets.} }
	\begin{tabular*}{\textwidth}{@{\extracolsep{\fill}}  l|c|c|c|c|c|c|c|c|c|c}
	Example      & Bodies & Joints  & Tetra  & Contacts &  Newton Iters & Linear Iters& \multicolumn{2}{c|}{Assembly (ms)} & \multicolumn{2}{c}{Solve (ms)} \\
	\hline
	&  \# & \# & \# & \# (island) & \# & \# & Avg & Std. Dev. & Avg & Std. Dev. \\	
	\hline
	Allegro DexNet & 21 & 21 & 0 & 104 \add{(104)} & 8 & 20     & \rem{0.8} \add{0.4}  & \rem{0.2} \add{0.1}  & \rem{20.1} \add{10.0}  & \rem{3.0} \add{1.5}  \\
Allegro Ball & 21 & 21 & 427 & 409 \add{(409)} & 6 & 50     & \rem{0.7} \add{0.35} & \rem{0.1} \add{0.05} & \rem{22.9} \add{11.4}  & \rem{2.4} \add{1.2} \\
PneuNet      & 1 & 0 & 2241 & 532 \add{(532)} & 4 & 80      & \rem{0.6} \add{0.3}  & \rem{0.1} \add{0.05} & \rem{45.5} \add{22.25} & \rem{2.7} \add{1.4} \\
Fetch Tomato & 29 & 28 & 319 & 371 \add{(371)} & 4 & 50     & \rem{0.9} \add{0.45} & \rem{0.1} \add{0.05} & \rem{24.4} \add{12.2}  & \rem{2.5} \add{1.25} \\
Fetch Beam   & 42 & 42 & 0 & 300 \add{(300)} & 6 & 50       & \rem{0.4} \add{0.2}  & \rem{0.2} \add{0.1}  & \rem{37.1} \add{18.6}  & \rem{4.5} \add{2.25} \\
Arch         & 20 & 0 & 0 & 134 \add{(134)} & 6 & 20        & \rem{0.3} \add{0.15} & \rem{0.1} \add{0.05} & \rem{15.7} \add{7.75}  & \rem{2.8} \add{1.4} \\
Table Pile   & 54 & 0 & 0 & 936 \add{(736)} & 4 & 20        & \rem{0.2} \add{0.1}  & \rem{0.1} \add{0.05} & \rem{7.4} \add{3.7}    & \rem{1.9} \add{0.95} \\
Humanoid Run & 5200 & 4800 & 0 & 10893 \add{(33)} & 4 & 10 & \rem{1.6} \add{0.8}  & \rem{0.1} \add{0.05} & \rem{17.8} \add{8.9}   & \rem{0.5} \add{0.25} \\
Yumi Cabinet & 6400 & 6600 & 0 & 4082 \add{(55)} & 5 & 25  & \rem{3.6} \add{1.8}  & \rem{0.5} \add{0.25} & \rem{112.0} \add{56.0} & \rem{12.8} \add{6.4} \\	\end{tabular*}
	\label{tab:stats}
\end{table*}

We now present a method for approximating $\v{H}$ that is related to the method of Broyden \shortcite{broyden1965class}. This method builds the Jacobian iteratively through successive finite differences. We apply this idea to build a diagonal approximation to the geometric stiffness matrix. Using the last two Newton iterates, we can define the following first-order approximation:

\begin{align}\label{eq:broyden_update}
\v{H}\Delta\v{u} = \left(\pddv{\v{g}}{\v{u}}\right)\Delta\v{u} \approx \v{g}(\v{u}^{n+1}) - \v{g}(\v{u}^n)
\end{align}

\noindent where $\v{H}$ is the unknown matrix we wish to find, and $\Delta\v{u} = \v{u}^{n+1} - \v{u}^{n}$ is the known difference between the last two iterates. This problem is under-determined since $\v{H}$ has $n_d\times n_d$ entries, while \eqref{eq:broyden_update} only provides $n_d$ equations, however if we assume $\v{H}$ has the special form:

\begin{align}
\tilde{\v{H}} \approx \v{M} - \tilde{\v{K}}
\end{align}

\noindent where $\tilde{\v{K}} = \text{diag}[c_1, c_2, \cdots, c_n]$ is a diagonal approximation to $\v{K}$, then the individual entries of $\tilde{\v{K}}$ are easily determined from \eqref{eq:broyden_update} by examining each entry:

\begin{align}
c_k = -\left[\frac{\v{g}(\v{u}^{n+1})_k - \v{g}(\v{u}^{n})_k + (\v{M}\Delta\v{u})_k}{\Delta\v{u}_k}\right].
\end{align}

\add{Each Newton iteration we then update the mass block's diagonal entries to form our approximate $\tilde{\v{H}}$ as}

\begin{align}\label{eq:mass_update_2}
\tilde{\v{H}}_{ii} = \tilde{\v{M}}_{ii} - \text{min}(0, c_k).
\end{align}

\add{For most models $\v{M}$ is block-diagonal, e.g.: 3x3 blocks for rigid body inertias, and so $\v{H}^{-1}$ is may be easily computed to form the Schur complement.} To ensure $\v{H}$ remains positive-definite we clamp the shift to be positive. \eqref{eq:schur_system}. This diagonal approximation is inspired by the method presented by Andrews et al. \shortcite{andrews2017geometric} however this approach has several advantages. First, we do not have to derive or explicitly evaluate the constraint Hessians for different constraint types. This means our method works automatically for contacts and complex deformable models. Second, we do not need to track the Lagrange multipliers from the previous frame, as our method updates the geometric stiffness at each Newton iteration. Finally, since our method does not change the solution to the problem it does not introduce any additional damping provided the solver is run to sufficient convergence. \add{Our approach bears considerable resemblance to a L-BFGS method \cite{nocedal1980updating}, however the use of a diagonal update means the system matrix structure is constant, allowing us to use the efficient block-diagonal inverse for the Schur complement.}

\subsection{Regularization}

Due to contact and constraint redundancy it is often the case that the system matrix is singular, \add{e.g.: for a chair resting with four legs on the ground the contact constraints are linearly dependent, meaning the system is underdetermined, and the Newton matrix becomes singular}. To improve conditioning we optionally apply an $\epsilon$ identity shift to the system matrix \eqref{eq:schur_system}. This type of regularization bears some resemblance to implicit penalty methods, however unlike penalty-based approaches it does not change the underlying physical model. The regularization only applies to the error at each Newton iteration, and the solution approaches the original one as the Newton iterations progress. For the examples shown here we set $\epsilon = 10^{-6}$, which is equivalent to adding a small amount of compliance to the system matrix. 

\begin{figure}
	\begin{subfigure}{\columnwidth}
		\includegraphics[width=\columnwidth]{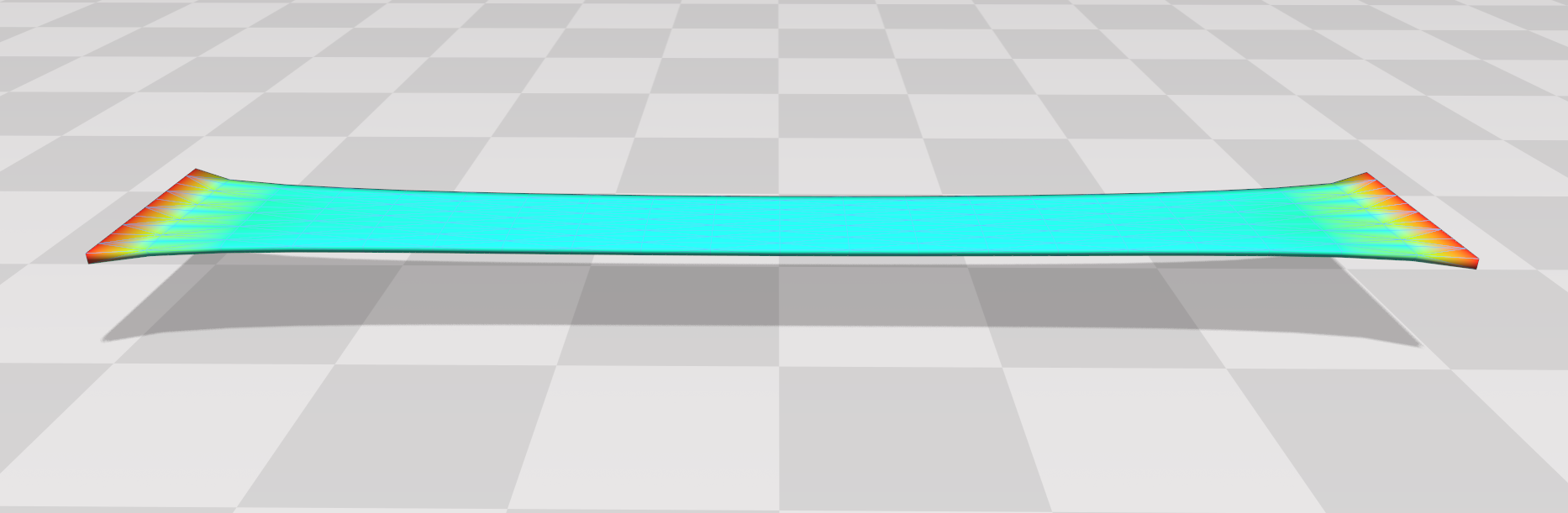}
	\end{subfigure}
	
	\begin{subfigure}{\columnwidth}
		\includegraphics[width=\columnwidth]{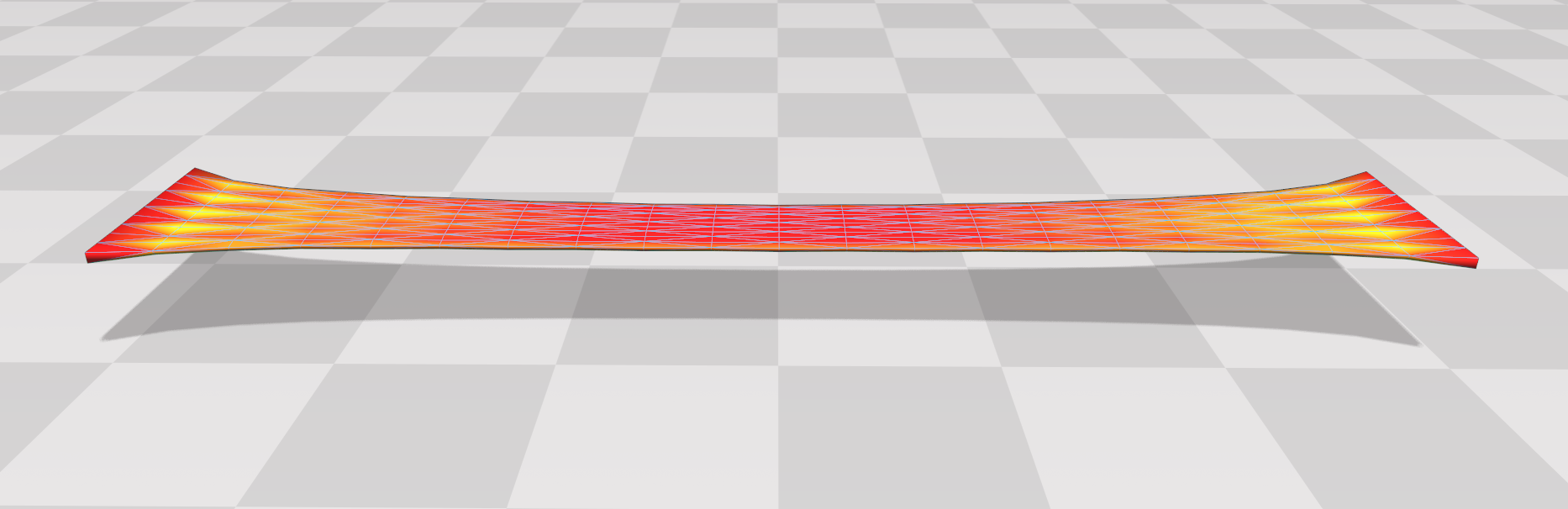}
	\end{subfigure}	
	\caption{A material extension test. Our generalized compliance formulation accommodates hyperelastic material models that strongly resist volume changes (top), while linear models show characteristic volume loss at large strains (bottom). }
	\label{fig:fem_stretch}
\end{figure}

\section{Modeling}\label{sec:deformables}

In the following sections we show how to generalize the compliance form of elasticity and discuss some of the physical models we have used in our examples.

\subsection{Generalized Compliance}

To extend our compliance form of elasticity to arbitrary material models and forces we make a new interpretation of the compliance transformation as a force factorization. Specifically, given a force $\v{f}(\q)$ we factor it as:

\begin{align}
\v{f}(\q) = -\v{J}^{T}(\q)\v{c}(\q)
\end{align}

\noindent where $\v{J}$ and $\v{c}$ may be scalar, vector, or matrix valued functions. The only requirements we place on the factors are that

\begin{align}
\pdv{\v{c}}{\q} = \v{N}(\q)\v{J}(\q),
\end{align}

\noindent where $\v{N}(\q)$ is an invertible linear mapping. The goal in choosing a factorization is to separate the force Jacobian into components where one is easily, even analytically, invertible. This splitting moves part of the force Jacobian out of the mass matrix, and onto the compliance matrix, where it becomes a regularization term. To perform the splitting we introduce a variable $\gv{\lambda} = -\v{c}$ which allows us to write our equations of motion (simplified for illustration) as,

\begin{align}
\v{M}\ddot{\q} - \v{J}^T\gv{\lambda} &= \v{0} \label{eq:balance} \\
\v{c} + \gv{\lambda} &= \v{0} \label{eq:constraint}.
\end{align}

In the context of a Newton solve, we require the linearization of \eqref{eq:constraint} which is given by

\begin{align}
\v{N}\v{J}\Delta\q + \v{I}\Delta\gv{\lambda} = - \left(\v{c}+\gv{\lambda}\right).
\end{align}

Here we have used the condition that $\pdv{\v{c}}{\q} = \v{N}(\q)\v{J}(\q)$. The advantage of this transformation occurs when we use the invertibility of $\v{N}$ to rewrite the linearization as,

\begin{align}
\v{J}\Delta\q + \v{E}\Delta\gv{\lambda} = -\v{E}\left(\v{c}+\gv{\lambda}\right)
\end{align}

\noindent where $\v{E} = \v{N}^{-1}$ is our compliance matrix. Essentially we have factored our force into two components, one of which has an analytically invertible component. This allows us to isolate a poorly conditioned components from the system matrix, e.g.: the material parameters, which may include large or even infinite values.

\subsection{Continuum Materials}

To illustrate the above transformation we show how to apply it to the case of a strain energy density $\Psi(\v{s})$, where $\v{s}$ is some parameterization of the density function, e.g.: strains, strain rates, or principal stretches. Given such a function the elastic potential energy is

\begin{align}
U(\q) = \int_V \Psi(\v{s}(\q)) dV,
\end{align}

\noindent where $V$ is the volume of integration, and the resulting force on the system is

\begin{align}
\v{f} = -\pdv{U}{\q}^T = -\left(\pdv{U}{\v{s}}\pdv{\v{s}}{\q}\right)^T.
\end{align}

To obtain the compliance form of this force we factorize it as $\v{c} = \pdv{U}{\v{s}}^T$ and $\v{J} = \pdv{\v{s}}{\q}^T$. The derivative of $\v{c}$ is then given by

\begin{align}
\pdv{\v{c}}{\q} = \pdv[2]{U}{\v{s}}\pdv{\v{s}}{\q} = \v{N}(\q)\v{J}(\q),
\end{align}

\noindent from which we can identify the compliance matrix

\begin{align}
\v{E} = \v{N}^{-1} = \left(\pdv[2]{U}{\v{s}}\right)^{-1}.
\end{align}

In terms of matrix assembly, each element of a finite element discretization contributes $n_s = \text{dim}(\v{s})$ constraints and Lagrange multiplier variables, and a $n_s\times n_s$ block to the system compliance matrix $\v{C}$.

\subsubsection{Linear Materials}

For a constant strain element with general linear Hookean constitutive model we have the strain energy

\begin{align}
U(\q) = V_e\frac{1}{2}\v{s}^T\v{K}_e\v{s},
\end{align}

\noindent where $\v{s} = [\epsilon_{xx}, \epsilon_{yy}, \epsilon_{zz}, \epsilon_{yz}, \epsilon_{xz}, \epsilon_{xy}]$ is a vector of strain elements, $\v{K}_e \in \mathbb{R}^{6\times 6} $ is the element's material stiffness matrix, and $V_e$ the element rest volume. The corresponding compliance matrix is $\v{E} = \left(V_e\v{K}_e\right)^{-1} $ which is a constant that may be precomputed.

\subsubsection{Hyperelastic Materials}

Linear material models may exhibit significant volume loss during large deformations. Hyperelastic models, where stiffness increases as a function of strain, are less prone to this artifact, as illustrated in Figure \ref{fig:fem_stretch}. To incorporate isotropic hyperelastic materials we may parameterize the strain energy density in terms of principal stretches $\v{s}(\q) = [s_1, s_2, s_3]$ of the deformation gradient $\v{F}(\q)$. The corresponding material stiffness matrix $\v{N}(\q) \in \mathbb{R}^{3\times 3}$ is no longer constant, but is a function of the system configuration. During the Newton solve we evaluate $\v{N}$ at each iteration and perform a direct inversion of it to obtain $\v{E}$. A practical consideration for hyperelastic materials is that, as $\v{E}$ is the inverse of a Hessian it may be indefinite, in which case it may be necessary to project it back to the positive definite cone before including it in our system matrices \cite{teran2005robust}, although we have also found that a simple diagonal approximation to $\v{E}$ is often sufficient. In Appendix \ref{app:smith} we give the derivation of $\v{E} = (\pdv[2]{U}{\v{s}})^{-1}$ for the stable Neo-Hookean model presented by Smith et al \shortcite{smith2018stable}.

\subsection{Particles}\label{sec:particles}

For deformable bodies we use tetrahedral finite elements defined over a set of particles where a particle with index $i$ contributes 3 degrees of freedom,

\begin{align}
\q_i = \begin{bmatrix} x \\ y \\ z\end{bmatrix}, \hspace{0.5em} \v{u}_i = \begin{bmatrix} v_x \\v_y \\v_z\end{bmatrix}.
\end{align}

The kinematic mapping for particles is the identity transform, $\v{G}_i = \v{1}_{3x3}$ and the lumped mass matrix is $\tilde{\v{M}}_i = m\v{1}_{3x3}$, where $m$ is the particle mass.

\subsection{Rigid Bodies}\label{sec:rigids}

We describe the state of a rigid body with index $i$ using a maximal coordinate representation consisting of the position of the body's center of mass, $\v{x}_i$, its orientation expressed as a quaternion $\gv{\theta}_i = [\theta_1, \theta_2, \theta_3, \theta_4]$, and a generalized velocity vector $\v{u}_i$

\begin{align}
\q_i = \begin{bmatrix}\v{x}_i \\ \gv{\theta}_i\end{bmatrix}, \hspace{0.5em} \v{u}_i = \begin{bmatrix}\dot{\v{x}}_i \\ \gv{\omega}_i\end{bmatrix},
\end{align}

\noindent where $\gv{\omega}$ is the body's angular velocity. The kinematic map from generalized velocities to the system coordinate time-derivatives is then given by $\qdot_i = \v{G}\v{u}_i$ where

\begin{align}
\v{G} = \begin{bmatrix}\v{1}_{3\times 3} & \v{0} \\ \v{0} & \frac{1}{2}\v{Q}\end{bmatrix},
\end{align}

\noindent where $\v{Q}$ is the matrix that performs a truncated quaternion rotation,

\begin{align}
\v{Q}(\gv{\theta}) = \begin{bmatrix}
-\theta_2 & -\theta_3 &  -\theta_4 \\
\theta_1 & \theta_4 &  -\theta_3 \\
-\theta_4 & \theta_1 &  \theta_2 \\
\theta_3 & -\theta_2 &  \theta_1 \\\end{bmatrix}.
\end{align}

The corresponding mass matrix is then

\begin{align}
\tilde{\v{M}}_i = \begin{bmatrix}m\v{1}_{3\times 3} & \v{0}\ \\ \v{0} & \v{I}\end{bmatrix},
\end{align}

\noindent where $m$ is the body mass and $\v{I}$ is the inertia tensor. When using quaternions to represent rotations there is an implicit constraint that $\|\gv{\theta}\|=1$, rather than include this directly in the Newton solve we periodically normalize the quaternion to unit-length.

\begin{figure}
	\begin{subfigure}{0.25\textwidth}
	\includegraphics[width=\textwidth,trim=0 20 0 20, clip]{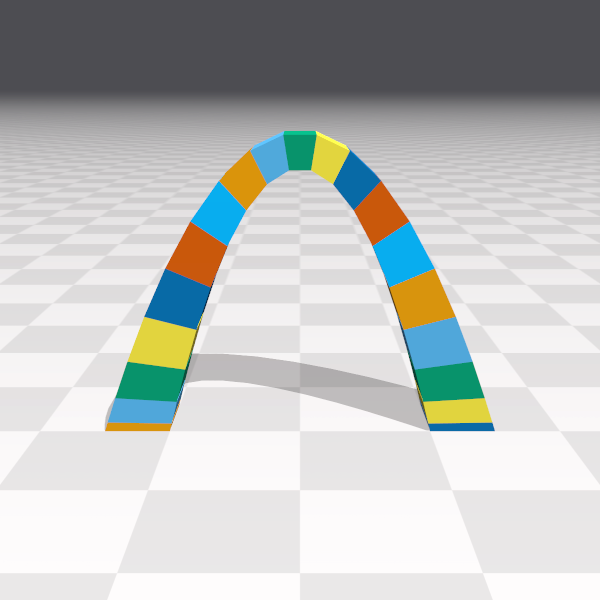}
	\end{subfigure}
	~
	\begin{subfigure}{0.25\textwidth}
	\includegraphics[width=\textwidth,trim=0 20 0 20, clip]{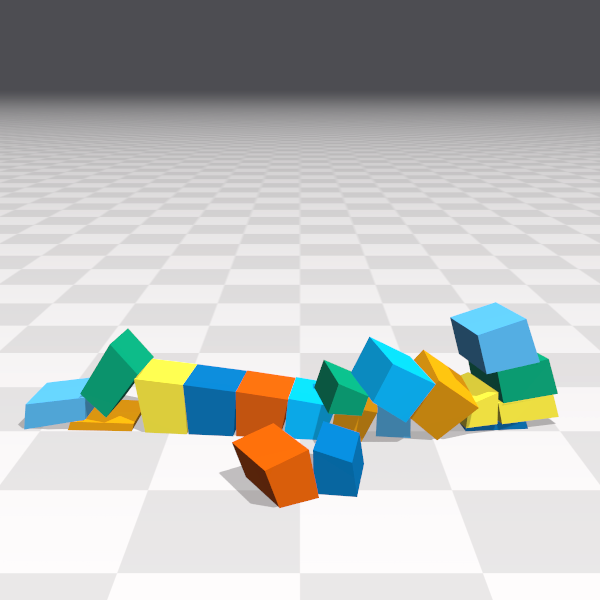}
	\end{subfigure}
	
	\par\vspace{0.1em}
	\begin{subfigure}{0.25\textwidth}
	\includegraphics[width=\textwidth,trim=0 20 0 20, clip]{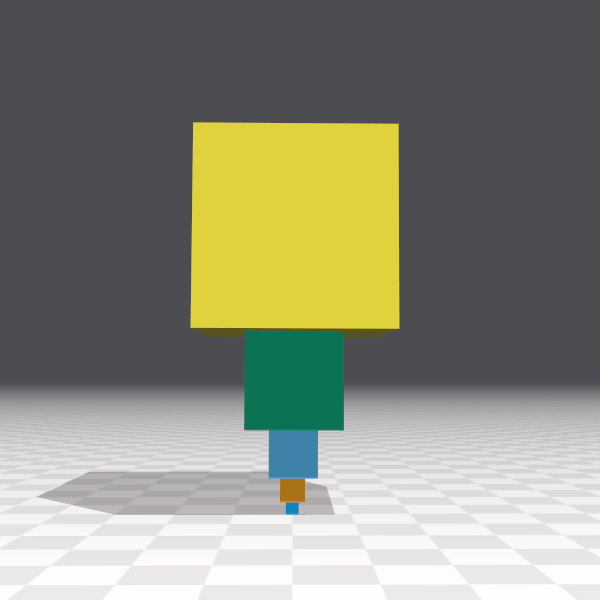}
	\end{subfigure}
	~
	\begin{subfigure}{0.25\textwidth}
		\includegraphics[width=\textwidth,trim=0 20 0 20, clip]{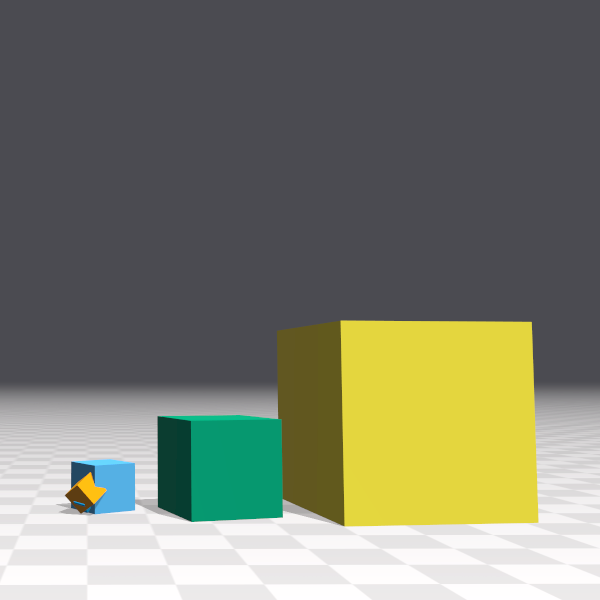}
	\end{subfigure}
	\par\vspace{0.1em}
	
	\begin{subfigure}{0.25\textwidth}
		\includegraphics[width=\textwidth,trim=150 40 200 0, clip]{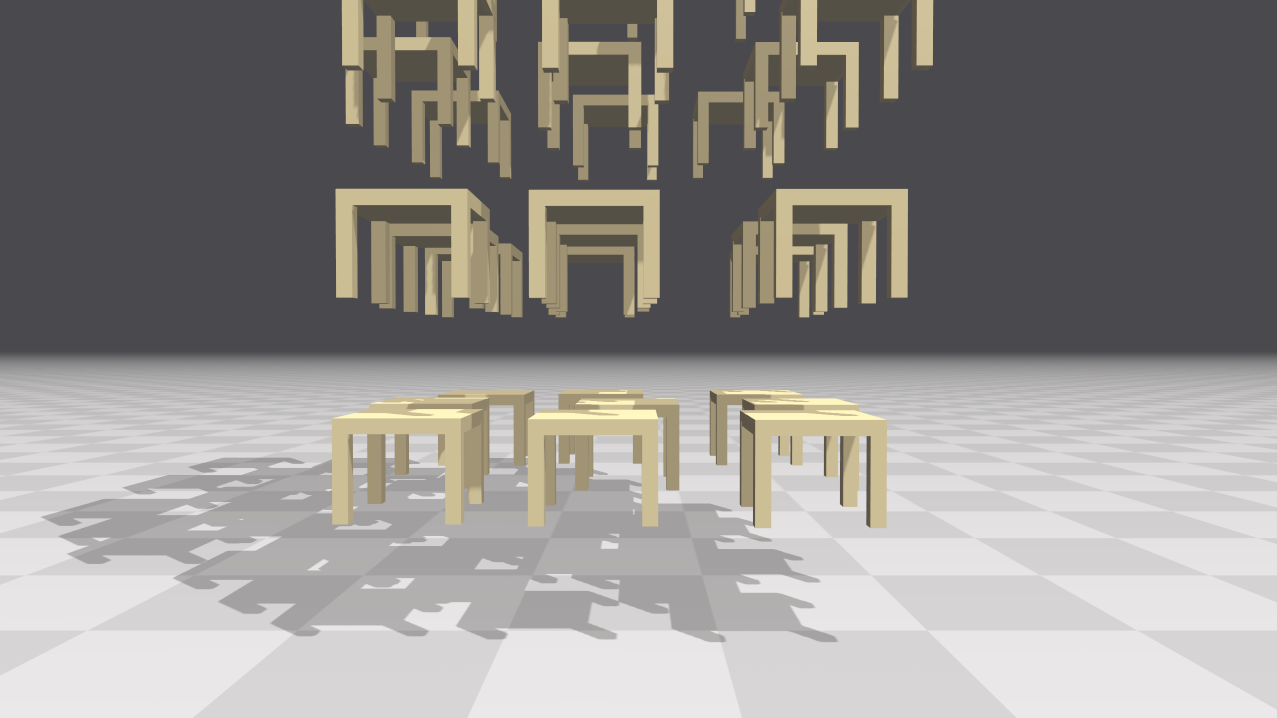}
	\end{subfigure}
	~
	\begin{subfigure}{0.25\textwidth}
		\includegraphics[width=\textwidth,trim=150 40 200 0, clip]{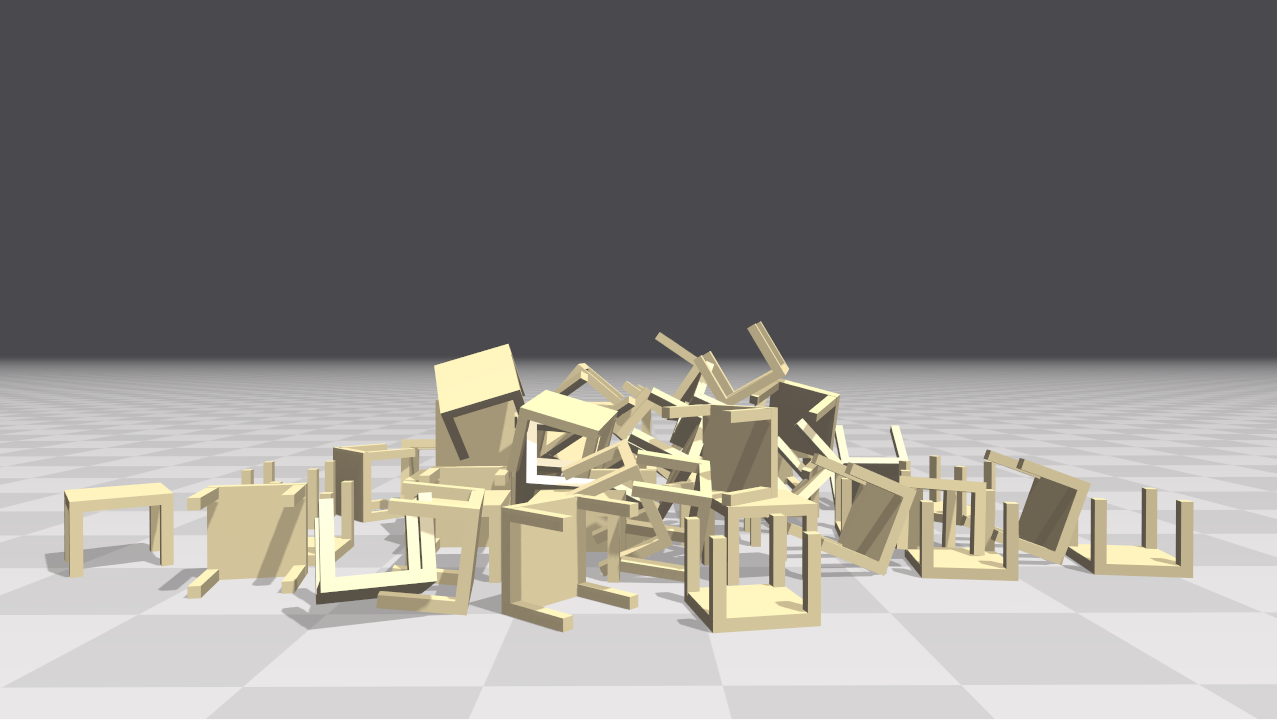}
	\end{subfigure}

	\caption{\textbf{Top}: A self-supporting parabolic arch. Direct or Krylov linear solvers can reduce the error sufficiently to form stable structures without drift (left). Relaxation methods like Jacobi and Gauss-Seidel eventually collapse even with hundreds of iterations (right). \textbf{Middle}: A stack of increasingly heavy boxes with a total mass ratio of 4096:1, such poorly conditioned problems are difficult for relaxation methods, leading to significant interpenetration (right). \textbf{Bottom}: An unstructured piling test, our method captures stick/slip transitions and forms stable piles of non-convex objects.}
	\label{fig:stacking}
\end{figure}

\section{Results}

We implemented our algorithm in CUDA and run it on an NVIDIA GTX 1070 GPU. The assembly of the system matrix is performed on the GPU in compressed sparse row (CSR) format for maximum flexibility. We build $\v{H}, \v{M}, \v{J}$, and $\v{C}$ separately and perform the matrix multiply operations necessary for Krylov methods through successive multiplications of individual matrices.

Collision detection is performed between triangle-mesh features once per-time step using the system's unconstrained velocity to generate candidate pairs. We define contact normals as the normalized vector between each feature pair's closest points using the configuration at the start of the time-step. Constraint manifold refinement (CMR) could be used to further improve robustness \cite{otaduy2009implicit}.

\subsection{Experimental Setup}

In this section we discuss the test scenarios we have used for our method. We report scene statistics and performance numbers in Table \ref{tab:stats}. \add{When running in an interactive setting we use a display rate of 60hz which corresponds to a frame time of 16.6ms. For robust collision detection we use two simulation time-steps per visual frame, each with $h = 8.3$ms. If the simulation computation takes longer than this the effect for the user is a slightly slower than real-time update rate}.

\subsubsection{Fetch Tomato}

We test our method on a pick-and-place task using the Fetch robot as shown in Figure \ref{fig:fetch_tomato}. The robot consists of rigid bodies connected by joints as defined by the Unified Robot Description Format (URDF) file. The tomato is modeled using tetrahedral FEM with Young's modulus of $Y=0.1$MPa, Poisson's ratio of $\nu=0.45$, density of $\rho=1000\text{kg}/\text{m}^3$, and a coefficient of friction of $\mu=0.75$ between the grippers and the tomato. The robot is controlled by a human operator who directs the arm and the grippers to grasp the object and transfer it to the mechanical scales. The scales are modeled through rigid bodies connected by prismatic and revolute joints that drive the needle and accurately reads the weight of the tomato. The most challenging part of this scenario is the grasp and transfer of the tomato, which requires tight coupling between frictional contact and the internal dynamics of the tomato. Our method forms stable grasps while undergoing large translational and rotational motion.

\subsubsection{Fetch Beam}

We test our method on a flexible beam insertion task by modeling the beam as a series of connected rigid bodies with finite bending stiffness as shown in Figure \ref{fig:fetch_beam}. The lightweight beam is modeled by 16 rigid bodies each with mass of $m=0.003$kg, and connected through joints with a bending stiffness of $250\text{N.m}$. This example highlights the limitations of traditional relaxation-based approaches that cannot achieve the desired stiffness on the beam's joints even with hundreds of iterations as shown in the convergence plot of Figure \ref{fig:solver}.

\subsubsection{DexNet}

We evaluate our method on the problem of dexterous grasping using the DexNet adversarial object database \cite{mahler2017dex}. These models are highly irregular with many concave areas that make forming stable grasps difficult. The underlying triangle meshes are high resolution and non-manifold which tends to generate many redundant contacts, a challenge for most complementarity solvers. We use the Allegro hand with a coefficient of friction $\mu=0.8$ to perform grasping using a human control interface, and find stable grasps for the objects in collection as shown in Figure \ref{fig:dexnet}.

\subsubsection{PneuNet}

To test coupling between deformable and rigid bodies we simulate a three-fingered gripper based on the PneuNet design \cite{ilievski2011soft}. We model the deformable finger using tetrahedral FEM with a linear isotropic material model and parameters for silicone rubber of $Y = 0.01 \text{GPa}, \nu=0.47, \rho = 1200\text{kg}/\text{m}^3$. To model inflation we use an activation function that uniformly adds an internal volumetric stress to each tetrahedra in the finger arches. We do not model the chamber cavity explicitly, however we found this simple activation model was sufficient to reproduce the characteristic curvature of the gripper. We observe robust coupling through contact by picking up a ball with mass $m=0.32\text{kg}$, using a friction coefficient of $\mu=0.7$ as shown in Figure \ref{fig:pnet}.

\subsubsection{Rigid Body Contact}

We test our method on a variety of rigid-body contact problems as illustrated in Figure \ref{fig:stacking}. Our method achieves stable configurations for difficult problems including self-supported structures, and stacks with extreme mass ratios. For the self-supported arch we use $\mu=0.6$ with masses in the range $m=[15, \cdots, 110]$. For the heavy stack of boxes we use $\mu=0.5$, with masses that increase geometrically as $m=[8, 64, \cdots, 32768]$kg. For the table piling scene each table has a mass of $m=4.7$kg with $\mu=0.7$. Particularly on the scenes with high-mass ratios we observe that relaxation methods struggle to reduce error, while Krylov methods form stable structures and successfully prevent interpenetration. 

\subsubsection{Material Extension}

We perform a material extension test and compare the behavior between a linear co-rotational model and the hyperelastic model of Smith et al. \shortcite{smith2018stable} with Young's modulus set to $E=10^5 \text{Pa}$ and Poisson's ratio of $\nu=0.45$. We visualize volumetric strain and observe high volume loss for the linear model as shown in Figure \ref{fig:fem_stretch}. In the supplementary video we show the impact of geometric stiffness on this test and find it is essential to obtain a stable simulation when strains are large. \add{This is also illustrated in in Figure \ref{fig:geometric_merit} which shows the iterate oscillating around the solution.}

\begin{figure}
	\begin{subfigure}{0.25\textwidth}
		\includegraphics[width=\textwidth,trim=200 0 100 0, clip]{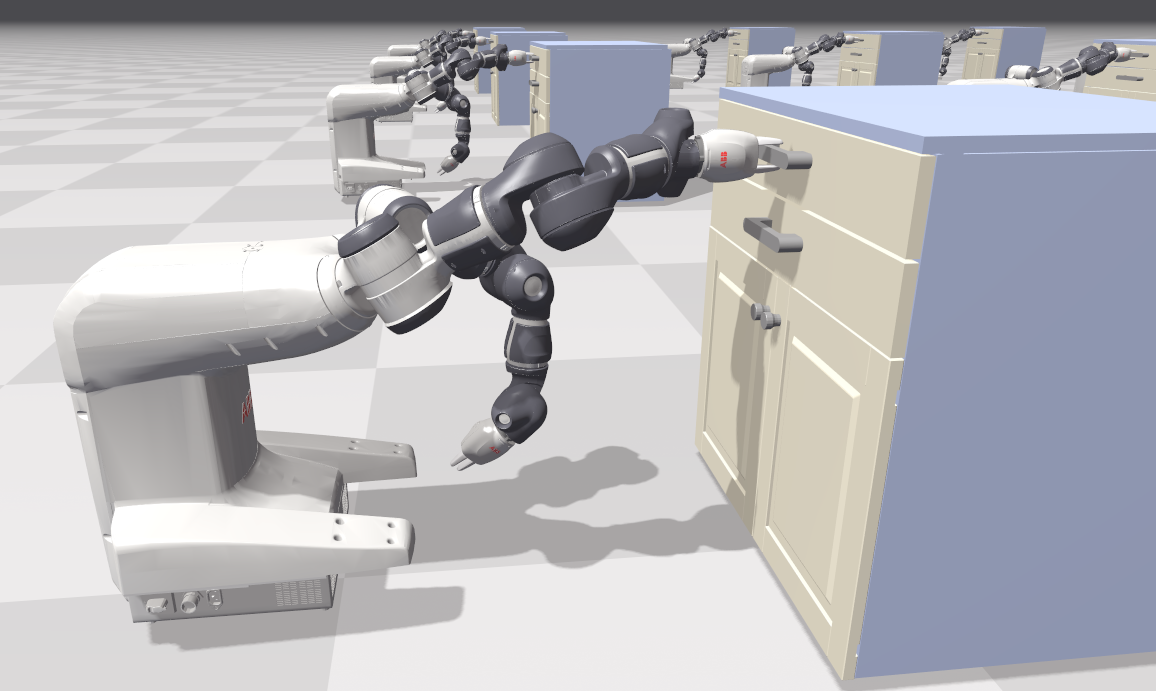}
	\end{subfigure}
	~
	\begin{subfigure}{0.25\textwidth}
		\includegraphics[width=\textwidth,trim=0 0 0 0, clip]{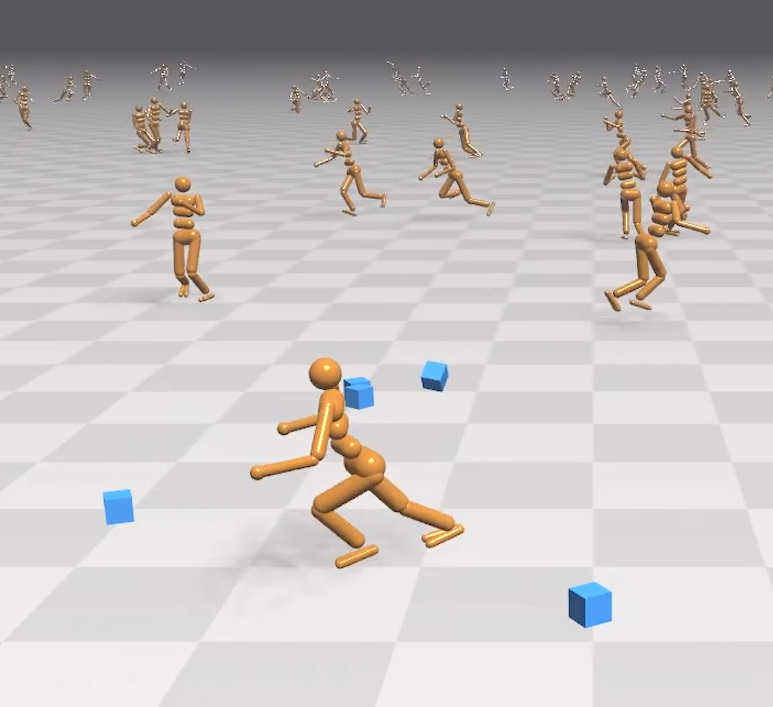}
	\end{subfigure}
		
	\caption{\textbf{Left}: The Yumi robot trained to open a cabinet drawer using reinforcement learning. The network learns a policy that reaches the handle, performs a grasp, and opens it through frictional forces from the fingers alone. \textbf{Right}: Reinforcement learning based locomotion based on the OpenAI Roboschool Humanoid Flagrun Harder environment. Using our simulator the network learns a robust policy that takes advantage of stick-slip transitions to change direction quickly, and recover from external disturbances.}
	\label{fig:reinforcement}
\end{figure}

\subsubsection{Reinforcement Learning}

Reinforcement learning (RL) is a good test of robustness for a simulator since it generates many random inputs in the form of forces, torques, and constraint configurations. We apply our simulator to two scenarios using reinforcement learning. The first is the problem of training the ABB Yumi robot to grasp a cabinet handle and open a drawer as shown in Figure \ref{fig:reinforcement} (left). We use the PPO algorithm \cite{schulman2017proximal} to train the network and find it quickly (less than 100 training iterations) learns a policy to extend, grasp and open the drawer through implicit PD controls applied through the joints. Our second RL example is the Humanoid Flagrun Harder scene adapted from the OpenAI Roboschool \shortcite{openairoboschool}. In this task a humanoid model must learn to stand up and run in a randomly assigned direction that changes periodically as shown in Figure \ref{fig:reinforcement} (right). The learned actions are torques, applied as external forces. Using our simulator we are able to achieve good running results and note the agent taking advantage of stick-slip transitions on the feet during fast turns.

\begin{figure}
					\begin{subfigure}{0.24\textwidth}
		\includegraphics[width=\textwidth]{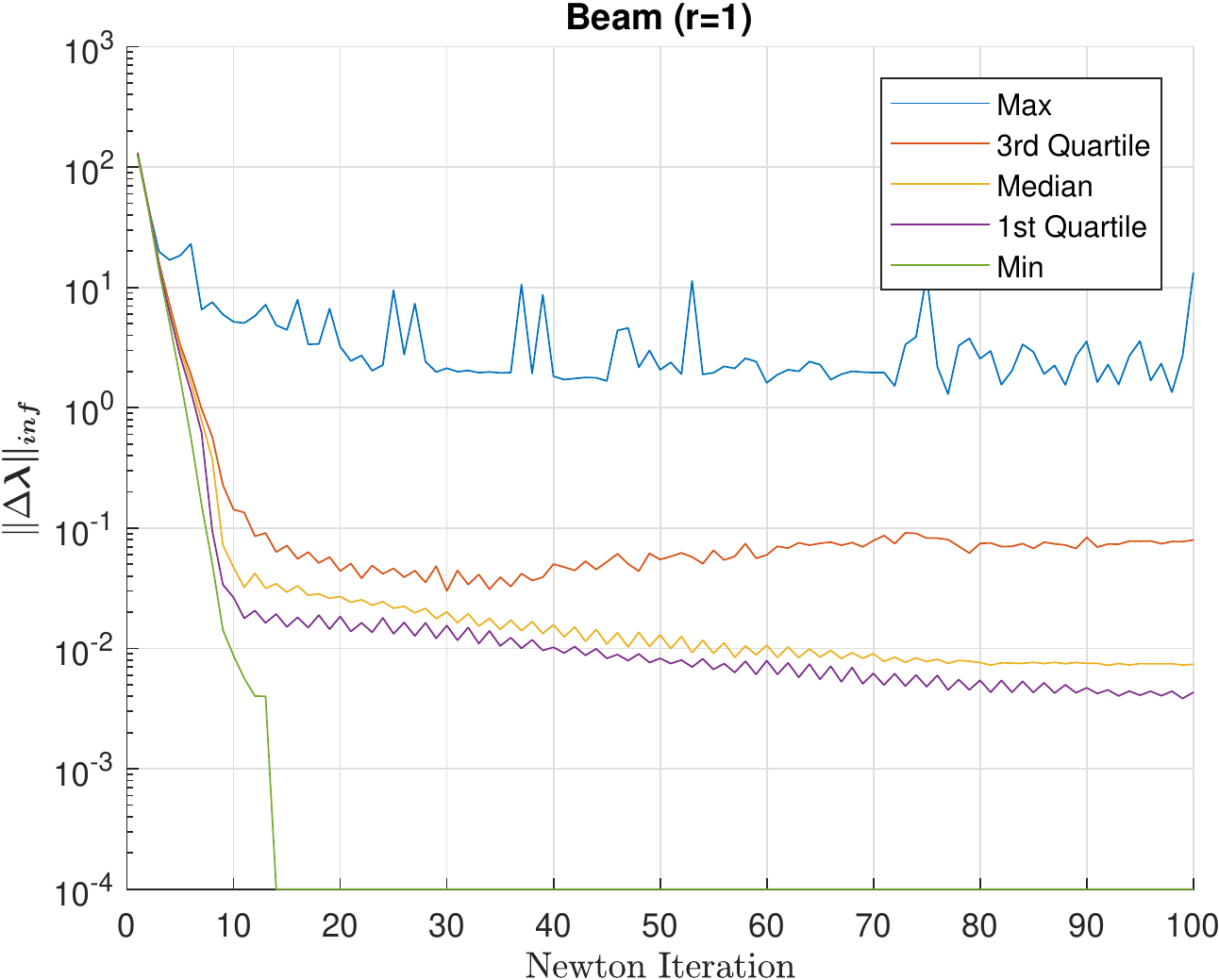}
	\end{subfigure}
	~
	\begin{subfigure}{0.24\textwidth}
		\includegraphics[width=\textwidth]{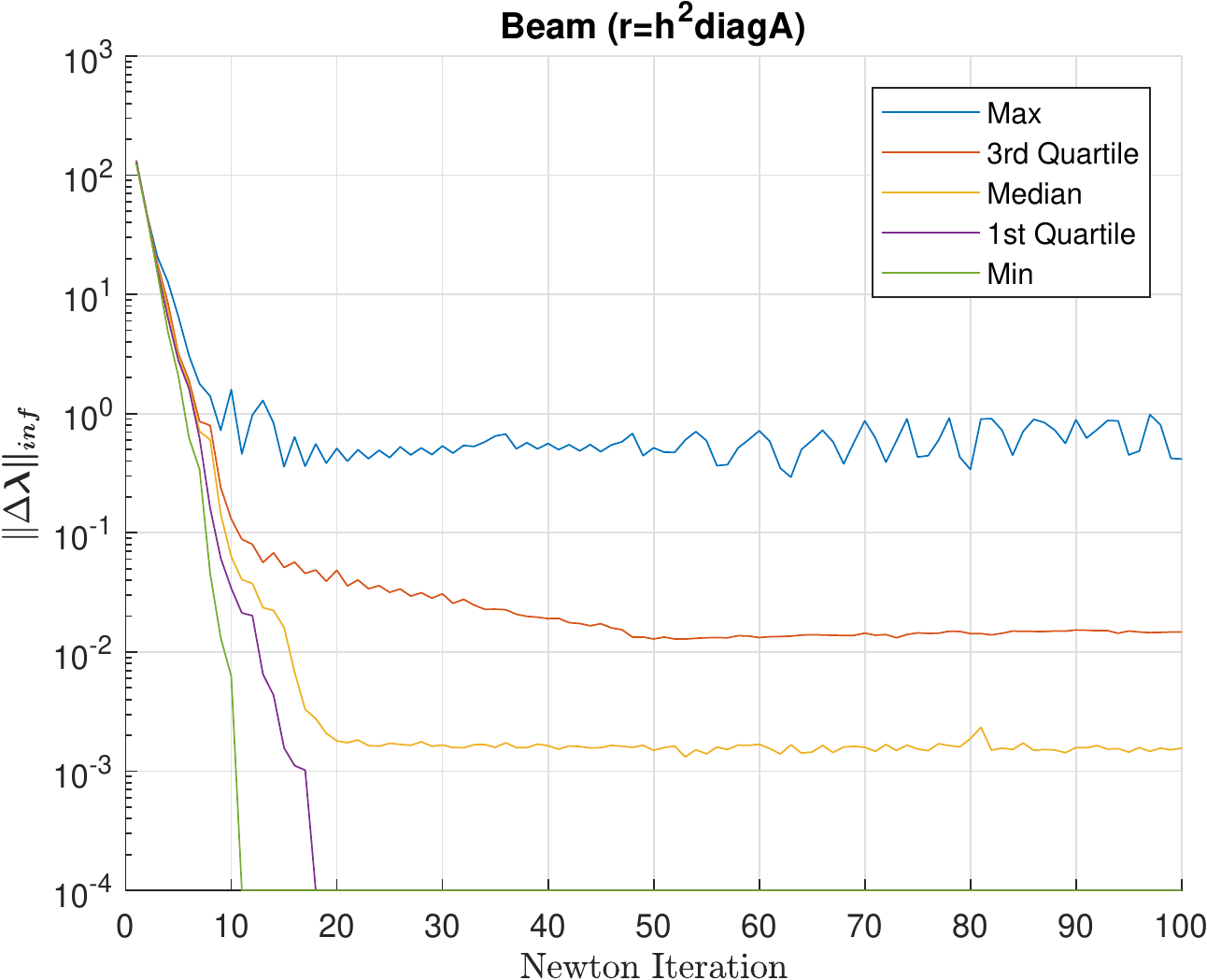}
	\end{subfigure}

	\caption{Quartile analysis of the flexible beam-insertion example with identity scaling (left), and our proposed preconditioning strategy (right). We plot statistics for 100 Newton iterations, taken over 132 simulation-steps. We observe super-linear convergence and lower final residual for significantly more cases with our proposed scheme.}
	\label{fig:quartile}
\end{figure}

\subsection{Effect of Complementarity Preconditiong}\label{sec:rscaling_discussion}

In Figure \ref{fig:quartile} we look at the distribution of convergence behavior over 132 simulation steps during the flexible beam insertion scenario. We use our proposed complementarity preconditioner with 40 PCR iterations per-Newton iteration and observe super-linear convergence in significantly more steps using our preconditioning strategy than with identity scaling. In addition, the median error with our strategy is often an order of magnitude lower for the same iteration count. We use the rigid body contact test scenarios to evaluate the effectiveness of our complementarity preconditioner, and compare the convergence of different strategies in Figure \ref{fig:rscaling}. We observe that identity scaling with $r=1$ may fail badly when there are large masses, or large contact sliding velocities. Constant global scaling by $r=\dt^2$ improves convergence, but still suffers from problems with large mass-ratios. The combination of time-step and effective mass leads to the most reliable observed convergence. Based on the poor performance of identity scaling we believe that a preconditioning strategy such as ours is a necessity to make such non-smooth formulations practical.

\subsection{Effect of NCP-Function}

We found that although both the minimum-map and Fischer-\linebreak Burmeister functions can achieve high accuracy given enough iterations, the minimum-map tends to produce noisy results where the contact forces are not distributed evenly around a contact area. This is primarily a problem when the contact set is redundant, and a small change in the problem may lead to a large change in the active-set. We found the Fischer-Burmeister function was less sensitive to this problem, and would produce smooth contact force distributions even for redundant contact sets. We suspect this is because the minimum-map has many non-smooth points while Fischer-Burmeister has only one, however further investigation to verify this is needed. Due to the improved stability in interactive environments we have used the Fischer-Burmeister function for all examples unless otherwise stated. \add{For scenarios where force distributions are critical, e.g.: force feedback based controllers, it may be appropriate to use a combination of warm-starting to provide temporal coherence, and redistribution of contact forces as a post-process after the contact solve, as proposed by Zheng \& James \shortcite{ZHENG11}. Yet another option is to change the contact model itself by introducing compliance. This makes the problem well-posed by allowing some interpenetration, and may be supported in our formulation by augmenting the compliance block on the contact constraints.}.

\subsection{Effect of Linear Solver}\label{sec:linear_discussion}

In Figure \ref{fig:solver} we compare convergence for different linear solvers over the course of a Newton solve during a single time-step. For performance sensitive applications it is typically not practical or desirable to run each Newton solve to convergence so for this test we fix the number of linear solver iterations per-step to 40. This early termination is generally no problem for relaxation and PCR methods, but it can cause problems for solvers like PCG which decrease the error non-monotonically, leading to erratic convergence. In Figure \ref{fig:linear_solve_convergence} we plot the behavior of each iterative method on a single linear subproblem. Please see the supplementary material for our test matrices and reference PCR implementation in MATLAB format.

The convergence of Jacobi and Gauss-Seidel with our contact formulation is consistent with the behavior observed in other engines \add{such as Bullet \cite{Coumans:2015:BPS:2776880.2792704} and XPBD \cite{macklin2016xpbd}}. While relaxation methods perform quite well for reasonably well-conditioned problems, they are very slow to converge for situations involving high mass-ratios as shown in Figure \ref{fig:solver}. This is highlighted in the heavy-stack example which shows catastrophic interpenetration.

\subsection{Error Analysis}

To better understand our results we perform an error analysis to establish a baseline accuracy limit given finite precision floating point. Our analysis is based on that given by Tisseur \shortcite{tisseur2001newton} who shows that Newton's method applied to the problem of solving $\v{r}(\v{x})=\v{0}$ will have a limiting step-size, or solution accuracy of

\begin{align}
\|\Delta\v{x}\| \approx \|\v{A}_*^{-1}\|\nu + \|\v{x}_*\|\delta.
\end{align}

Here $\v{x}_*$ is the true solution, $\v{A}_*$ is the system Jacobian at the solution, $\nu$ is an upper bound on the residual error, and $\delta$ is the machine epsilon.  In general we do not know the true solution $\v{x}^*$ so we use the lowest error solution as an approximation. Likewise we use $\nu = \delta\|\v{r}\|$ as the residual error bound, where $\v{r}$ is the lowest achieved error. Likewise, the minimum residual is limited by available precision and Tisseur show that its predicted limiting magnitude is 

\begin{align}
\|\v{r}\| \approx \|\v{A}_*\|\|\v{x}_*\|\delta.
\end{align}

\begin{figure*}[p]	
	\centering
	\begin{subfigure}{0.3\textwidth}
		\centering
		\includegraphics[width=\textwidth]{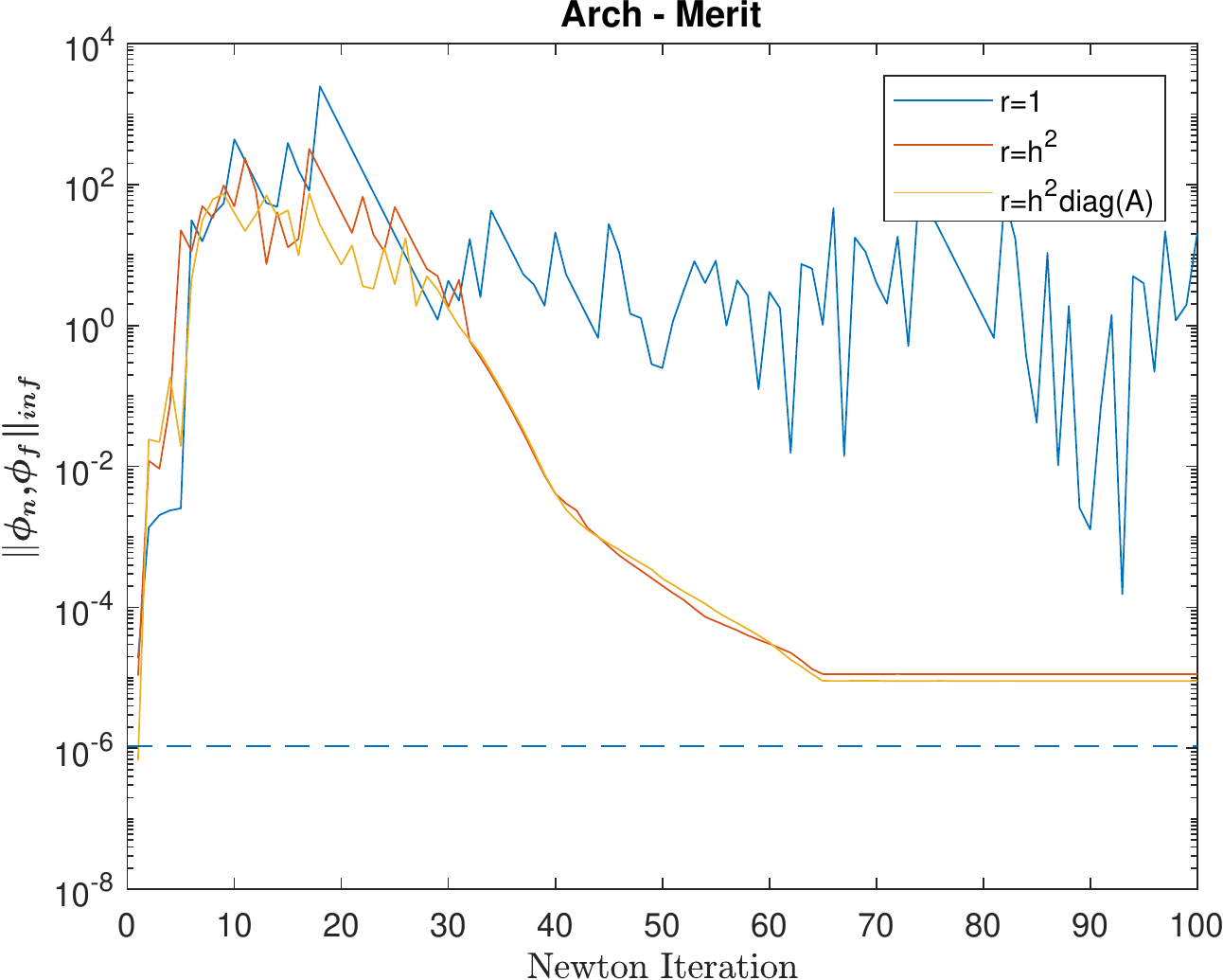}
	\end{subfigure}
	~	
	\begin{subfigure}{0.3\textwidth}
		\centering
		\includegraphics[width=\textwidth]{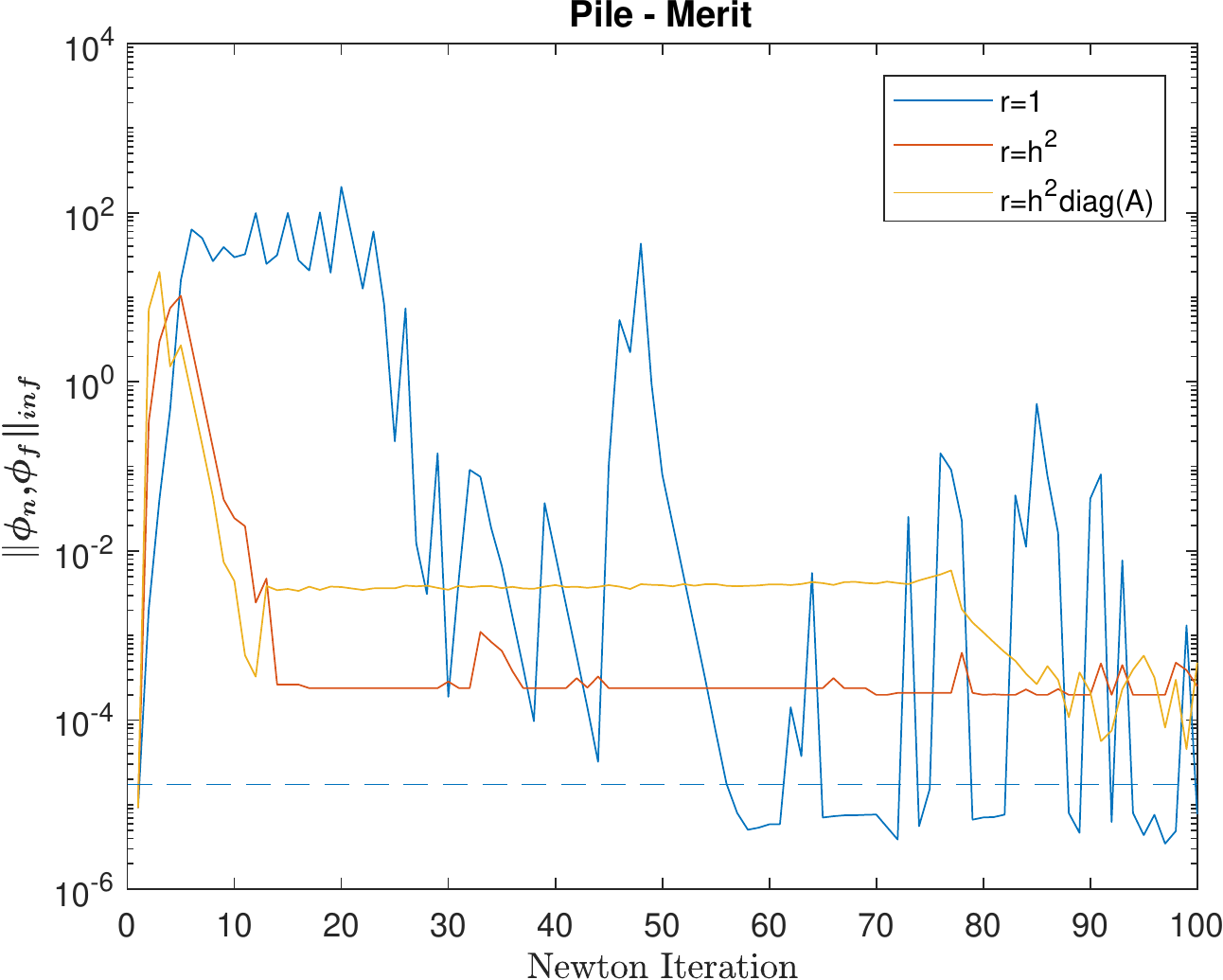}
	\end{subfigure}
	~
	\begin{subfigure}{0.3\textwidth}
		\centering
		\includegraphics[width=\textwidth]{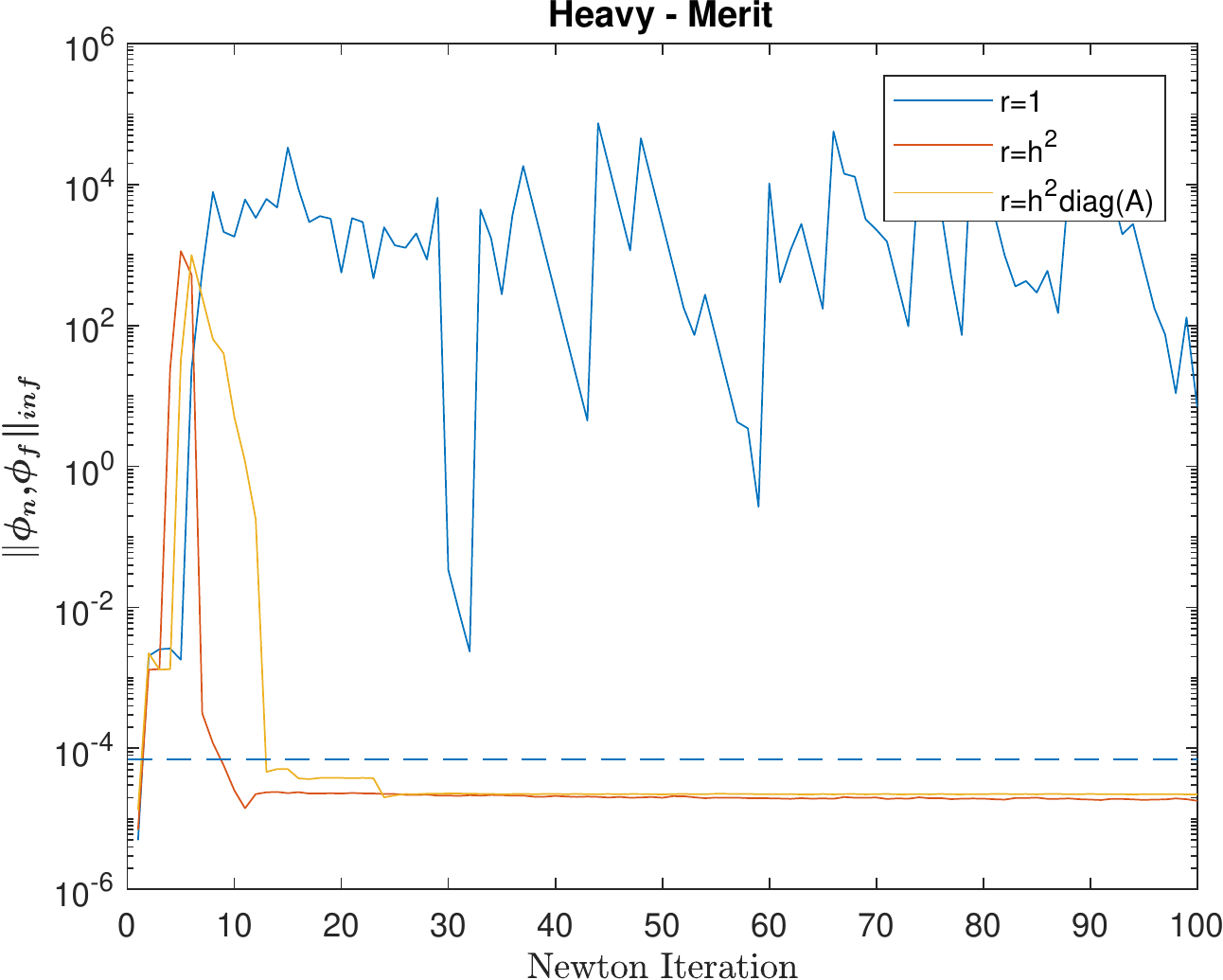}
	\end{subfigure}
	
	\par\bigskip
	
	\begin{subfigure}{0.3\textwidth}
		\centering
		\includegraphics[width=\textwidth]{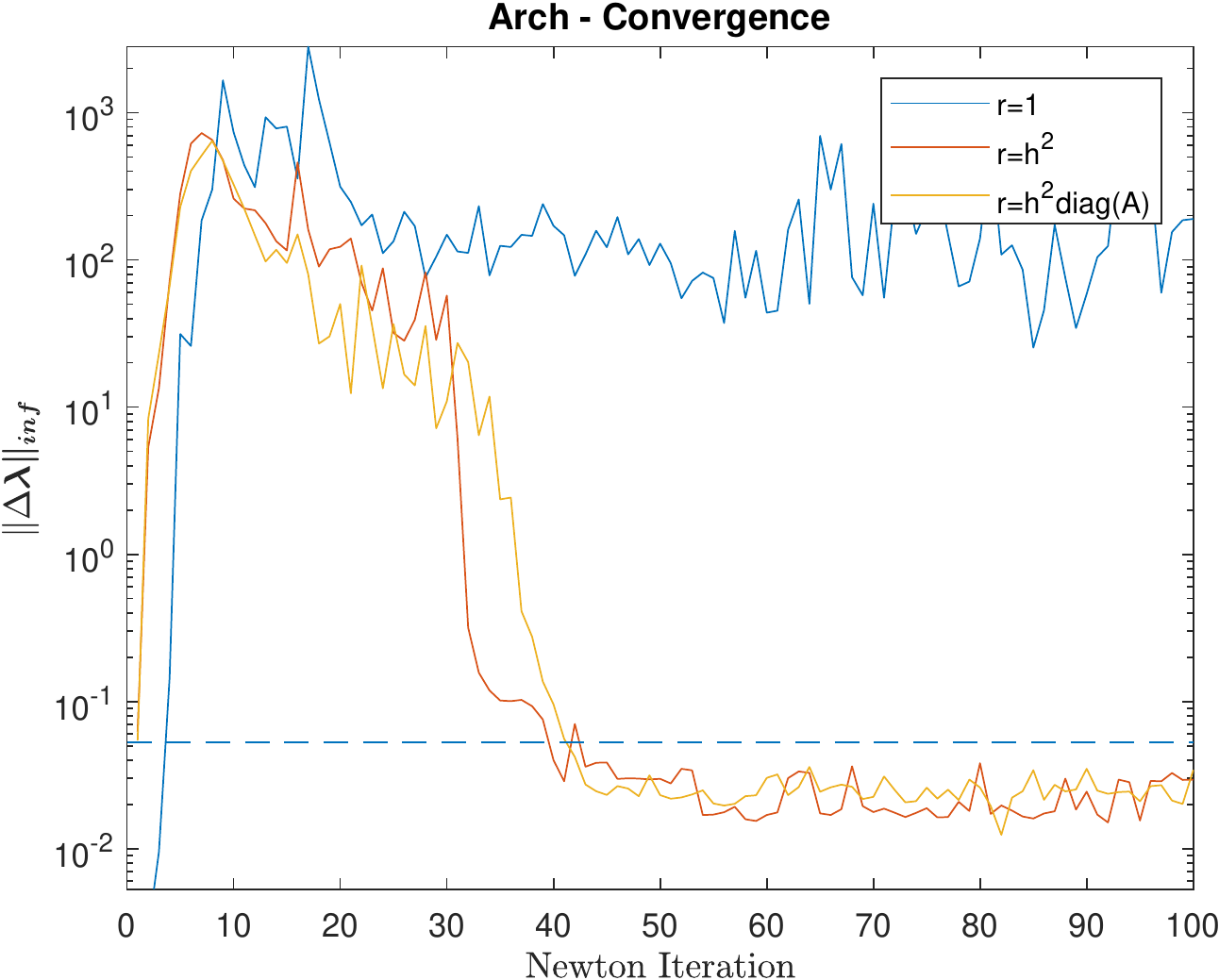}
	\end{subfigure}
	~
	\begin{subfigure}{0.3\textwidth}
		\centering
		\includegraphics[width=\textwidth]{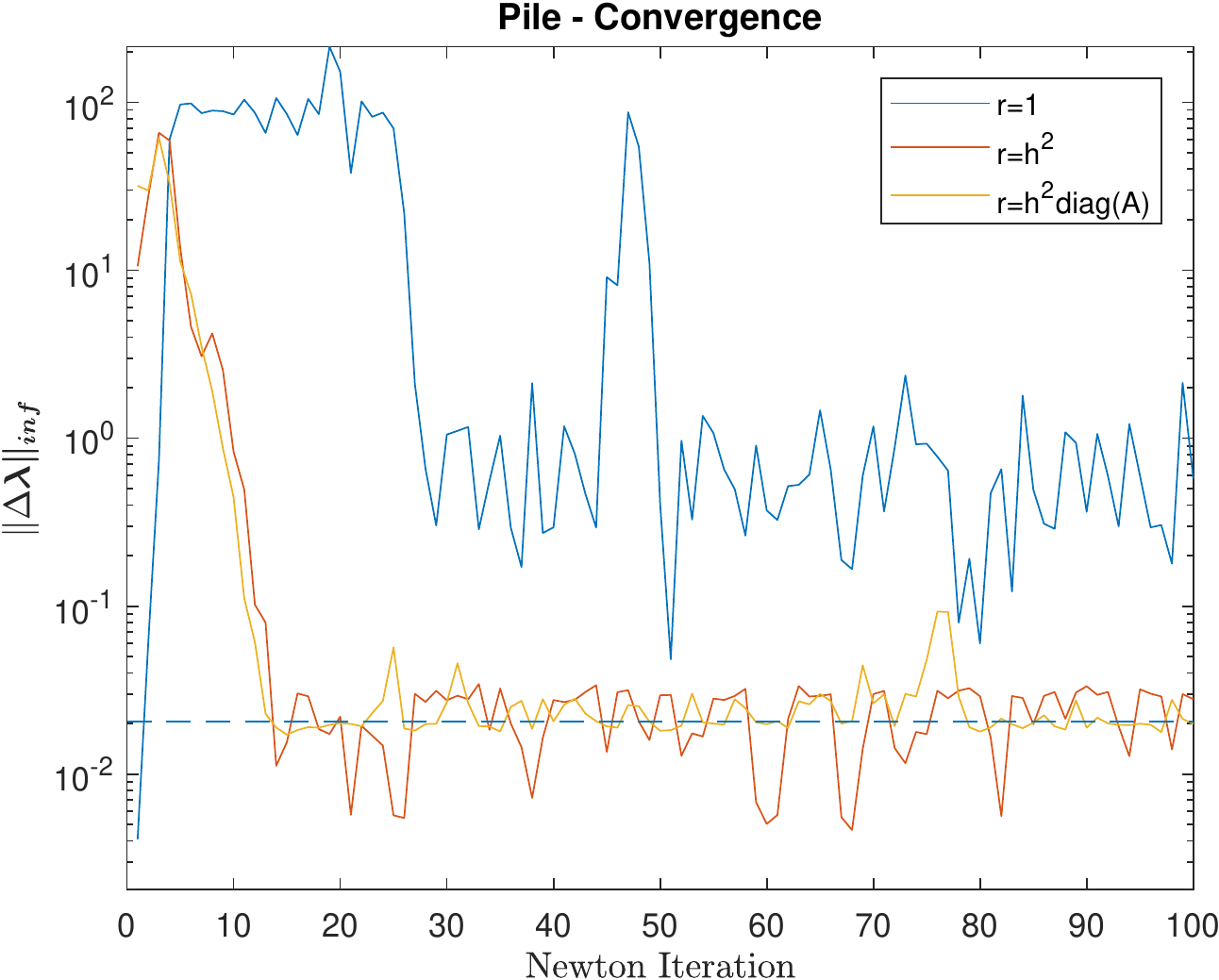}
	\end{subfigure}
	~
	\begin{subfigure}{0.3\textwidth}
		\centering
		\includegraphics[width=\textwidth]{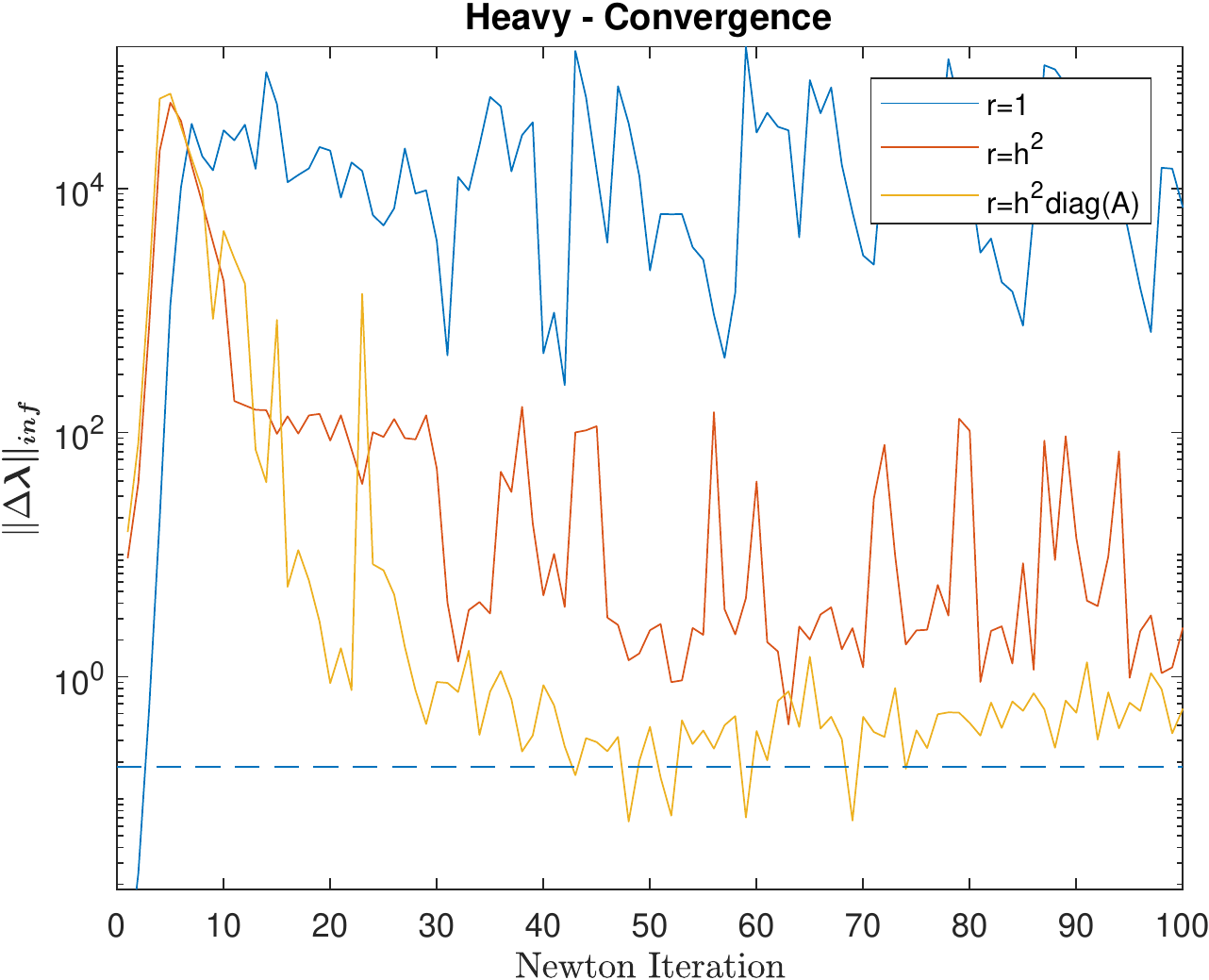}
	\end{subfigure}
	
	\caption{An evaluation of different complementarity preconditioning strategies on three contact problems. We plot the maximum complementarity error for normal and frictional forces (upper row), and the step size (lower row) for each Newton iteration over a single time-step. Identity scaling (blue) often fails to converge. A time-step scaled strategy with (red) performs well when mass-ratios are small, such as in the arch and pile scenes (left, middle). For situations with high-mass ratios (right) we also take into account the system's effective mass (yellow).}
	\label{fig:rscaling}
	\par\bigskip
\centering
\begin{subfigure}{0.3\textwidth}
	\centering
	\includegraphics[width=\textwidth]{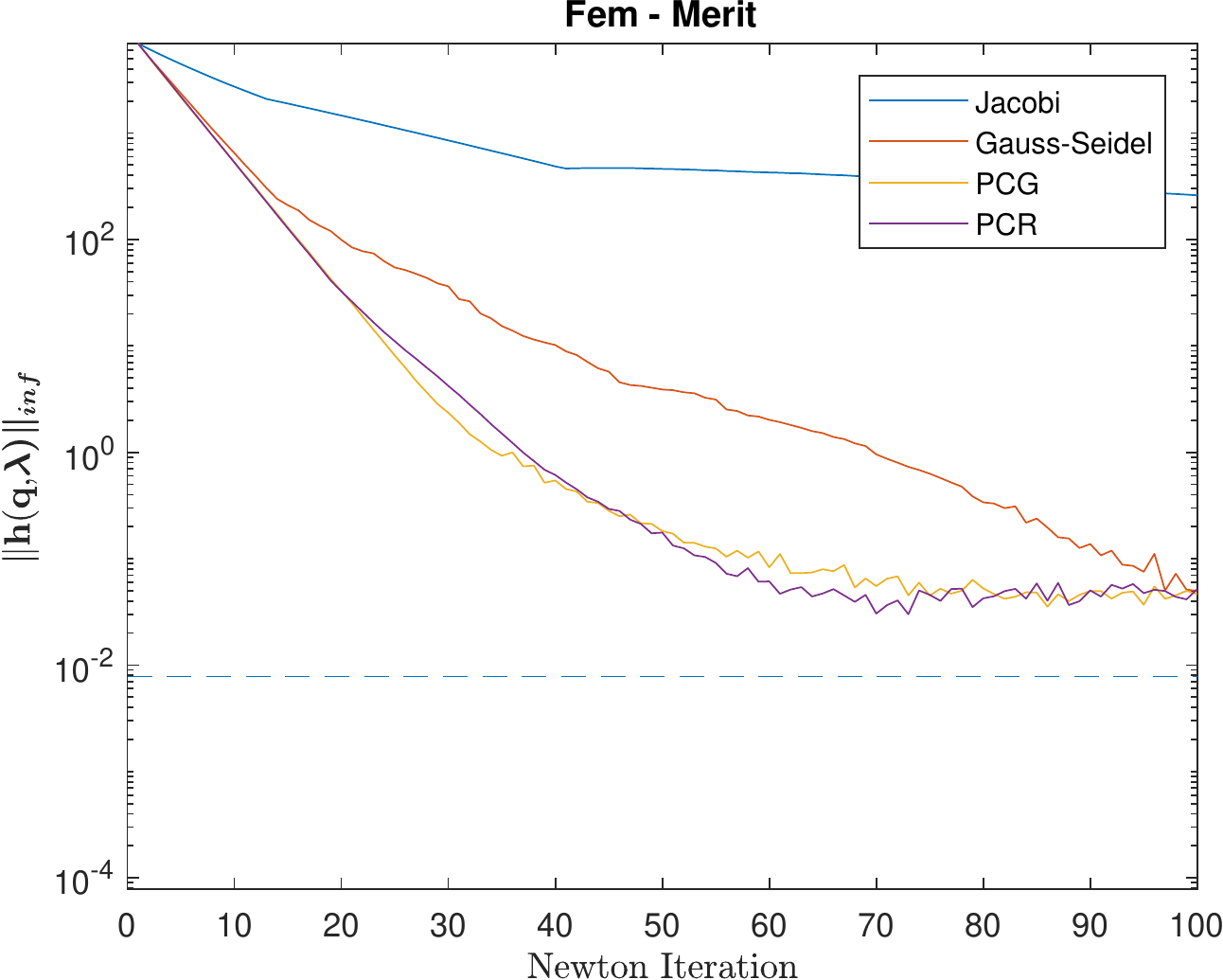}
\end{subfigure}
~
\begin{subfigure}{0.3\textwidth}
	\centering
	\includegraphics[width=\textwidth]{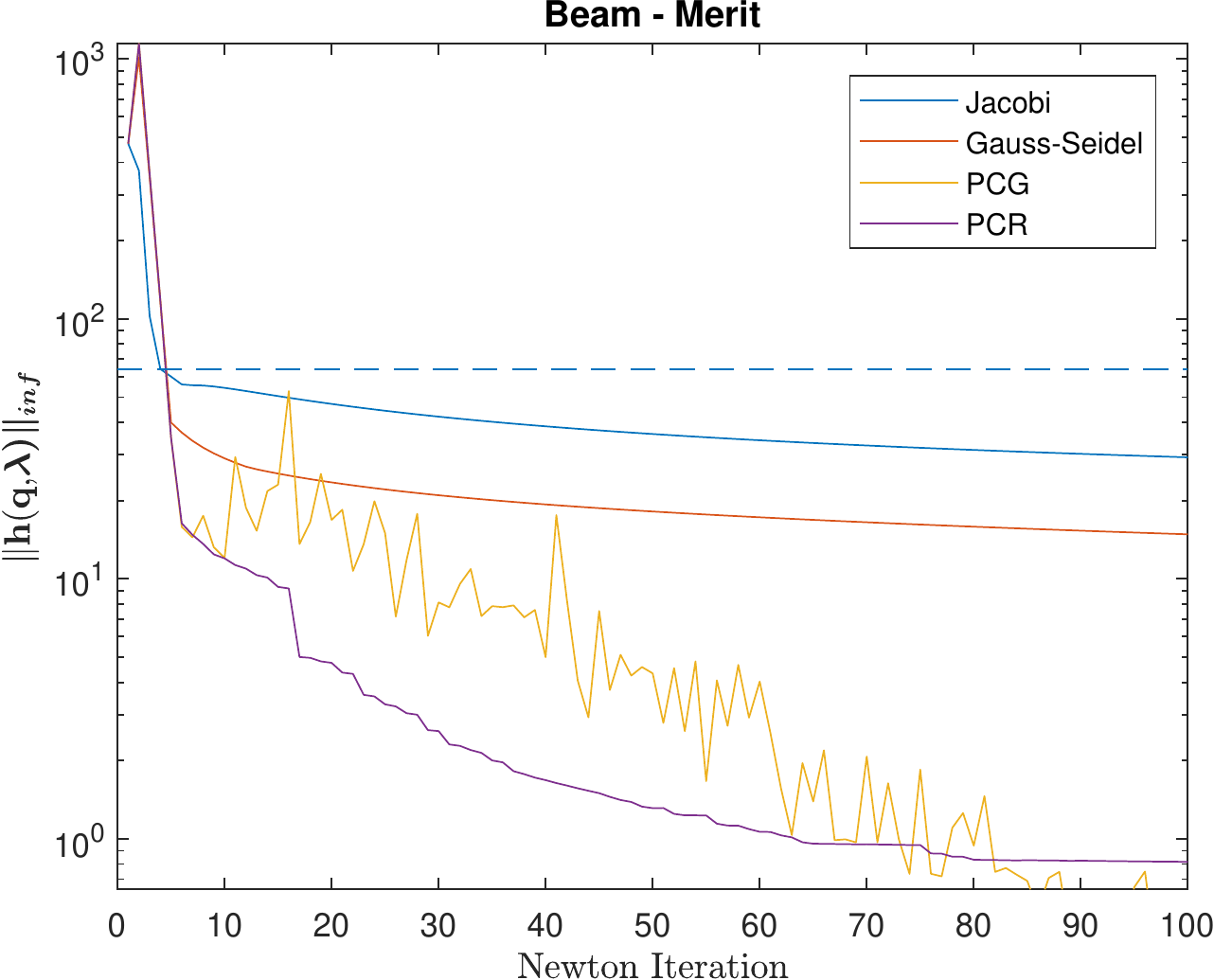}
\end{subfigure}
~
\begin{subfigure}{0.3\textwidth}
	\centering
	\includegraphics[width=\textwidth]{images/heavy_merit.pdf}
\end{subfigure}

\par\bigskip

\begin{subfigure}{0.3\textwidth}
	\centering
	\includegraphics[width=\textwidth]{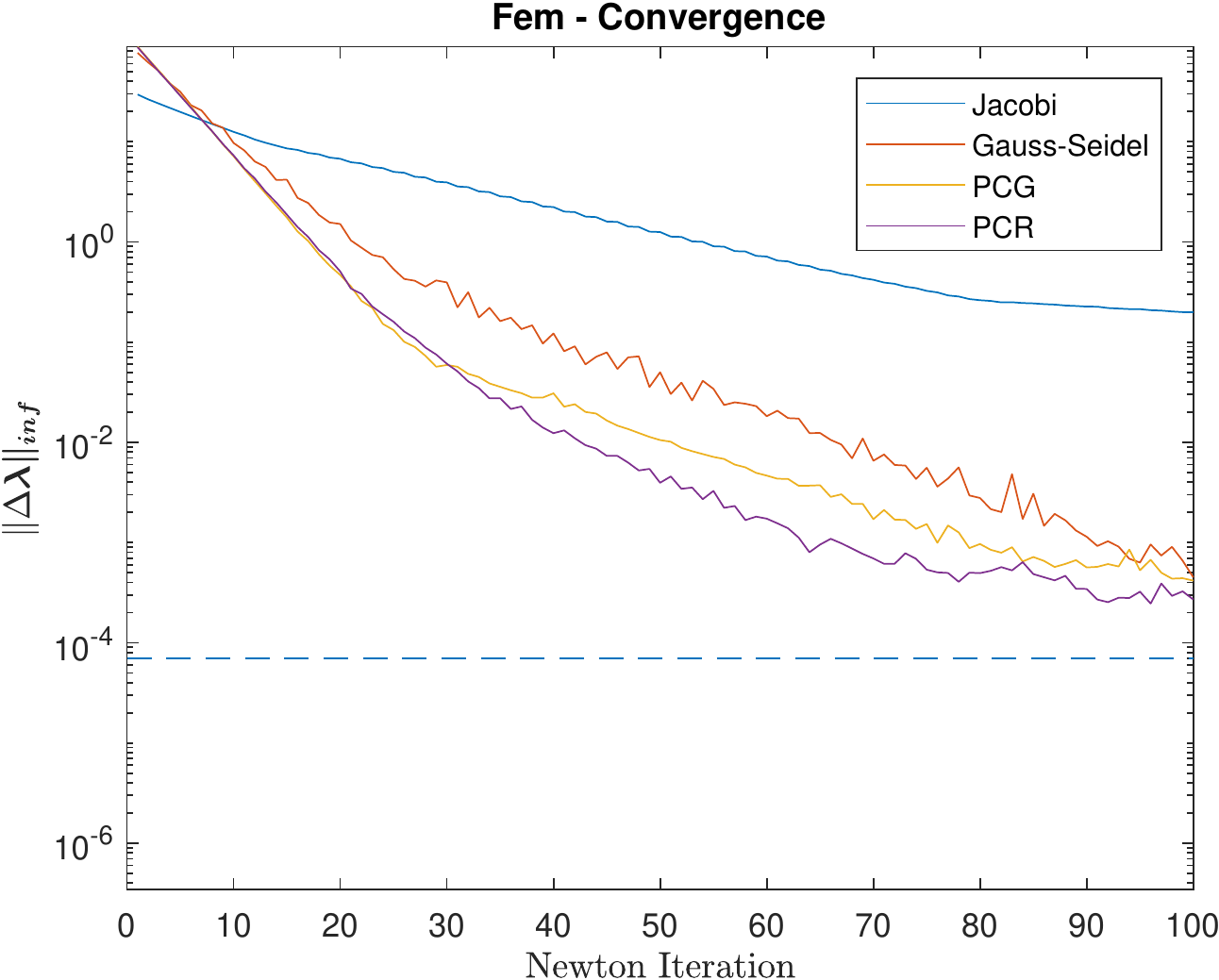}
\end{subfigure}
~
\begin{subfigure}{0.3\textwidth}
	\centering
	\includegraphics[width=\textwidth]{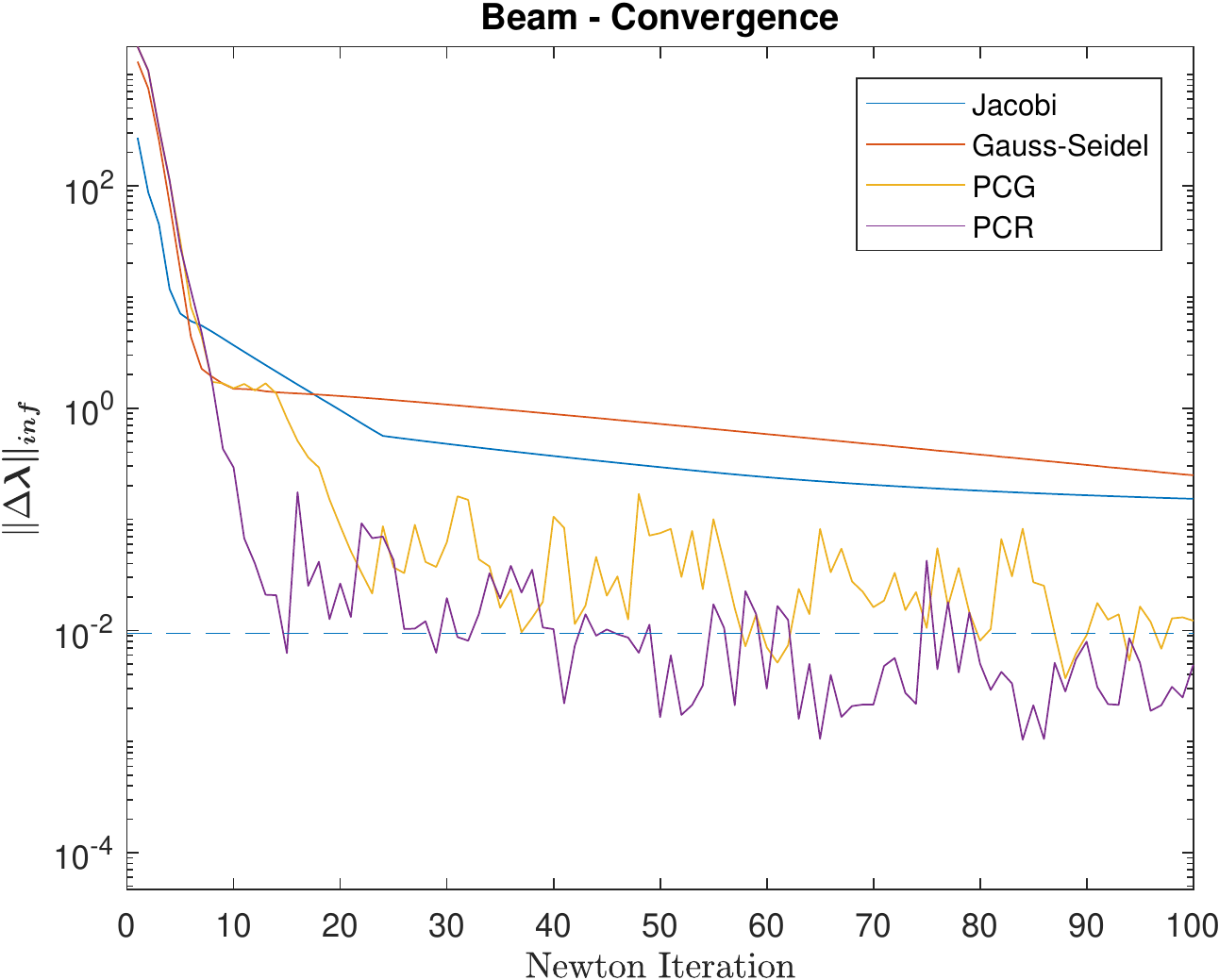}
\end{subfigure}
~
\begin{subfigure}{0.3\textwidth}
	\centering
	\includegraphics[width=\textwidth]{images/heavy_convergence.pdf}
\end{subfigure}
\caption{A comparison of different iterative methods on three test cases. We plot the maximum constraint error (upper row) and the step size (lower row) for each Newton iteration over a single time-step. Jacobi and Gauss-Seidel methods show characteristic stagnation for ill-conditioned problems (blue, red). Conjugate Gradient (yellow) is less sensitive to conditioning but has large residual fluctuations. Conjugate Residual (purple) provides fast convergence and smooth error reduction. The dashed lines represent the limiting accuracy and residual bounds predicted by an error analysis assuming 32-bit floating point.}
\label{fig:solver}

\end{figure*}

We use 32-bit floating point for all calculations, and plot these limiting values for the $L_\infty$ norm as dashed lines in Figures \ref{fig:rscaling}-\ref{fig:solver}. We find that this error model does a good job of predicting the observed accuracy, with the exception of the flexible beam example where the predicted residual accuracy is under-estimated.

\section{Limitations and Future Work}

\add{In robotics it is common to use a reduced coordinate representation to model rigid articulations. This establishes hard constraints directly in the system degrees of freedom, and reduces the number of parameters and explicit constraint equations. We have derived our formulation in terms of generalized coordinates, so our method should be applicable to any parameterization simply by using the appropriate mass matrix $\v{M}$, kinematic map $\v{G}$, and constraint Jacobians.}

Krylov methods such as PCG and PCR use a globally optimal line-search step at each iteration. This has the sometimes undesirable effect that error in one row can affect the rate of residual reduction in other rows, leading to slower than necessary convergence for independent parts of the simulation. A natural solution would be to include an island-detection step and solve these independent systems separately.

We have not dealt specifically with elastic collisions, or energy preserving integrators, however we believe that the approach of Smith et al. \shortcite{smith2012reflections} can be combined with non-smooth complementarity formulations such as ours.

We found our geometric stiffness approximation to be quite effective on particle-based objects, but less so on rigid-articulated mechanisms, and in the worst case can cause some jitter at joint limits. To avoid this we apply geometric stiffness only to the particle degrees of freedom. In the future we plan to explore more advanced quasi-Newton methods e.g.: those based on symmetric rank-1 (SR1) updates.

Our method's primary computational cost is the solution of a symmetric linear system at each step. All the iterative solvers considered here may be implemented in a matrix-free way which would likely improve performance. \rem{In addition we plan to explore the space of complementarity preconditioners.} \add{Because the linear solver is treated as a black box it would be interesting to apply more advanced methods, such as algebraic multi-grid (AMG), to accelerate convergence for larger-scale linear subproblems}.

\add{Finally, the design space for choosing the $r$ factor in the complementarity preconditioner is large, and we think that heuristics based on global information could provide significant performance improvements.}

\section{Conclusion}

We have presented a framework for multi-body dynamics that allows off-the-shelf linear solvers to be used for rigid and deformable contact problems. We evaluate our method on a variety of scenarios in robotics and found that it performs well on grasping and dexterous manipulation of deformable objects, as well as ill-conditioned rigid body contact problems. We believe our complementarity preconditioner makes non-smooth formulations of contact practical for interactive applications for the first time, and hope that this opens the door for future works to apply new contact models, preconditioners, and linear solver methods to multi-body problems.

\begin{acks}

The authors would like to thank Ken Goldberg and Jeff Mahler at Berkeley AUTOLAB for their help with the DexNet models. Many thanks to Jacky Liang for creating the Fetch flexible beam insertion environment and the Allegro hand interface. We also thank Yevgen Chebotar for contributing the RL Yumi scene, Jan Issac for creating the cabinet model, and Richard Tonge for creating the table model and piling scene. The Allegro hand URDF and models are provided courtesy SimLab Co., Ltd. The Fetch robot URDF and model is provided courtesy of Fetch Robotics, Inc. The Yumi robot URDF is provided courtesy of ABB robotics. The humanoid model is adapted from the MuJoCo model created by Vikash Kumar.

\end{acks}

\bibliographystyle{ACM-Reference-Format}
\bibliography{nonsmoothnewton}

\appendix

\section{Contact Formulation}\label{app:phif}

In this appendix we give the full form of the contact constraints and derivatives, including the $r$-factor scaling parameter, for the minimum-map and Fischer-Burmeister NCP-functions. 

\subsection{Contact}
The contact constraint given in terms of the Fischer-Burmeister function can be written as,

\begin{align}
{\phi_{n}}_\text{FB} = C_n(\q) + r\lambda_n - \sqrt{C_n(\q)^2 + r^2\lambda_n^2},
\end{align}
\noindent which has the following derivatives,

\begin{align}
\pdv{{\phi_{n}}_\text{FB}}{\q} &= \left(1-\frac{C_n(\q)}{\sqrt{C_n(\q)^2 + r^2\lambda_n^2}}\right)\nabla C_n \\
\pdv{{\phi_{n}}_\text{FB}}{\lambda_{n}} &= \left(1-\frac{r\lambda_n}{\sqrt{C_n(\q)^2 + r^2\lambda_n^2}}\right)r.
\end{align}
Likewise, for the minimum-map,

\begin{align}
{\phi_{n}}_\text{min} &= \begin{cases} C_n & C_n(\q) \le r\lambda_{n} \\
r\lambda_{n} & \text{otherwise}\end{cases}
\end{align}
\noindent with derivatives given by,

\begin{align}
\pdv{{\phi_{n}}_\text{min}}{\q} &= \begin{cases} \nabla C_n & C_n(\q) \le r\lambda_{n} \\
\v{0} & \text{otherwise}\end{cases} \\
\pdv{{\phi_{n}}_\text{min}}{\lambda_{n}} &= \begin{cases} 0 & C_n(\q) \le r\lambda_{n} \\
r & \text{otherwise}\end{cases}.
\end{align}

\subsection{Friction}

We now give the closed form of the frictional compliance terms including $r$-factor scaling. First, we write in full the Fischer-Burmeister NCP-function for the frictional conditions \eqref{eq:friction_complementarity},

\begin{align*}
{\psi_f}_\text{FB} &=|\v{D}^T\qdot| + r(\mu\lambda_{n} - |\gv{\lambda}_f|) - \sqrt{|\v{D}^T\qdot|^2 + r^2(\mu\lambda_{n} - |\gv{\lambda}_f|)^2}.
\end{align*}

We use the equation ${\psi_f}_\text{FB} = 0$ to obtain a fixed point iteration for the quantities $|\v{D}^T\qdot|$ and $|\gv{\lambda}_f|$ as follows,

\begin{align}
|\v{D}^T\qdot|^{n+1} &= |\v{D}^T\qdot| - {\psi_f}_\text{FB}(\qdot, \gv{\lambda}_f) \\
&= \sqrt{|\v{D}^T\qdot|^2 + r^2(\mu\lambda_{n} - |\gv{\lambda}_f|)^2} - r(\mu\lambda_{n} - |\gv{\lambda}_f|) 
\end{align}
\begin{align}
|\gv{\lambda}_f|^{n+1} &= |\gv{\lambda}_f| + r^{-1}{\psi_f}_\text{FB}(\qdot, \gv{\lambda}_f) \\
&= r^{-1}\left(|\v{D}^T\qdot|^2 + r\mu\lambda_n - \sqrt{|\v{D}^T\qdot|^2 + r^2(\mu\lambda_{n} - |\gv{\lambda}_f|)^2}\right).
\end{align}
Using these we can write our frictional compliance quantity $W_{FB}$,

\begin{align}
W_\text{FB} &= \frac{|\v{D}^T\qdot|^{n+1}}{|\gv{\lambda}_f|^{n+1}} \\ 
 &= r\left(\frac{\sqrt{|\v{D}^T\qdot|^2 + r^2(\mu\lambda_{n} - |\gv{\lambda}_f|)^2} - r(\mu\lambda_{n} - |\gv{\lambda}_f|)}{|\v{D}^T\qdot|^2 + r\mu\lambda_n - \sqrt{|\v{D}^T\qdot|^2 + r^2(\mu\lambda_{n} - |\gv{\lambda}_f|)^2}}\right).
\end{align}
Following the same pattern we find that $W_\text{min}$ for the minimum-map is given by,
\begin{align}
W_\text{min} &= \begin{cases}
0 & |\v{D}^T\qdot| \le r(\mu\lambda_{n} - |\gv{\lambda}_f|) \\
\frac{|\v{D}^T\qdot| - r(\mu\lambda_n - |\gv{\lambda}_f|)}{\mu\lambda_n}, & \text{otherwise} \end{cases}.
\end{align}

Using the $W$ for one or the other NCP-function, the compliance matrix sub-block for a single contact with index $i$ is

\begin{align}
\v{W}_{ii} &= \pdv{\gv{\phi}_{f,i}}{\gv{\lambda}_{f,i}} \begin{cases}\v{1}_{2\times 2}W_i & \mu\lambda_{n,i} > 0 \\ \\ \v{1}_{2\times 2} & \text{otherwise} \end{cases}
\end{align}
	
\section{Compliance Form of Stable Neo-Hookean Materials}\label{app:smith}

Here we give the derivatives required for the compliance form of the stable Neo-Hookean material model introduced by Smith et al \shortcite{smith2018stable}. The elastic strain-energy density is given by
\begin{align}
\Psi_E = C_1(I_C - 3) + D_1(J - \alpha)^2.
\end{align}

The material constants $C_1$, $D_1$, and $\alpha$ can be related to the Lam\'{e} parameters according to the original paper. Defining $\v{s} = [s_1, s_2, s_3]^T$ as the vector of principal stretches of the deformation gradient $\v{F}$ we have $I_C = s_1^2 + s_2^2 + s_3^2$ the first invariant of stretch, and $J = s_1s_2s_3$, the relative volume change induced by $\v{F}$. To perform the compliance transformation we require the Jacobian and Hessian of $\Psi$ with respect to $\v{s}$, which we provide below:

\begin{align*}
& \pdv{\Psi_E}{\v{s}} = 2C_1\begin{bmatrix} s_1 \\ s_2 \\ s_3\end{bmatrix}^T + 2D_1(J-1)\begin{bmatrix} s_2s_3 \\ s_1s_3 \\ s_1s_2\end{bmatrix}^T \\ \\
& \pdv[2]{\Psi_E}{\v{s}} = 2\begin{bmatrix}D_1 s_2^2  s_3^2  + C_1 & k_1 & k_2 \\ k_1 & D_1 s_1^2  s_3^2 + C_1 & k_3 \\ k_2 & k_3 & D_1 s_1^2  s_2^2 + C_1\end{bmatrix} \\ \\
& k_0 = 2J - \alpha, \\
& k_1 = D_1 s_3 k_0, \\ 
& k_2 = D_1 s_2 k_0, \\
& k_3 = D_1 s_1 k_0.
\end{align*}

For a constant strain tetrahedron with rest volume $V_e$ the elastic potential energy is $U(\q) = V_e\Psi_E(\v{s})$, and the compliance matrix is $\v{E} = \left(V_e\pdv[2]{\Psi_E}{\v{s}}\right)^{-1}$ which can be obtained through a $3\times 3$ matrix inverse. The derivatives of the singular values with respect to the vertices, $\v{J} = \pdv{\v{s}}{\v{q}}$, are given in \cite{perez2013strain}.

\end{document}